\documentclass[11pt,logo,letterpaper]{berkeley}

\usepackage[square,sort,comma,numbers]{natbib}

\usepackage[utf8]{inputenc} 
\usepackage[T1]{fontenc}    
\usepackage{hyperref}       
\usepackage{color}
\usepackage{url}            
\usepackage{booktabs}       
\usepackage{amsfonts}       
\usepackage{nicefrac}       
\usepackage{microtype}      
\usepackage{xcolor}         

\usepackage{adjustbox}
\usepackage{titletoc}
\usepackage{latexsym}
\usepackage{amsfonts}
\usepackage{multirow}
\usepackage{array}
\usepackage{caption}
\usepackage{booktabs}
\usepackage{colortbl}
\usepackage{xcolor}
\usepackage{multicol}
\usepackage{bm}
\usepackage{graphicx}
\usepackage{subfigure}
\usepackage{amsthm,amsmath,amssymb}
\usepackage{mathrsfs}
\usepackage{enumitem}
\usepackage{algorithm,algpseudocode}

\usepackage{tikz}
\usepackage{enumitem}
\usepackage{wrapfig}
\usepackage{float}
\usepackage{graphicx}
\usepackage{caption}
\usepackage{color}
\usepackage{colortbl}
\usepackage{bbding}
\usepackage{pifont}
\usepackage{tabularx} 
\usepackage{hyperref}
\usepackage{url}
\usepackage[utf8]{inputenc} 
\usepackage[T1]{fontenc}    
\usepackage{hyperref}       
\usepackage{url}            
\usepackage{booktabs}       
\usepackage{amsfonts}       
\usepackage{nicefrac}       
\usepackage{microtype}      
\usepackage{xcolor}         
\usepackage{makecell}
\usepackage{graphicx}
\usepackage{amsmath}
\usepackage{multirow}
\usepackage{wrapfig}
\usepackage{tcolorbox}
\usepackage{fontawesome}

\usepackage{colortbl}
\usepackage{threeparttable}
\usepackage{bxcoloremoji}

\usepackage{marvosym}

\usepackage{tikz}
\usetikzlibrary{shapes.geometric, arrows.meta, positioning, fit}

\newcommand{\github}{\raisebox{-1.5pt}{\includegraphics[height=1.05em]{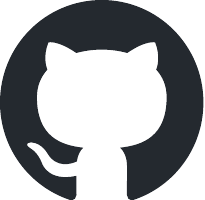}}}

\usepackage[edges]{forest}
\definecolor{hidden-draw}{RGB}{0,0,0}
\definecolor{hidden-pink}{rgb}{0.98, 0.94, 0.75}
\definecolor{level0}{rgb}{0.67, 0.88, 0.69}
\definecolor{level1}{rgb}{0.98, 0.92, 0.84}
\definecolor{level2}{rgb}{0.8, 0.8, 1.0}
\definecolor{level3}{rgb}{1.0, 0.71, 0.76}
\definecolor{level4}{rgb}{0.49, 0.99, 0.0}
\definecolor{level5}{rgb}{0.87, 0.63, 0.87}
\definecolor{darkblue}{rgb}{0, 0, 0.5}

\definecolor{TikTokPink}{HTML}{FF2B54} 
\definecolor{CalGoldHex}{HTML}{FDF7F2}
\definecolor{IronGrey}{HTML}{6D6E71}
\definecolor{YaleBlue}{HTML}{2A5487}
\definecolor{LinkPurple}{HTML}{FF2B54}

\definecolor{hidden-red}{RGB}{205, 44, 36}
\definecolor{hidden-blue}{RGB}{194,232,247}
\definecolor{hidden-orange}{RGB}{243,202,120}
\definecolor{hidden-green}{RGB}{34,139,34}
\definecolor{hidden-pink}{RGB}{255,245,247}
\definecolor{hidden-black}{RGB}{20,68,106}
\definecolor{purple}{RGB}{144,153,196}
\definecolor{yellow}{RGB}{255,228,123}
\definecolor{hidden-yellow}{RGB}{255,248,203}
\definecolor{tkcolor}{RGB}{224,223,255}
\definecolor{darkblue}{rgb}{0, 0.40, 0.75}
\newcommand{\eg}{\textit{e.g.,}}

\hypersetup{colorlinks=true, linkcolor=YaleBlue}

\newcommand{\MYhref}[3][YaleBlue]{\href{#2}{\color{#1}{#3}}}%

\tcbset{
  aibox/.style={
    width=\linewidth,
    top=8pt,
    bottom=4pt,
    colback=blue!6!white,
    colframe=black,
    colbacktitle=black,
    enhanced,
    center,
    attach boxed title to top left={yshift=-0.1in,xshift=0.15in},
    boxed title style={boxrule=0pt,colframe=white,},
  }
}

\newtcolorbox{AIbox}[2][]{aibox,title=#2,#1}

\tcbset{
  takeawaybox/.style={
    width=\linewidth,
    top=8pt,
    bottom=4pt,
    colback=hidden-yellow,
    colframe=black,
    colbacktitle=black,
    enhanced,
    center,
    attach boxed title to top left={yshift=-0.1in,xshift=0.15in},
    boxed title style={boxrule=0pt,colframe=white,},
  }
}

\newtcolorbox{TakeawayBox}[2][]{takeawaybox,title=#2,#1}

\definecolor{mygreen}{RGB}{144, 238, 144} 
\definecolor{darkmygreen}{RGB}{0, 100, 0} 

\usepackage{latexsym}

\definecolor{box-pink}{RGB}{255,72,162}
\definecolor{box-cyan}{RGB}{29,234,221}
\definecolor{box-red}{RGB}{255,1,25}
\definecolor{box-purple}{RGB}{61,31,255}
\definecolor{box-green}{RGB}{71,212,90}
\definecolor{box-yellow}{RGB}{255,146,1}
\definecolor{box-blue}{RGB}{5,188,248}

\definecolor{tree-pink}{RGB}{255,232,242}
\definecolor{tree-cyan}{RGB}{199,241,240}
\definecolor{tree-red}{RGB}{255,215,214}
\definecolor{tree-purple}{RGB}{230,231,255}
\definecolor{tree-green}{RGB}{192,242,213}
\definecolor{tree-yellow}{RGB}{255,242,230}
\definecolor{tree-blue}{RGB}{223,244,255}

\title{Speed Always Wins: A Survey on Efficient Architectures for Large Language Models}

\runningtitle{Speed Always Wins: A Survey on Efficient Architectures for Large Language Models}

\author{Weigao Sun*\textsuperscript{\rm 1}\quad Jiaxi Hu*\textsuperscript{\rm 2 }\quad Yucheng Zhou*\textsuperscript{\rm 3}\quad Jusen Du*\textsuperscript{\rm 1}\quad Disen Lan*\textsuperscript{\rm 1}\quad Kexin Wang*\textsuperscript{\rm 4}\\ \textbf{Tong Zhu}*\textsuperscript{\rm 5}\quad \textbf{Xiaoye Qu}*\textsuperscript{\rm 1}\quad \textbf{Yu Zhang}\textsuperscript{\rm 5}\quad \textbf{Xiaoyu Mo}\textsuperscript{\rm 6}\quad \textbf{Daizong Liu}\textsuperscript{\rm 7}\quad \textbf{Yuxuan Liang}\textsuperscript{\rm 2}\\  \textbf{Wenliang Chen}\textsuperscript{\rm 5}\quad \textbf{Guoqi Li}\textsuperscript{\rm 4}\quad \textbf{Yu Cheng}\textsuperscript{\rm 8 \Letter}\\
\vspace{1mm}
\textsuperscript{\rm 1}Shanghai AI Laboratory \quad
\textsuperscript{\rm 2}HKUST (GZ) \quad
\textsuperscript{\rm 3}University of Macau \\
\textsuperscript{\rm 4}Institute of Automation, Chinese Academy of Sciences \quad
\textsuperscript{\rm 5}Soochow University  \\
\textsuperscript{\rm 6}KTH Royal Institute of Technology \quad
\textsuperscript{\rm 7}Peking University \quad
\textsuperscript{\rm 8}The Chinese University of Hong Kong \\
\vspace{-1mm}
}

\begin{document}

\begingroup
\renewcommand{\thefootnote}{} 
\footnotetext{*Equal Contribution \quad \textsuperscript{\rm \Letter}Corresponding Author (chengyu@cse.cuhk.edu.hk)}
\endgroup

\begin{abstract}
Large Language Models (LLMs) have delivered impressive results in language understanding, generation, reasoning, and pushes the ability boundary of multimodal models.
Transformer models, as the foundation of modern LLMs, offer a strong baseline with excellent scaling properties.
However, the traditional transformer architecture requires substantial computations and poses significant obstacles for large-scale training and practical deployment.
In this survey, we offer a systematic examination of innovative LLM architectures that address the inherent limitations of transformers and boost the efficiency.
Starting from language modeling, this survey covers the background and technical details of linear and sparse sequence modeling methods, efficient full attention variants, sparse mixture-of-experts, hybrid model architectures incorporating the above techniques, and emerging diffusion LLMs.
Additionally, we discuss applications of these techniques to other modalities and consider their wider implications for developing scalable, resource-aware foundation models.
By grouping recent studies into the above category, this survey presents a blueprint of modern efficient LLM architectures, and we hope this could help motivate future research toward more efficient, versatile AI systems.

\vspace{2mm}
\centering
\github{} \textbf{GitHub}: \MYhref{https://github.com/weigao266/Awesome-Efficient-Arch}{https://github.com/weigao266/Awesome-Efficient-Arch}

\end{abstract}

\maketitle

\begin{figure}[ht]
    \centering
    \vspace{-3mm}
    \includegraphics[width=0.92\linewidth]{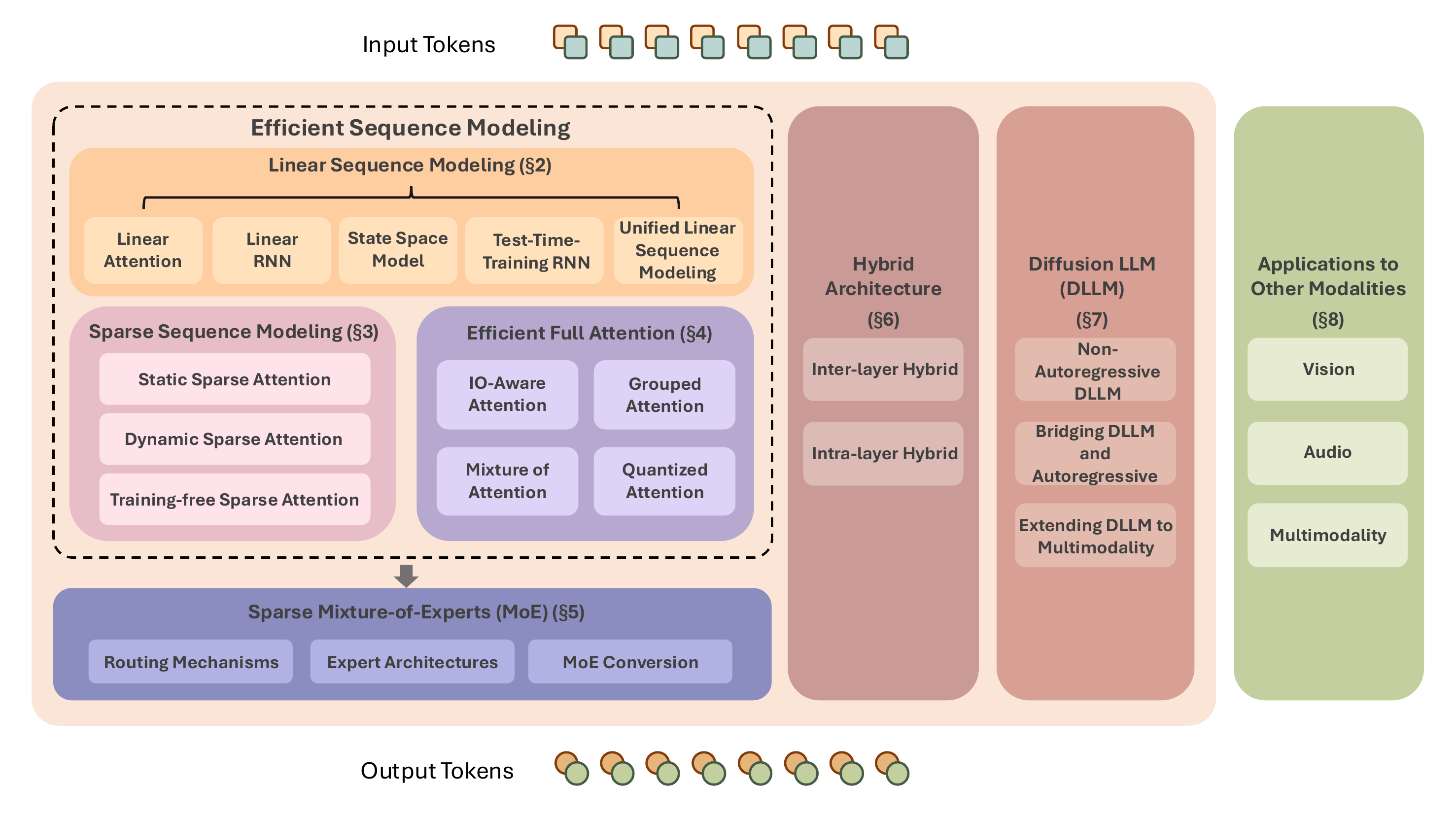}
    \vspace{-3mm}
    \caption{{Overview of Efficient Architectures for Large Language Models.}}
    \label{fig:fig1}
\end{figure}

\onecolumn
\clearpage
\addtocontents{toc}{\protect\setcounter{tocdepth}{2}}
\vskip 3mm
\startcontents[sections]\vbox{\sc Table of Contents}
\vspace{4mm}
\hrule height .8pt
\vspace{-2mm}

{\setlength{\baselineskip}{8pt} 
\setlength{\parskip}{3pt}
\printcontents[sections]{l}{1}{\setcounter{tocdepth}{2}}}

\vspace{4mm}
\hrule height .8pt
\vskip 10mm
\clearpage

\section{Introduction}
\label{sec:intro}

\subsection{Background}

In recent years, Large Language Models (LLMs) have emerged extraordinary capabilities in understanding and generating natural language have driven substantial progress across a wide range of tasks, including text generation~\cite{zhao2023survey, brown2020language, raffel2020exploring}, code generation~\cite{feng2020codebert, li2022competition, fried2022incoder}, question answering~\cite{lewis2020retrieval, guu2020retrieval}, and machine translation~\cite{raffel2020exploring, ranathunga2023neural}. Prominent LLM families such as ChatGPT~\cite{radford2018improving, radford2019language, brown2020language, achiam2023gpt, hurst2024gpt, jaech2024openai, OpenAI_GPT-o3_SystemCard,gpt-oss,gpt5}, Claude~\cite{claude1,claude21,claude37,claude4,claude41}, Gemini~\cite{team2023gemini, team2024gemini, Gemini2025}, DeepSeek~\cite{bi2024deepseek, liu2024deepseekv2, liu2024deepseekv3, guo2025deepseek}, Qwen~\cite{bai2023qwen, team2024qwen2, Yang2024Qwen25TR, yang2025qwen3}, LLaMA~\cite{touvron2023llama, touvron2023llama2, grattafiori2024llama, meta2025llama}, GLM~\cite{glm2024chatglm}, Minimax-Text~\cite{li2025minimax}, InternLM~\cite{team2023internlm,cai2024internlm2}, Hunyuan~\cite{sun2024hunyuan,liu2025hunyuan} have continuously pushed the boundaries of performance, while also reshaping how people interact with machines in daily life.
Beyond their initial role in language tasks, LLMs are increasingly being applied in two demanding areas: multimodality and complex reasoning. In multimodal applications, LLMs now play a key role in systems that integrate and generate information across multiple data types. Recent advances in Vision-Language Models (VLMs), such as Qwen-VL~\cite{wang2024qwen2,bai2025qwen2,xu2025qwen2}, InternVL~\cite{chen2024internvl,chen2024far,chen2024expanding,zhu2025internvl3}, Seed-VL~\cite{guo2025seed1}, Kimi-VL~\cite{team2025kimivl}, Minimax-VL~\cite{li2025minimax},  illustrate this shift, showcasing enhanced abilities in handling cross-modal understanding and generation by combining language skills with visual processing. 
At the same time, a growing line of work focuses on strengthening the reasoning capabilities of LLMs, often referred to as Large Reasoning Models (LRMs). Representative systems like OpenAI o1/o3~\cite{jaech2024openai, OpenAI_GPT-o3_SystemCard}, DeepSeek-R1~\cite{guo2025deepseek}, Seed1.5-Thinking~\cite{seed2025seed1}, Minimax-M1~\cite{chen2025minimax}, Kimi-k1.5/K2~\cite{team2025kimi,kimiteam2025kimik2openagentic} incorporate strategies such as long-chain Chain-of-Thought (CoT) prompting~\cite{wei2022chain} and Reinforcement Learning (RL)~\cite{qu2025survey} to support multi-step reasoning and more deliberate cognitive behavior.

Although LLMs, VLMs, and LRMs have brought major advances in language understanding, multimodal processing, and complex reasoning, they also introduce considerable computational demands~\cite{cottier2024rising, bogmans2025power, faiz2023llmcarbon}. These increased requirements result in significantly higher development and deployment costs, which present practical barriers to widespread adoption. This challenge is shared across LLMs, VLMs, and LRMs, highlighting a common trade-off between model capability and efficiency. While such models offer a promising path toward intelligence, their high resource consumption raises important questions about the sustainability and practicality of pursuing even more powerful systems under current computational constraints.
This keeps us thinking: \textbf{Have we paused to consider the immense hidden costs behind such unrivaled capabilities, and what is the true price of this "intelligence"?}



The core architecture behind many of latest breakthroughs is the Transformer~\cite{vaswani2017attention}, introduced in 2017. Its self-attention mechanism allows models to capture long-range dependencies more effectively than traditional Recurrent Neural Networks (RNNs)~\cite{greff2016lstm}, enabling the scaling of LLMs to hundreds of billions or even trillions of parameters~\cite{brown2020language}.
However, one major limitation of the Transformer lies in the quadratic complexity of its self-attention mechanism, which scales as $O(N^2)$ with the input sequence length $N$~\cite{katharopoulos2020transformers}. This computational inefficiency leads to extremely high training and inference costs, particularly in tasks that involve long-context inputs~\cite{sun2025linear}.
With the continued advancement of artificial intelligence (AI), long-sequence scenarios are becoming increasingly popular. 

\begin{figure*}[t]
    \centering
    \includegraphics[width=\linewidth]{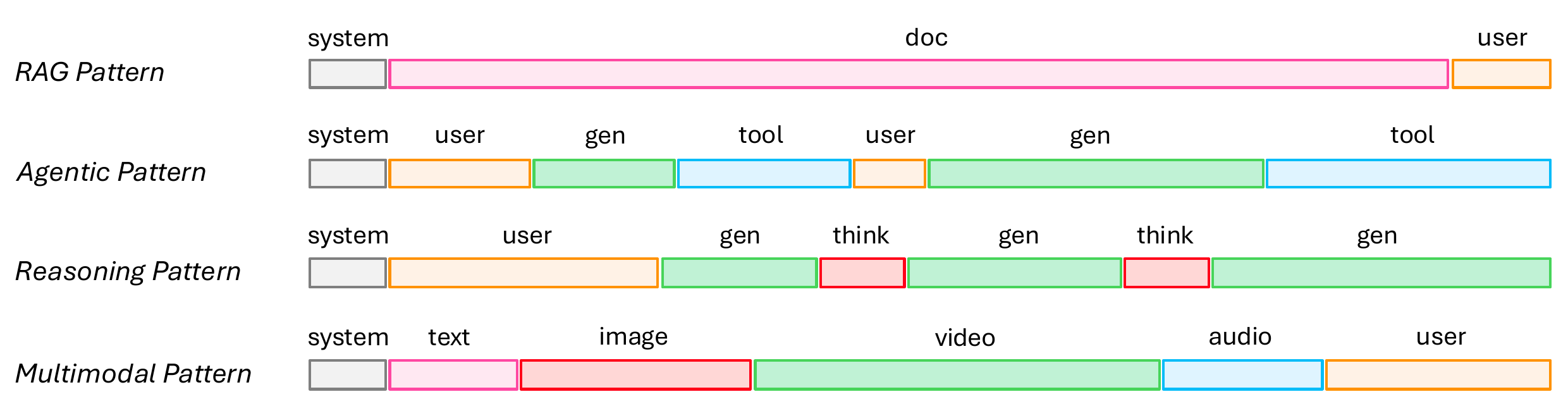}
    \caption{{Long Context Patterns.} We provide representative examples of long-context usage patterns across various scenarios, including retrieval-augmented generation (RAG), agentic, reasoning, and multimodal applications.}
    \label{fig:long_context}
\end{figure*}

As shown in Figure~\ref{fig:long_context}, tasks such as Retrieval-Augmented Generation (RAG)~\cite{lewis2020retrieval} often requires LLMs to process entire documents. In the emerging era of AI agents~\cite{qin2025ui}, long sequences frequently arise from repeated generations and multiple tool invocations. When models are equipped with enhanced reasoning abilities~\cite{qu2025survey}, forming LRMs, they must handle lengthy chains of thought, which also result in long sequences. Similarly, in multimodal applications\cite{yin2024survey}, high-resolution images, videos, and audio introduce additional long-sequence challenges.
Another key component of the Transformer architecture, the Feed-Forward Network (FFN)~\cite{masoudnia2014mixture}, also faces challenges as model size increases. When the number of parameters grows beyond a certain scale, the training cost and inference efficiency of the FFN layer become increasingly difficult to manage.
In this condition, another question arises: \textbf{How can we break through the Transformer's efficiency ceiling? Is costly "intelligence" our only path forward?}

\tikzstyle{my-box}=[
rectangle,
draw=black,
rounded corners,
text opacity=1,
minimum height=1.5em,
minimum width=5em,
inner sep=2pt,
align=left,
fill opacity=.5,
]

\tikzstyle{linear-seq}=[my-box, fill=tree-pink]
\tikzstyle{sparse-seq}=[my-box, fill=tree-cyan]
\tikzstyle{softmax-attn}=[my-box, fill=tree-red]
\tikzstyle{moe}=[my-box, fill=tree-purple]
\tikzstyle{hybrid}=[my-box, fill=tree-green]
\tikzstyle{diffusion}=[my-box, fill=tree-yellow]
\tikzstyle{applications}=[my-box, fill=tree-blue]

\tikzstyle{leaf}=[
my-box, 
minimum height=1.5em,
fill=tree-pink,
text=black,
align=left,
font=\normalsize,
inner xsep=5pt,
inner ysep=4pt,
align=left,
text width=45em,
]
\tikzstyle{leaf2}=[
my-box, 
minimum height=1.5em,
fill=tree-cyan, 
text=black,
align=left,
font=\normalsize,
inner xsep=5pt,
inner ysep=4pt,
]
\tikzstyle{leaf3}=[
my-box, 
minimum height=1.5em,
fill=tree-red, 
text=black,
align=left,
font=\normalsize,
inner xsep=5pt,
inner ysep=4pt,
]

\begin{figure}[htbp]
\vspace{-2mm}
\centering
\resizebox{0.93\textwidth}{!}{
	\begin{forest}
		forked edges,
		for tree={
			grow=east,
			reversed=true,
			anchor=base west,
			parent anchor=east,
			child anchor=west,
			base=left,
			font=\large,
			rectangle,
			draw=black,
			rounded corners,
			align=left,
			minimum width=4em,
			edge+={darkgray, line width=1pt},
			s sep=8pt,
			inner xsep=2pt,
			inner ysep=4pt,
			line width=1.1pt,
			ver/.style={rotate=90, child anchor=north, parent anchor=south, anchor=center},
		},
		where level=1{text width=10em,font=\normalsize,}{},
        where level=2{text width=12em,font=\normalsize,}{},
        where level=3{text width=9.5em,font=\normalsize,}{},
        where level=4{text width=42em,font=\normalsize,}{},
[\ \ Efficient Architectures, text width=12em, ver
	[\ \ \ \ Linear Sequence \\ \ \ \ \ \ \ Modeling~(\S\ref{sec:linear}), text width=10em, linear-seq, draw=box-pink
            [\ \ \ Linear Attention~(\S\ref{subsec:linear attention}), linear-seq, draw=box-pink
                [
    			\eg
                    Linear Transformer~\citep{katharopoulos2020transformers}{,}
                    ABC~\citep{peng2021abc}{,}
                    Lightning Attention~\citep{qin2024various}{,}
                    GLA~\citep{yang2023gated}{,} 
                    GSA~\citep{zhang2024gated}{,}\\    
                    LightNet~\citep{qin2024you}{,}
                    Based~\citep{arora2024simple}{,}
                    Rebased~\citep{aksenov2024rebased}{,}
                    DeltaNet~\citep{yang2024parallelizing}{,}  
                    Gated DeltaNet~\citep{yang2024gated}{,}
                    MoM~\citep{du2025mom}{,}
                    \textit{etc.},
                    leaf, text width=42em, linear-seq, draw=tree-pink
    		  ]
            ]
            [\ \ \ \ \ \ \ Linear RNN~(\S\ref{subsec:linear_rnn}), linear-seq, draw=box-pink
                [\eg 
    			HGRN~\citep{qin2024hierarchically}{,} 
                HGRN2~\citep{qin2024hgrn2}{,} 
                RWKV4~\citep{peng2023rwkv}{,} 
                RWKV6~\citep{peng2024eagle}{,} 
                RWKV7~\citep{peng2025rwkv}{,} 
                LRU~\citep{orvieto2023resurrecting}{,}\\ 
                xLSTM~\citep{beck2024xlstm}{,}
                GateLoop~\citep{katsch2023gateloop}{,} 
                \textit{etc.},
                leaf, text width=42em, linear-seq, draw=tree-pink
    		  ]
            ]
		[\ \ State Space Model~(\S\ref{subsec:ssm}), linear-seq, draw=box-pink
            [\eg~LegRNN~\citep{voelker2019legendre}{,} Hippo~\citep{gu2020hippo}{,} LSSL~\citep{gu2021combining}{,} S4~\citep{gu2021efficiently}{,} 
            HTTYH~\citep{gu2022howtotrain}{,} 
            DSS~\citep{gupta2022diagonal}{,} 
            S4D~\citep{gu2022parameterization}{,} \\
            H3~\citep{gu2021efficiently}{,}
            S5~\citep{smith2022simplified}{,}
            SpaceTime~\citep{zhang2023effectively}{,}
            Time-SSM~\citep{hu2024time}{,} Stable-SSM~\citep{wang2023stablessm}{,} Hippo-PTD~\citep{yu2023robustifying}{,}\\
            Liquid-S4~\citep{hasani2022liquid}{,} 
            Longhorn~\citep{liu2024longhorn}{,}
            Mamba~\cite{gu2023mamba}{,}
            Mamba2~\citep{dao2024transformers}{,} 
            Comba~\citep{hu2025comba}{,} \textit{etc.}
            , leaf, text width=42em, linear-seq, draw=tree-pink]
		]
		[\ \ \ \ \ \ Test-Time-Training\\ \ \ \ \ \ \ \ \ \ \ \ \ RNN~(\S\ref{subsec:ttt}), linear-seq, draw=box-pink
            [\eg~TTT~\citep{sun2024learning}{,} 
                Titans~\citep{behrouz2024titans}{,} 
                Lattice~\citep{karami2025lattice}{,} 
                Miras~\citep{behrouz2025s}{,} 
                Atlas~\citep{behrouz2025atlas}{,} 
                MesaNet~\citep{von2025mesanet}{,} \textit{etc.}
            , leaf, text width=42em, linear-seq, draw=tree-pink]
		]
        [\ \ Unified Linear Sequence\\ \ \ \ \ \ \ \ \ \ Modeling~(\S\ref{subsec:unified}), linear-seq, draw=box-pink
            [
            \eg
                LCSM~\citep{qin2024unlocking}{,}
                Linear-MoE~\citep{sun2025linear}{,}
                Mamba2~\citep{dao2024transformers}{,} 
                Comba~\citep{hu2025comba}{,}
                TTT~\citep{sun2024learning}{,} 
                Titans~\citep{behrouz2024titans}{,} 
                \textit{etc.},
                leaf, text width=42em, linear-seq, draw=tree-pink
          ]
        ]
        [\ \ \ \ \ Linearization~(\S\ref{subsec:linearization}), linear-seq, draw=box-pink
            [
            \eg
                T2R~\citep{kasai2021finetuning}{,}
                MambaInLlama~\citep{wang2024mamba}{,}
                SUPRA~\citep{mercat2024linearizing}{,}
                LoLCATs~\citep{zhang2024lolcats}{,}
                Liger~\citep{lan2025liger}{,}
                \textit{etc.},
                leaf, text width=42em, linear-seq, draw=tree-pink
          ]
        ]
        [\ \ \ \ \ \ Hardware-efficient\\ \ \ \ Implementation~(\S\ref{subsec:hardware-linear}), linear-seq, draw=box-pink
            [
            \eg
                Lightning Attention~\citep{qin2024various}{,}
                GLA~\citep{yang2023gated}{,} 
                S4~\citep{gu2021efficiently}{,} 
                Mamba~\cite{gu2023mamba}{,}
                Mamba2~\citep{dao2024transformers}{,} 
                \textit{etc.},
                leaf, text width=42em, linear-seq, draw=tree-pink
          ]
        ]
	]
[\ \ \ \ Sparse Sequence \\ \ \ \ \ \ \ Modeling~(\S\ref{sec:sparse}), text width=10em, sparse-seq, draw=box-cyan
    [\ \ \ \ \ \ \ \ \ \ \ \ \ \ Static\\  \ \ \ Sparse Attention~(\S\ref{subsec:static sa}), sparse-seq, draw=box-cyan
        [\eg Sparse Transformer~\citep{child2019generating}{,}
        Star-Transformer~\citep{guo2019star}{,}
        BlockBERT~\citep{qiu2019blockwise}{,}
        Longformer~\citep{beltagy2020longformer}{,}\\
        ETC~\citep{ainslie2020etc}{,}
        BigBird~\citep{zaheer2020big}{,}
        LongT5~\citep{guo2021longt5}{,}
        LongNet~\citep{ding2023longnet}{,}
        Axial Attention~\citep{ho2019axial}{,} \textit{etc.}
        , leaf3, text width=42em, sparse-seq, draw=tree-cyan]
    ]
    [\ \ \ \ \ \ \ \ \ \ \ \ Dynamic\\ \ \ \ Sparse Attention~(\S\ref{subsec:dynamic sa}), sparse-seq, draw=box-cyan
        [\eg Reformer~\citep{kitaev2020reformer}{,}
        Routing Transformer~\citep{roy2021efficient}{,}
        Sparse Sinkhorn Attention~\citep{tay2020sparse}{,}\\
        Memorizing Transformers~\citep{wu2022memorizing}{,}
        Unlimiformer~\citep{bertsch2024unlimiformer}{,}
        NSA~\citep{yuan2025native}{,}
        MoSA~\citep{pikekos2025mixture}{,} \textit{etc.}
        , leaf3, text width=42em, sparse-seq, draw=tree-cyan]
        ]
    [\ \ \ \ \ \ \ \ \ \ Training-free\\ \ \ \ Sparse Attention~(\S\ref{subsec: training-free sa}), sparse-seq, draw=box-cyan
        [\eg SpAtten~\citep{wang2021spatten}{,}
        {MInference}~\citep{jiang2024minference}{,}
        {SeerAttention}~\citep{gao2024seerattention}{,}
        {StreamingLLM}~\citep{xiao2023efficient}{,}
        {H2O}~\citep{zhang2024h2o}{,}\\
        {FastGen}~\citep{ge2023fastgen}{,}
        {Quest}~\citep{tang2024quest}{,}
        {LongHeads}~\citep{lu2024longheads}{,}
        LServe~\cite{yang2025lserve}{,}
        XAttention~\cite{xu2025xattention}{,} \textit{etc.}
        , leaf3, text width=42em, sparse-seq, draw=tree-cyan]
    ]
    [\ \ \ \ \ \ Hardware-efficient\\ \ \ \ \ Implementation~(\S\ref{subsec:hardware sa}), sparse-seq, draw=box-cyan
        [\eg 
        Longformer~\citep{beltagy2020longformer}{,}
        FlashAttention-1~\citep{dao2022flashattention}{,} 
        FlashAttention-2~\citep{dao2023flashattention}{,} 
        SeerAttention~\citep{gao2024seerattention}{,} \\
        NSA~\citep{yuan2025native}{,}
        MoBA~\citep{lu2025moba}{,}
        \textit{etc.}
        , leaf3, text width=42em, sparse-seq, draw=tree-cyan]
    ]
]
	[\ \ \ \ \ \ \ Efficient Full \\ \ \ \ \ \ \ Attention~(\S\ref{sec:softmax}), softmax-attn, draw=box-red
    [\ \ \ \ \ \ IO-Aware Attention\\ \ \ \ \ \ \ \ \ \ \ \ \ \ \ (\S\ref{subsec:io-aware attn}), softmax-attn, draw=box-red
			[\eg 
			FlashAttention-1~\cite{dao2022flashattention}{,} FlashAttention-2~\cite{dao2023flashattention}{,}FlashAttention-3~\cite{shah2024flashattention} \textit{etc.}, leaf2, text width=42em, softmax-attn, draw=tree-red]
	]
	[\ \ \ \ \ \ \ Grouped Attention\\ \ \ \ \ \ \ \ \ \ \ \ \ \ \ (\S\ref{subsec:grouped attn}), softmax-attn, draw=box-red
			[\eg MQA~\cite{shazeer2019fast}{,}
            GQA~\cite{ainslie2023gqa}{,}
            MLA ~\cite{liu2024deepseekv2,liu2024deepseekv3}{,}
            GTA~\cite{zadouri2025hardware}{,} 
            GLA~\cite{zadouri2025hardware} \textit{etc.}
			, leaf2, text width=42em, softmax-attn, draw=tree-red]
	]
	[\ \ \ \ \ Mixture of Attention\\ \ \ \ \ \ \ \ \ \ \ \ \ \ \ (\S\ref{subsec:mixture of attn}), softmax-attn, draw=box-red
			[\eg MoA~\cite{fu2024moa}{,} Llama-MoE-v2~\cite{Qu2024LLaMAMoEVE}{,} MoH~\cite{jin2024moh}{,} MoBA~\cite{lu2025moba}{,} MoM~\cite{du2025mom}{,} MoSA~\cite{pikekos2025mixture}{,} \textit{etc.}
			, leaf2, text width=42em, softmax-attn, draw=tree-red]
	]
	[\ \ \ \ \ Quantized Attention\\ \ \ \ \ \ \ \ \ \ \ \ \ \ \ (\S\ref{subsec:quantized attn}), softmax-attn, draw=box-red
		[\eg SageAttention-V1~\citep{zhang2024sageattention}{,} SageAttention-V2~\citep{zhang2024sageattention2}{,} SageAttention-V3~\citep{zhang2025sageattention3}{,} Q-Bert~\citep{shen2020q}{,}\\ I-BERT~\citep{kim2021bert}{,}
        INT-FlashAttention~\citep{chen2024int}{,} Q8BERT~\citep{zafrir2019q8bert}{,} FullyQT~\citep{prato2019fully}{,} TurboAttention~\citep{kang2024turboattention}{,}\\ HACK~\citep{zhang2025hack}{,} BitDistiller~\citep{du2024bitdistiller}{,} \textit{etc.}
			, leaf2, text width=42em, softmax-attn, draw=tree-red]
    ]
]
	[\ \ \ \ Sparse Mixture- \\ \ \ \ \ \ of-Experts~(\S\ref{sec:moe}), moe, draw=box-purple
    [\ \ \ \ \ Routing Mechanisms\\ \ \ \ \ \ \ \ \ \ \ \ \ \ \ (\S\ref{subsec: routing}), moe, draw=box-purple
        [\eg 
        Expert-Choice~\cite{zhou2022mixture}{,}
        BASE Layer~\cite{Lewis2021BASELS}{,}
        Hash Layer~\cite{Roller2021HashLF}{,}
        MoE-Dynamic~\cite{Huang2024HarderTN}{,} \\
        DynMoE~\cite{Guo2024DynamicMO}{,} AdaMoE~\cite{Zeng2024AdaMoETR}{,}
        Ada-K~\cite{Yue2024AdaKRB}{,}
        AuxLossFree~\cite{Wang2024AuxiliaryLossFreeLB}{,}
        Global Batch~\cite{Qiu2025DemonsIT}{,}
        \textit{etc.}, leaf2, text width=42em, moe, draw=tree-purple]
	]
	[\ \ \ \ \ Expert Architectures\\ \ \ \ \ \ \ \ \ \ \ \ \ \ \ (\S\ref{subsec: expert}), moe, draw=box-purple
        [\eg 
        DeepSeekMoE~\cite{dai2024deepseekmoe}{,}
        DeepSpeed-MoE~\cite{Rajbhandari2022DeepSpeedMoEAM}{,}
        Qwen3~\cite{yang2025qwen3}{,}
        OLMoE~\cite{muennighoff2024olmoe}{,}
        MoD~\cite{Raposo2024MixtureofDepthsDA}{,}
        \textit{etc.}, leaf2, text width=42em, moe, draw=tree-purple]
    ]
	[\ \ \ \ \ \ \ MoE Conversion\\ \ \ \ \ \ \ \ \ \ \ \ \ \ \ (\S\ref{subsec:moe_conversion}), moe, draw=box-purple
        [\eg 
            MoEBERT~\cite{Zuo2022MoEBERTFB}{,}
            MoEfication~\cite{Zhang2021MoEficationTF}{,}
            LLaMA-MoE~\cite{zhu-etal-2024-llama}{,}
            LLaMA-MoE-v2~\cite{Qu2024LLaMAMoEVE}{,}\\
            Sparse Upcycling~\cite{Komatsuzaki2022SparseUT}{,}
            BTM~\cite{Li2022BranchTrainMergeEP}{,}
            BTX~\cite{Sukhbaatar2024BranchTrainMiXME}{,}
        \textit{etc.}, leaf2, text width=42em, moe, draw=tree-purple]
    ]
]
	[\ \ \ \ \ \ \ \ \ \ Hybrid\\ \ \ \ \  Architectures~(\S\ref{sec:hybrid}), text width=10em, hybrid, draw=box-green
    [\ \ \ \ \ \ Inter-layer Hybrid\\ \ \ \ \ \ \ \ \ \ \ \ \ \ \ (\S\ref{sec:hybrid_inter}), hybrid, draw=box-green
		[\eg 
			Zamba~\citep{zamba}{,} 
            Zamba2~\citep{glorioso2024zamba2}{,} 
            Samba~\citep{samba}{,} 
            Jamba~\citep{jamba}{,} 
            RWKV-X~\citep{rwkv-x}{,} \\
            Minimax-01~\citep{li2025minimax}{,} 
            Mamba-in-Llama~\citep{wang2024mamba}{,}
            HunYuan-Turbos~\citep{liu2025hunyuan}{,} 
            Zebra-Llama~\citep{yang2025zebra}{,} \\
            YOCO~\citep{sun2024yoco}{,}
            RecurrentGemma~\citep{botev2024recurrentgemma}{,} 
            LaCT~\citep{zhang2025test}{,}
            \textit{etc.},
            leaf3, text width=42em, hybrid, draw=tree-green]
		]
	[\ \ \ \ \ \ Intra-layer Hybrid\\ \ \ \ \ \ \ \ \ \ \ \ \ \ \ (\S\ref{sec:hybrid_intra}), hybrid, draw=box-green
		[\eg 
        Hymba~\citep{hymba}{,}
        TransMamba~\citep{transmamba}{,} 
        Liger~\citep{lan2025liger}{,}
        LoLCATs~\citep{zhang2024lolcats}{,} 
        LoLA~\citep{mcdermott2025lola}{,}
        \textit{etc.}, leaf3, text width=42em, hybrid, draw=tree-green]
		]
]
	[\ \ \ Diffusion LLM~(\S\ref{sec:diffusion}), diffusion, draw=box-yellow
    [\ \ \ \ \ \ Non-Autoregressive\\  \ \ \ \ \ Diffusion LLM (\S\ref{subsec:diff_non-auto}), diffusion, draw=box-yellow
        [\eg LLaDA~\citep{nie2025largelanguagediffusionmodels}{,}
        Diffusion-LM~\citep{li2022diffusionlmimprovescontrollabletext}{,}
        DiffuSeq~\citep{gong2023diffuseqsequencesequencetext}{,}
        SEDD~\citep{lou2024discretediffusionmodelingestimating}{,}
        Plaid~\citep{gulrajani2023likelihoodbaseddiffusionlanguagemodels}{,}
        \textit{etc.}
        , leaf2, text width=42em, diffusion, draw=tree-yellow]
	]
	[\ \ \ Bridging Diffusion LLM \\ \ and Autoregressive (\S\ref{subsec:diff_semi-auto}), diffusion, draw=box-yellow
        [\eg  BD3-LMs~\citep{arriola2025blockdiffusioninterpolatingautoregressive}{,}
        Scaling diffusion~\citep{gong2025scalingdiffusionlanguagemodels}{,}
        \textit{etc.} 
        , leaf2, text width=42em, diffusion, draw=tree-yellow]
    ]
	[\ \ Extending Diffusion LLM \\ \ \ \ to Multimodality (\S\ref{subsec:dllm multimodal}), diffusion, draw=box-yellow
        [\eg  LLaDA-V ~\citep{you2025lladavlargelanguagediffusion}{,}
        UniDisc~\citep{swerdlow2025unified}{,}
        LaViDa~\citep{li2025lavida}{,}
        MMaDA~\citep{yang2025mmadamultimodallargediffusion}{,}
        \textit{etc.} 
        , leaf2, text width=42em, diffusion, draw=tree-yellow]
    ]
]
	[\ Applications to Other \\ \ \ \ \ \ \ Modalities~(\S\ref{sec:application}), applications, draw=box-blue
    [\ \ \ \ \ \ \ \ \ \ Vision (\S\ref{subsec:vision}), applications, draw=box-blue
        [\eg 
        Vig~\cite{liao2025vig}{,} 
        Vision-rwkv~\cite{duan2024vision}{,} 
        Tutel~\cite{hwang2023tutel}{,} 
        InsectMamba~\cite{wang2025insectmamba}{,} 
        Voxel mamba~\cite{zhang2024voxel}{,} \\
        DiM~\cite{mo2024scaling}{,} 
        U-mamba~\cite{ma2024u}{,}
        Vm-unet~\cite{ruan2024vm}{,}
        Rwkv-unet~\cite{jiang2025rwkv}{,}
        Mambabev~\cite{you2024mambabev}{,} 
        \textit{etc.}
        , leaf2, text width=42em, applications, draw=tree-blue]
	]
	[\ \ \ \ \ \ \ \ \ \ Audio (\S\ref{subsec:audio}), applications, draw=box-blue
		[\eg 
        Audio mamba~\cite{lin2024audio}{,} 
        Mamca~\cite{zhang2024MAMC}{,} 
        Rawbmamba~\cite{chen2024rawbmamba}{,}
        SaShiMi~\cite{goel2022its_raw}{,} 
        Music-Diff~\cite{liu2024perturbing}{,}\\
        BiMamba~\cite{zhang2024mamba}{,} 
        Dual-path mamba~\cite{jiang2024dual}{,} 
        Spmamba~\cite{li2024spmamba}{,} 
        VAD~\cite{zuo2023advancing}{,} 
        \textit{etc.}
	, leaf2, text width=42em, applications, draw=tree-blue]
    ]
	[\ \ \ \ \ Multimodality (\S\ref{subsec:multimodality}), applications, draw=box-blue
		[\eg 
        MaTAV~\cite{li2024mamba}{,} 
        Avs-mamba~\cite{gong2025avs}{,}
        VisualRWKV-UHD~\cite{li2024visualrwkv}{,}
        Llada-v~\cite{you2025lladavlargelanguagediffusion}{,}
        Mmada~\cite{yang2025mmadamultimodallargediffusion}{,}\\
        Fragkiadaki~\cite{swerdlow2025unified}{,}
        VL-MoE~\cite{shen2023scaling}{,}
        Moe-llava~\cite{lin2024moe}{,}
        MoCLE~\cite{gou2023mixture}{,}
        Llava-mole~\cite{chen2024llava}{,}
        \textit{etc.}
	, leaf2, text width=42em, applications, draw=tree-blue]
    ]
]
]
	\end{forest}
}
\caption{{A Comprehensive Taxonomy of Efficient Architectures for Large Language Models.}}
\label{fig:arch-survey}
\end{figure}
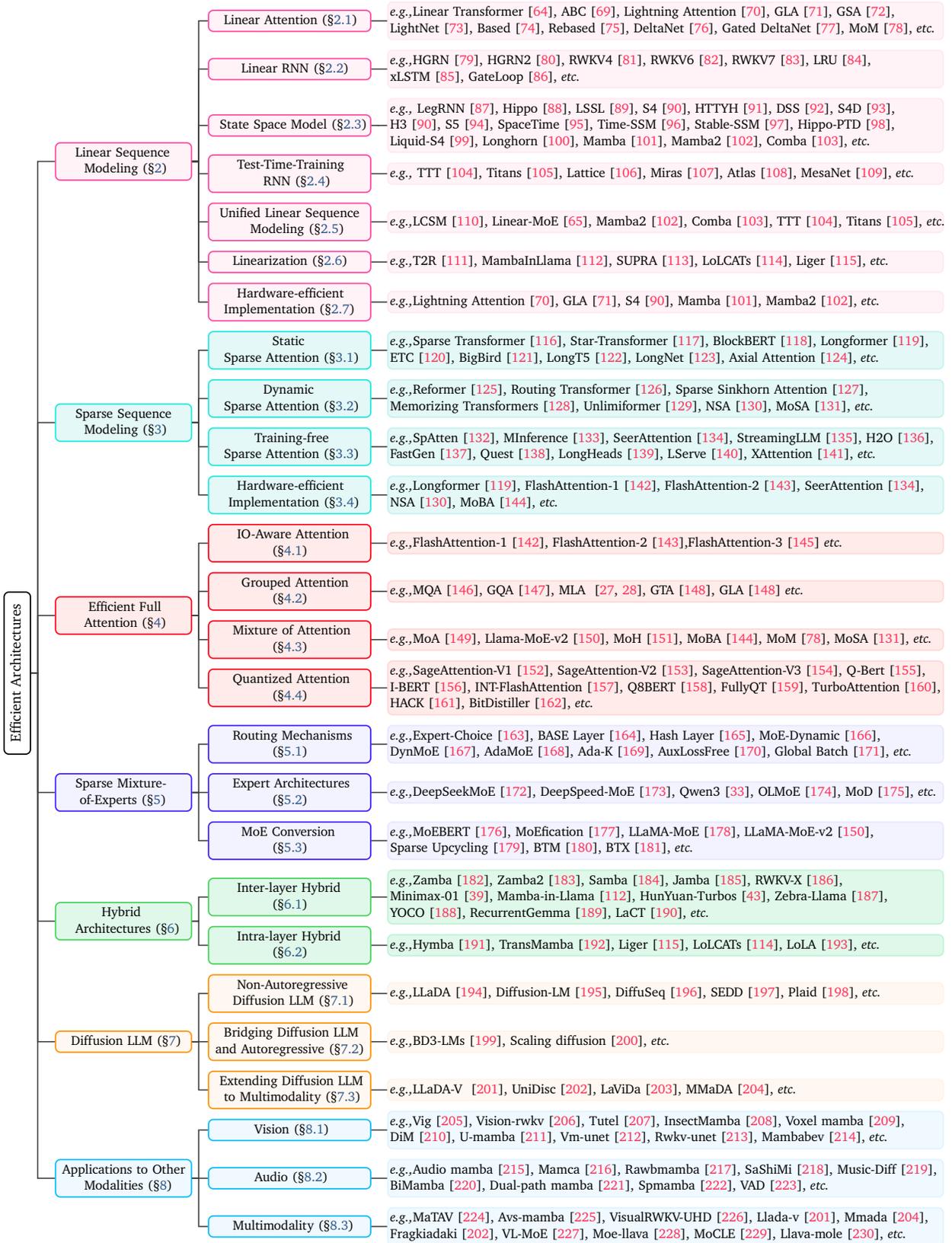

To address these pressing challenges and unlock the full potential of LLMs, the research community has been actively exploring a spectrum of innovative architectural designs and optimization strategies. This survey delves into these innovative approaches, systematically categorizing them to provide a comprehensive overview. The specific methods encompassed within each category can be found in Figure~\ref{fig:arch-survey}. Here we summary each category as below:
\begin{itemize}[leftmargin=*]
    \item \textbf{Linear Sequence Modeling}: These methods aim to reduce the quadratic complexity of self-attention to linear complexity ($O(N)$) by reformulating the attention mechanism, often drawing inspiration from conventional attention, RNNs or state-space models (SSMs). These methods also eliminate the need to store Key-Value (KV) cache during inference, thereby lower the deployment cost.
    \item \textbf{Sparse Sequence Modeling}: Instead of computing attention over all token pairs, these methods selectively focus on a subset of interactions (i.e., the attention map), thereby reducing computational and memory requirements while striving to preserve performance.
    \item \textbf{Efficient Full Attention}:  These methods enhance the standard softmax attention's efficiency while retaining its theoretical quadratic complexity, such as improving memory access efficiency through IO-aware attention mechanisms, and reducing the KV cache size through grouped query mechanisms.
    \item \textbf{Sparse Mixture of Experts}: This paradigm introduces a conditional computation approach where only a subset of a model's parameters (called experts) are activated for each input token, allowing for a massive increase in model capacity without a proportional increase in computational cost.
    \item \textbf{Hybrid Architectures}: These designs strategically combine linear sequence modeling components with traditional full attention layers. This can be achieved through intra-layer hybrid, where both types of operations co-exist within the same layer, or inter-layer hybrid, where different layers utilize distinct attention types, leveraging the strengths of each to trade-off both efficiency and model capacity.
    \item \textbf{Diffusion LLMs}: An emerging area that explores the non-autoregressive diffusion models for language generation, potentially offering new avenues for efficient and high-quality text synthesis.
    \item \textbf{Applications to Other Modalities}: Importantly, the architectural principles driving efficiency in LLMs are not confined to language; their adaptability is increasingly evident in other domains such as vision, audio, and multi-modality, a trend we will also explore.
\end{itemize}

\subsection{Position and Contributions}

The pursuit of more efficient model architectures has attracted substantial attention, resulting in a number of survey articles that chart the evolution of this research area. For example, Tay et al.~\citep{tay2022efficient} provides a detailed examination of Efficient Transformers, discussing a wide range of strategies designed to improve the self‑attention mechanism, including early efforts in linear attention. Recent years, Patro et al.~\citep{patro2024mamba} offers an extensive survey of SSMs, positioning these architectures as promising alternatives to Transformer‑based approaches for handling very long input sequences; their work systematically classifies different SSM designs and highlights applications that span language, vision, and other domains. More recently, Tiezzi et al.~\citep{tiezzi2025back} reviews the renewed interest in recurrent processing, presenting models that blend core ideas from both Transformers and traditional recurrent networks, as well as the latest advances in state‑space formulations.
Sun et al.~\citep{sun2025efficient} summarizes recent advances in linear attention and sparse attention, and hybrid pretrained LLMs with these techniques.
Although these surveys cover some of the efficient architectures like linear models, they only focus on a limit scope of methods to address Transformer's high computation problem.


Compared with the above reviews, this survey starts from key components of the Transformer model and provides a more comprehensive and organized overview of recent advances in efficient architectures of LLMs.
We focus on summarizing their key design principles, performance benefits, known limitations, and possible future developments. Through this synthesis of current progress and emerging trends, we aim to support researchers and practitioners in developing more efficient and scalable LLMs and beyond.

The key contributions of this survey can be summarized as follows:
\begin{itemize}[leftmargin=*]
  \item \textbf{Comprehensive Survey of Efficient Sequence Modeling:}  
    We present an in‑depth review of recent progress in efficient sequence modeling. This includes methods that reduce the quadratic cost of attention through linear sequence models, strategies that selectively sparsify attention scores, and approaches that optimize the full softmax attention operation. By examining their design principles, performance trade‑offs, and implementation details, we identify common themes and trace each technique back to its original motivation.

  \item \textbf{Broader Transformer Component Optimizations:}  
    In addition to sequence modeling, we extend our analysis to other critical Transformer sub‑modules. We cover advances in sparse MoE layers that enable conditional computation, hybrid architectures that blend linear and standard attention within or across layers, and the emerging area of diffusion‑based language generation architectures. This wider perspective highlights how different parts of the Transformer can be re‑imagined for efficiency.

  \item \textbf{Multi-Modal and Multi‑Domain Applications:}  
    Recognizing that efficient architectures are not limited to text processing, we explore their adaptation to other data modalities. We discuss applications in vision, audio, and multimodal settings, demonstrating how efficiency gains in token and channel mixing can benefit tasks ranging from high‑resolution image understanding to multimodal sequence modeling.

\end{itemize}

\section{Linear Sequence Modeling}
\label{sec:linear}

\begin{figure}[t]
    \centering
    \includegraphics[width=\linewidth]{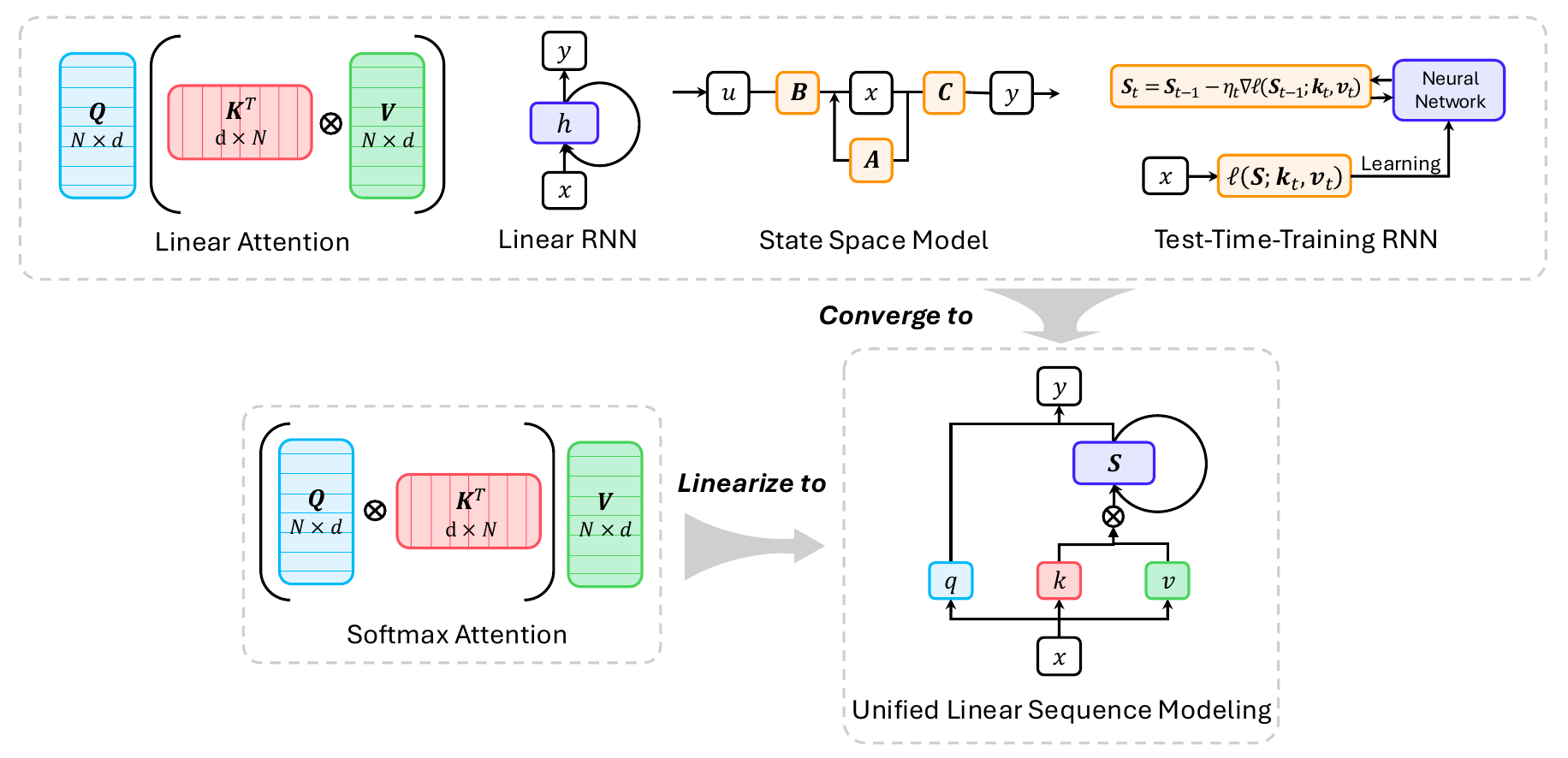}
    \caption{{Linear Sequence Modeling Methods and Their Connections.} The formulations of linear attention, linear RNNs, state space models, and test-time training RNNs have gradually converged toward a unified representation. Moreover, softmax attention can also be transformed into a linear sequence modeling form through the linearization techniques.
}
    \label{fig:linear}
\end{figure}

In general, linear sequence modeling approaches can be broadly categorized into four groups: Linear Attention, Linear Recurrent Neural Network (Linear RNN), State Space Model (SSM), and Test-Time-Training (TTT) RNN, each originating from distinct motivations and mathematical formulations~\cite{sun2025linear}. Following we first present these methods individually, and then describe how they can be unified under a common linear sequence modeling framework. We further explore a new direction called Linearization, which aims to convert standard Transformer models with softmax attention into linear sequence modeling architectures, allowing them to benefit from the efficiency of linear methods at a low cost~\cite{lan2025liger}. In addition, this section also covers implementation strategies that enhance the hardware efficiency of these linear sequence modeling approaches.

\subsection{Linear Attention}
\label{subsec:linear attention}
Standard transformer \cite{vaswani2017attention} employs softmax attention mechanism which takes the input token $\bm{x}_t$ and computes the output $\bm{o}_t$ through:

\begin{equation}
\label{eq:exp}
\begin{aligned}
    \bm q_t, \bm k_t, \bm v_t = \bm x_t \bm W_Q, \bm x_t \bm W_K, \bm x_t \bm W_V,\quad 
    \bm{o}_t = \frac{\sum_{i=1}^{t}\exp(\bm q_t \bm k_i^\top) \bm v_i}{\sum_{i=1}^{t}\exp(\bm q_t \bm k_i^\top)}
\end{aligned}
\end{equation}

Linear Attention was initially proposed in the Linear Transformer~\cite{katharopoulos2020transformers}, which replaces the standard $\operatorname{softmax}$ attention with a linear approximation based on feature maps or kernel functions. This modification addresses the computational inefficiencies of conventional attention mechanisms and enables linear-time complexity. The generalized form of attention using an arbitrary similarity function can be expressed as:

\begin{equation}
\label{eq:sim}
\begin{aligned}
    \bm{o}_t = \frac{\sum_{i=1}^t \operatorname{sim}(\bm{q}_t,  \bm{k}_i)\bm{v}_i}{\sum_{i=1}^t \operatorname{sim}(\bm{q}_t, \bm{k}_i)}
\end{aligned}
\end{equation}
Note that Eq. (\ref{eq:sim}) is equivalent to Eq. (\ref{eq:exp}) if we substitute the similarity function with the specific implementation $\operatorname{sim}({\bm{q}, \bm{k}}) = \operatorname{exp}(\bm{q}\bm{k}^\top)$. Linear attention introduces kernel-based function to represent $\operatorname{sim}(\bm{q}, \bm{k})$ as $\phi(\bm{q}) \phi(\bm{k})^\top$ with feature mapping $\phi(\cdot)$. Then Eq. (\ref{eq:sim}) can be rewritten as:

\begin{equation}
\label{eq:linear}
\begin{aligned}
    \bm{o}_t = \frac{\sum_{i=1}^t \phi (\bm{q}_t) \phi (\bm{k}_i) \bm{v}_i}{\sum_{i=1}^t \phi (\bm{q}_t) \phi (\bm{k}_i)}
\end{aligned}
\end{equation}
Eq. \ref{eq:linear} can be further simplified by using the associative property of matrix multiplication as:

\begin{equation}
\label{eq:linear2}
\begin{aligned}
    \bm{o}_t = \frac{\phi (\bm{q}_t) \sum_{i=1}^t \phi (\bm{k}_i) \bm{v}_i}{\phi (\bm{q}_t) \sum_{i=1}^t \phi (\bm{k}_i)}
\end{aligned}
\end{equation}

Let $\bm{S}_t = \sum_{i=1}^t \phi(\bm{k}_i^\top)\bm{v}_i$ and $\bm{z}_t =\sum_{i=1}^t \phi(\bm{k}_t)$, the above formulation in Eq. (\ref{eq:linear2}) can be written in a recurrent form \cite{katharopoulos2020transformers} as:

\begin{equation}
    \left\{
    \begin{aligned}
        \bm{S}_t &= \bm{S}_{t-1} + \phi(\bm{k}_t)^\top \bm{v}_t, \\
        \bm{z}_t &= \bm{z}_{t-1} + \phi(\bm{k}_t)^\top,
    \end{aligned}
    \right.
    \ \ 
    \bm{o}_t = \frac{\phi(\bm{q}_t)\bm{S}_t}{\phi(\bm{q}_t)\bm{z}_t}
\end{equation}
By reconfiguring the order of operations $(\bm{Q} \bm{K}^\top) \bm{V}$ into $\bm{Q} (\bm{K}^\top \bm{V})$ using the associative property of matrix multiplication can reduce both computational and memory complexity from quadratic complexity $\mathcal{O}(N^2d)$ to linear complexity $\mathcal{O}(Nd^2)$ over sequence length $N$. 

\textbf{Softmax Approximation via Feature Mapping.} The pursuit of linear computational complexity in linear attention mechanisms fundamentally relies on decoupling attention weight computation from the sequential dependencies inherent in standard softmax attention. This is achieved by reformulating $\operatorname{softmax}(\bm{Q}\bm{K}^\top)\bm{V}$ into $\phi(\bm{Q})(\phi(\bm{K}^\top)\bm{V})$ through the associative property of matrix multiplication. Core to this approach is the approximation of exponential similarity $\exp(\bm{q}^\top\bm{k})$ via feature mapping functions $\operatorname{sim}(\bm{q}, \bm{k}) = \phi(\bm{q}) \phi(\bm{k})^\top$. For example, Linear Transformer~\citep{katharopoulos2020transformers} employs a feature map $\phi(x)= \operatorname{elu}(x)+1$ ensuring non-negative similarity while mitigating gradient vanishing for negative inputs compared to $\operatorname{relu(\cdot)}$. Random feature attention (RFA) ~\citep{peng2021random} proposes to utilize random feature projection \cite{rahimi2007random} as feature mapping function $\phi(\cdot)$ for efficient softmax function approximation, outperforming its baseline $\phi_{\operatorname{elu}}(\cdot)$. \text{Nystromformer} ~\citep{xiong2021nystromformer} and Skyformer~\citep{chen2021skyformer} adopt a similar idea to utilize low-rank kernel-based method for softmax attention approximation. Subsequent work by \cite{han2023flatten} points out that the distribution of attention weights in linear transformers is relatively smooth, limiting focus on informative features, therefore proposes focused function $\phi_p$ as query-key feature mapping to improve feature diversity in linear attention weights and a simple rank restoration module by using depthwise convolution (DWC) to the attention matrix, enhancing the expressiveness of linear attention while maintaining low computation complexity. 

While linear attention models offer considerable gains in sequence modeling efficiency, the fidelity of their approximation to standard attention mechanisms remains deficient, especially when applied to long-context recall tasks. Based \cite{arora2024simple} introduces a simple hybrid architecture combining linear attention and sliding window attention, achieving better throughput and recall tradeoff. ReBased \cite{aksenov2024rebased} improves long-context modeling performance by further developing Based architecture using polynomial kernels with learnable parameters and normalization operations. Concurrently, Zhang et al.~\citep{zhang2024hedgehog} propose Hedgehog feature mapping for mimicking low-entropy ("spiky") weights and dot-product monotonicity properties in softmax attention, achieving better recall performance. However, Qin et al.~\citep{qin2022devil} observes that kernel-based linear attention for softmax attention approximation suffer from unbounded gradients, which would cause unstable convergence during training, and proposes output normalization for gradients stability. ReGLA \cite{lu2025regla} addresses this problem by using the normalized exponential feature mapping function, and introduces variance reduction factor for training stability.

\textbf{Gating Mechanism in Linear Attention.} Linear attention mechanisms often exhibit suboptimal sequence modeling capability compared to softmax attention, with empirical observations indicating performance disparities exceeding marginal levels; this limitation is partly attributed to vanilla linear attention's reliance on cumulative memory updating, which induces memory conflicts and hinders effective sequence modeling \cite{yang2023gated}. Furthermore, this cumulative process intrinsically lacks mechanisms for dynamically regulating information flow over sequences. Consequently, the introduction of gating mechanism becomes essential to endow the model with adaptive forgetting and controlled updating capabilities, which  allows the model to selectively attenuate or discard obsolete or irrelevant accumulated states while strategically integrating new inputs, mitigating memory interference and enabling more precise, contextually relevant representation learning essential for complex sequence tasks. Early work like TransNormerLLM \cite{qin2023transnormerllm} utilizes time-invariant memory decay and introduces Lightning Attention \cite{qin2024various} for I/O-aware hardware-efficient optimization, and further introduces Lightning Attention-2 \cite{qin2024lightning} by refining block-wise attention computations, which significantly improves computation efficiency. RetNet \cite{sun2023retentive} adopts a retention mechanism by introducing a \textit{data-independent} decay term into linear attention and gain improvement in sequence modeling. 

While prior approaches utilized \textit{data-independent} gating mechanisms for memory management, the ideal scenario involves models dynamically removing less important key-value associations based on the interaction between new inputs and existing memory content to accommodate new information. Recent advancements in linear attention variants demonstrate significant improvements by incorporating gating or decay factors for forgetting and updating. However, a more effective strategy is the adoption of \textit{data-dependent} gating mechanisms, which dynamically control memory retention and updates through input-driven selection. For instance, Gated Linear Attention (GLA) \cite{yang2023gated} employs a data-dependent gating scheme to enhance the performance of sequence modeling and hardware efficiency, while Gated Slot Attention (GSA) \cite{zhang2024gated} leverages context-aware gated linear attention with bounded-memory control (ABC) \cite{peng2021abc} to improve sequence modeling and long-context recall. To address the inefficiency of linear attention in multi-dimensional sequence modeling, where multiplicative decay requires multiple scans over the input, LightNet \cite{qin2024you} introduces an additive linear recurrence that enables efficient single-pass processing of multi-dimensional data. Theoretical work in MetaLA \cite{chou2024metala} establishes the necessity of time-varying gating for optimal softmax attention approximation within linear attention frameworks. Further refinement is achieved by ReGLA \cite{lu2025regla}, which introduces an additional refining gate to modify the forget gate when activations near saturation, thereby enhancing overall model performance and stability.

\textbf{Delta Learning Rule in Linear Attention}
An effective sequence model should be able to remove less relevant key-value associations to accommodate new information, and this removal should depend on the interaction between incoming inputs and the memory state. From the perspective of fast weight programming, the recurrent hidden state or memory in linear attention mechanisms, such as in GLA, functions as a fast weight matrix that maps the input query $\bm{q}_t$ to the output $\bm{o}_t$. This mapping is updated using a Hessian-like rule \cite{chakraverty2019hebbian}, which imposes limitations on memory capacity. To improve the expressiveness and adaptability of memory, recent work has explored the delta (Widrow-Hoff) learning rule \cite{prados1989neural,widrow1988adaptive}, enabling meta-learning or online adaptation during inference, which updates memory as follows:
$$\bm{S}_t = \bm{S}_{t-1} - \beta_t (\bm{S}_{t-1} \bm{k}_t - \bm{v}_t)\bm{k}_t^\top$$

where the update is driven by the difference of predicted output $\bm{S}_{t-1}\bm{k}_t$ and the target value $\bm{v}_t$. This mechanism enhances memory capacity and gives rise to several delta-rule-based models for linear sequence learning. DeltaNet \cite{yang2024parallelizing} is closely related to the Test-Time Training (TTT) framework \cite{sun2024learning}, where memory is treated as a trainable component and updated through gradient descent steps. To better highlight the distinctions, we introduce the TTT family separately in Section~\ref{subsec:ttt}. To overcome the limitation of DeltaNet, which updates only a single key-value pair at each step, Gated DeltaNet \cite{yang2024gated} introduces a gating mechanism that enables more flexible memory control and allows rapid removal of outdated or irrelevant information. MesaNet \cite{von2025mesanet} further advances this approach by incorporating the Mesa layer \cite{von2023uncovering}, which can dynamically adjust computational cost at inference time. It employs a recursive least squares loss for fast weight updates and can be viewed as a second-order online learner. MesaNet reduces the computational overhead of matrix inversion using gradient conjugation and supports hardware-efficient chunkwise parallelization. Additionally, models such as Comba \cite{hu2025comba} and RWKV7 \cite{peng2025rwkv} also fall under this category. However, we present them separately in the sections on state-space models and linear RNNs, respectively, to highlight their unique structural characteristics and connections to these model families.

\textbf{Log-Linear Memory in Linear Attention.} Although linear attention (or more generally, linear sequence modeling) has achieved performance comparable to or even surpassing that of Transformers in various downstream tasks, it still suffers from an inherent limitation: the existence of only a single, fixed-size state, which constrains its memory capacity. As a result, its performance degrades in long-context retrieval scenarios. A possible compromise is to design a model whose memory grows at a rate that is logarithmic plus linear in the sequence length (another potential solution is to expand multiple memories, as done in MoM \cite{du2025mom}).
Log-Linear Attention \cite{guo2025log} introduce a general framework for linear attention models by replacing the fixed size hidden state in linear attention with a logarithmically growing set of hidden states enables achieving $O(N\log N)$ training and $O(\log N)$ inference complexity, thus balancing the efficiency of linear attention and the expressiveness of softmax attention. PSM \cite{yau2025sequential} and Attraos \cite{hu2024attractor} adopt a similar idea to extend linear recursion to a logarithmic scale through Blelloch's parallel scanning \cite{blelloch1990prefix}.

\subsection{Linear RNN}
\label{subsec:linear_rnn}
RNNs are one of the most common methods for processing and modeling sequence data. At any given time step $t$, an RNN processes an input $\bm{x}_t$ and updates its hidden recurrent state $\bm{h}_t$. This updated hidden state is subsequently used to generate an output $\bm{y}_t$:

\begin{equation}
\label{eq:rnn}
\begin{aligned}
    \bm{h}_t &= \sigma (\bm{W}_{hh}\bm{h}_{t-1} + \bm{W}_{hx} \bm{x}_t + \bm{b}_h),\quad 
    \bm{y}_t = f(\bm{h}_t)
\end{aligned}
\end{equation}
where $\bm{h}_{t-1}$ represents the previous hidden state,  $f(\cdot)$ is the projection function, and $\sigma(\cdot)$ is the activation function. Even over various time steps, an RNN maintains a fixed-size hidden state $\bm{h}_t$ containing historical sequence information. Consequently, the parametrization cost for an RNN remains constant regardless of the increase in time steps, enabling linear-time complexity sequence modeling in a recurrent form.

Traditional RNNs face the problem of inability to conduct parallel training and low efficiency due to its recurrence characteristics, limiting the capability in modeling long-term dependencies and difficulty in scaling up. The main reason is that the update of hidden state involves matrix multiplication and nonlinear activation functions which not only leads to gradient issues but also prevents parallel training. Linear RNNs are proposed to address these issues by removing nonlinearity. In some variants, the recurrent transformation is restricted to element-wise operations for efficiency, while others maintain structured or diagonalized matrices. LRU \cite{orvieto2023resurrecting} leverages complex-valued diagonalization for efficient parallel computation and stable exponential parameterization of its recurrent dynamics to effectively capture long-range dependencies. Typical linear RNN such as Gated Linear Recurrent Unit (GLRU) \cite{chung2014empirical,qin2024hierarchically,qin2024hgrn2} is formulated as follows:

\begin{equation}
\label{eq:linear_rnn}
\begin{aligned}
    \bm{g}_t &= \sigma(\bm{W}_g \bm{x}_t + \bm{b}_g),\quad 
    \bm{i}_t = \tau(\bm{W}_i \bm{x}_t + \bm{b}_i),\quad 
    \bm{o}_t = \sigma(\bm{W}_o \bm{x}_t + \bm{b}_o), \\
    \bm{h}_t &= \bm{g}_t \odot \bm{h}_{t-1} + (1 - \bm{g}_t) \odot \bm{i}_t,\quad 
    \bm{y}_t = \bm{h}_t \odot \bm{o}_t
\end{aligned}
\end{equation}
where $\odot$ denotes element-wise product. Since linear RNN (in Eq. (\ref{eq:linear_rnn})) removes nonlinearity, it enables efficient parallelized training and achieve linear recurrent inference for sequence modeling. 

Recent works have made effort to explore more expressive Linear RNN architecture with efficient recurrence or gating mechanisms. HGRN \cite{qin2024hierarchically} maintains a similar RNN form as in Eq.(~\ref{eq:linear_rnn}), using dot products and accumulation to update memory. Additionally, it employs a novel forget gate mechanism featuring a learnable, layer-wise monotonically increasing lower bound across network layers. This allows lower layers to prioritize local, short-term information by forgetting more rapidly, while upper layers, constrained by a higher forget gate lower bound, effectively retain historical context for long-term dependency modeling. To enhance its expressive power within the linear recurrence framework, HGRN incorporates complex-valued hidden states and recurrence, where the gate's magnitude governs memory retention and its phase can encode relative positional information. RWKV4 \cite{peng2023rwkv} adopts a similar RNN structure but uses a channel-wise exponential decay mechanism inspired by relative position modeling in AFT \cite{zhai2021attention}. Moreover, it employs token shifts to integrate the input tokens. These linear RNN models use $d$-dimensional vectors as memory, which limits their capacity. Research has increasingly shown that expanding memory capacity is crucial for enhancing the performance of linear RNNs. Consequently, subsequent approaches have started using the outer product of two $d$-dimensional vectors to form a $d\times d$ matrix as the memory representation.

HGRN2 \cite{qin2024hgrn2} modifies the hidden state update process to Eq.(~\ref{eq:hgrn2}). By using the outer product with the forget gate, it expands the memory to a $d\times d$ matrix, thereby increasing memory capacity. GateLoop \cite{katsch2023gateloop} replaces their static state transition matrices with diagonal state transitions that dynamically adapt to the input. This approach allows for a matrix form of memory update within its recurrent mechanism, enabling more flexible state evolution. RWKV6 \cite{peng2024eagle} changes the original element-wise vector product to a $\bm{k^T}\bm v$ formulation to achieve a multi-headed matrix-valued memory, thus expanding its capacity. RWKV6 introduces an advancement with its dynamic recurrence. This means the channel-wise decay rates within the linear state update are no longer static but become data-dependent and vary at each timestep, achieved efficiently through Low-Rank Adaptation (LoRA) to augment learned base decay vectors. Additionally, it applies similar data-dependency to its token-shift mechanism, allowing for more flexible and context-aware integration of past and current token information. Similarly, xLSTM \cite{beck2024xlstm} extends traditional LSTMs by introducing variants like mLSTM, which replaces the standard vector cell state with a matrix-valued memory to significantly expand storage capacity.

\begin{equation}
\label{eq:hgrn2}
    \bm{h_t} = \bm{h_{t-1}} \cdot \mathrm{Diag}\{\bm{f_t}\}+\bm{i_t} \otimes (1-\bm{f_t}) \in \mathbb{R}^{d \times d}
\end{equation}

At this stage, the memory structure of both linear RNNs and linear attention mechanisms has converged, utilizing matrix-based memory derived from the outer product of vectors. This allows for the application of Test-Time-Training on the memory as will be discussed in $\S$\ref{subsec:ttt}. RWKV7 \cite{peng2025rwkv} incorporates test-time gradient descent by introducing a "dynamic state evolution" powered by a generalized delta rule for its recurrent state updates. This mechanism allows the model to dynamically adapt its multi-headed matrix-valued state at each time step, effectively performing a form of test-time learning or in-context adaptation. This update rule, while more complex than prior RWKV iterations, endows RWKV7 with enhanced expressivity, further enhancing performance.

The motivation behind linear RNNs is to optimize the hidden state update for parallel training. Their recurrent formulations now closely resemble those of linear attention mechanisms. Although their starting points differ, they have evolved towards structurally similar designs, differing mainly in notation and specific architectural choices.

\subsection{State Space Model}
\label{subsec:ssm}

State Space Model (SSM) \cite{kalman1960new} is a historical mathematical framework used to describe dynamic systems that evolve over time in control systems. In this section, we systematically go through its development and applications on efficient sequence modeling.



\textbf{From Hippo Theory to Continuous-Time SSM.}
Unlike the development path of traditional control theory \citep{glasser1985control}, which progresses from state space models to spectral projections, deep state-space models follow a reverse trajectory. According to Legendre-RNN \citep{voelker2019legendre} and $\mathsf{Hippo}$ \cite{gu2020hippo} theory, given an input function $u(s)$, a set of orthogonal polynomial basis $\phi_n(t,s)$ that $\int_{-\infty}^t \phi_m(t,s)\phi_n(t,s)\mathrm{d}s=\delta_{m,n}$, and an inner product probability measure $\mu(t,s)$. This enables us to project the $u(s)$ onto the polynomial basis along the time dimension:

\begin{equation}
\langle u, \phi_n\rangle_{\mu} = \int_{-\infty}^t u(s)\phi_n(t,s)\omega(t,s)\mathrm{d}s
\label{hippo}
\end{equation}

This process can also be viewed as a spectral transformation $\mathcal{K}u(s)=\int_D K(t, s) u(s) \mathrm{d}s$ with kernel $\bm{K}_n=\phi_n(t,s)\omega(t,s)$ \cite{gu2022howtotrain,hu2024time}. According to Time-SSM \cite{hu2024time}, by adjusting the polynomial basis $\phi_n(t,s)$ and inner product probability measure $\mu(t,s)$, various integral transforms, e.g., Gabor and Laplace transforms, can be realized. Moreover, When $\phi$ is the orthogonal basis in closed-recursive-form (e.g., Legendre, Chebyshev, Laguerre), by collecting all coefficients of order $n$, we can obtain a continuous ODE system. Together with the corresponding polynomial reconstruction process \cite{hu2024time}, this leads to the form of a time-invariant continuous-time SSM formulated in LSSL \cite{gu2021combining} and S4 \cite{gu2021efficiently}:

\begin{equation}
\begin{aligned}
    \bm{x}'(t) &= \bm{A}\bm{x}(t) + \bm{B}\bm{u}(t), \quad
    \bm{y}(t) = \bm{C}\bm{x}(t) + \bm{D}\bm{u}(t)
\end{aligned}
\label{eq:ssm}
\end{equation}
This can also be regarded as parameterized maps that transform the input $u(t)$ into an $ N$-dimensional latent space and project it onto the output $y(t)$. $\bm{D}\bm{u}(t)$ term is typically regarded as a residual connection \cite{he2016deep} and can often be omitted.

\textbf{Discretization.}
Since real-world data is typically in discrete form, it is necessary to convert continuous-time SSM parameters ($\bm{A}\bm{B}$) into their discrete counterparts ($\bm{\overline{A}},\bm{\overline{B}}$) with step $\bm{\Delta}$. Early approaches like S4\cite{gu2021efficiently,gu2021combining} employ Zero-Order Hold (ZOH) discretization (use the general solution for Eq. (\ref{eq:ssm})), while later studies like Mamba \cite{gu2023mamba} found that using hybrid discretization methods (use forward Euler method \cite{gu2020hippo} for $\bm{B}$) could lead to more compact and efficient representations.
\begin{align}
    (\text{ZOH}):\quad\bm{\overline{A}} & = \exp(\bm{\Delta} \bm{A}), \quad  \bm{\overline{B}} = (\Delta \bm{A})^{-1}(\operatorname{exp}(\Delta \bm{A}) - \bm{I})\cdot \Delta \bm{B} \\
    (\text{In Practice}):\quad\bm{\overline{A}} & = \exp(\bm{\Delta} \bm{A}), \quad 
    \bm{\overline{B}} = \bm{\Delta} \bm{B}
\end{align}

\textbf{Diagonal SSMs.}
Early SSMs such as S4 and LSSL leveraged the HiPPO theory for initialization, while in practice, the naive recursive calculation of SSM Kernel can be quite computationally intensive. Ideally, when matrix $\bm{A}$ is diagonal, the computation simplifies to exponentiating the diagonal elements only. S4 \cite{gu2021efficiently} initialize the state matrix as a diagonal plus low-rank structure. Later, DSS \cite{gupta2022diagonal} expressed it as a negative diagonal matrix (since any real skew-symmetric matrix derived from Hippo can be conjugate-diagonalized into a complex matrix, and the process $\bm{A} = \bm{V}^{-1}\bm{\Lambda}\bm{V}$ equal to a coordinate transform \cite{hu2024time}), and in S4D \cite{gu2022parameterization}, this is empirically simplified to a real diagonal matrix, which can be regarded as a rough approximation of the HiPPO-LegS matrix, and S4D introduces a method known as the left half-plane control to ensure that the diagonal elements of matrix $\bm{A}^{(D)}$ remain negative. Subsequently, HTTYH \cite{gu2022howtotrain} provided a theoretical extension to the HiPPO framework to encompass the initialization scheme used in S4D. S5 \cite{smith2022simplified} expands the single-input/output (\textbf{SISO} \cite{gu2021efficiently}) SSM to multi-input/output (\textbf{MIMO}) systems and introduces Blelloch parallel scan \cite{blelloch1990prefix} to faster recurrent computation. SpaceTime \cite{zhang2023effectively} enhances model expressiveness by augmenting a diagonal state matrix with full-rank column vectors, effectively transforming it into a companion matrix. In contrast, H3 \cite{fu2022hungry} adopts a two-layer SSM design: it initializes the state matrix of the first SSM layer as a shift matrix, and computes the diagonal parameters of the second SSM via key-value dot products. This design realizes token shift operations and thereby strengthens the model’s ability to recall relevant information. StableSSM \cite{wang2023stablessm} and Hipo-PTD \cite{yu2023robustifying} further discuss how to carry out more stable initialization. Formally, these models can all be expressed as convolution operations during the recursive computation process:

\begin{equation}
    \overline{\bm{K}}  = (\bm{C}\overline{\bm{B}}, \bm{C}\overline{\bm{A}} \overline{\bm{B}}, \dots, \bm{C}\overline{\bm{A}}^L\overline{\bm{B}}), \quad
    \bm{y} = \bm{x} * \overline{\bm{K}}
\end{equation}

\textbf{Time-variant (Selective) SSM.}
Although initialization based on the HiPPO framework is theoretically well-grounded, it results in system dynamics that remain the same at every time step. This uniformity prevents the model from selectively attending to important information, as attention mechanisms do, thereby limiting its expressive capacity. Based on the original formulation of S4 (with low-rank correction), Liquid-S4 \cite{hasani2022liquid} combines with a liquid time-constant network to obtain a new linear time-varying system. Mamba \cite{gu2023mamba}, on the other hand, abandons HiPPO-based initialization and instead directly derives data-dependent SSM parameter matrices through a linear projection layer. Moreover, it simplifies the computation into element-wise multiplications to accelerate inference. Building upon this, Attraos \cite{hu2024attractor} initializes the diagonal matrix A with all ones values to approximate a Lebesgue measure, and further proposes a multi-resolution SSM based on piecewise orthogonal polynomial projection. Mamba-2 \cite{dao2024transformers} further simplifies the state matrix to a scalar and introduces a hardware-efficient algorithm for blockwise parallel computation. 

Recently, inspired by closed-loop control theory, Comba \cite{hu2025comba} introduced two key improvements to enhance the effectiveness of recurrent sequence modeling. First, it replaces the scalar state transition in Mamba2 with a scalar-plus-low-rank (SPLR) matrix, updated via a delta learning mechanism \cite{prados1989neural}. This matrix-based formulation improves the expressiveness of the transition dynamics and, interpreted as a form of Householder transformation \cite{joffrain2006accumulating, bischof1987wy}, introduces supervised memory management. This mitigates memory conflicts typically found in recurrent models that rely on fixed-size hidden states. Furthermore, the SPLR structure naturally supports negative eigenvalues, which further increases representational capacity. A similar formulation also arises in Longhorn \cite{liu2024longhorn}, which extends Mamba to L2-based objectives. Second, Comba incorporates an output correction mechanism that introduces a learnable scalar to connect queries and keys. From the perspective of neural memory, this design ensures that the value is stored with high fidelity and can be accurately retrieved by the corresponding query. The output correction mechanism plays a central role in enabling this functionality. Moreover, Comba improves computational efficiency by eliminating unnecessary matrix inversions, which allows for the use of a highly optimized Triton kernel, resulting in significantly faster execution.


\subsection{Test-Time-Training RNN}
\label{subsec:ttt}
Although models like Comba \cite{hu2025comba}, (Gated)-DeltaNet \cite{yang2024parallelizing,yang2024gated}, and RWKV7 \cite{peng2025rwkv} can also be formulated in an SGD-like manner for test-time training, a range of models such as TTT \cite{sun2024learning} offer a more explicit formulation. These models treat the model’s state matrices directly as fast-adapting weights, which are updated through a learnable optimizer. From this perspective, the model is no longer restricted to a fixed linear or bilinear kernel but can instead leverage more advanced optimization algorithms to gain stronger expressive power.

In terms of formulation, this class of models abandons closed-form recurrent representations and instead expresses the model directly in an online learning paradigm, similar to early meta-learning \cite{munkhdalai2017neural,munkhdalai2019metalearned} and fast weight programming \cite{irie2021going,schlag2021linear}. For the general form:
\begin{equation}
\bm{S}_t=\alpha_t \bm{S}_{t-1}-\eta_t \nabla_{\bm{S}} \ell\left(\bm{S}_{t-1} ; \bm{k}_t, \bm{v}_t\right)
\end{equation}

Initially, TTT \cite{sun2024learning} still employed an SGD-based optimizer for updates and introduced a deep state based on a two-layer MLP, which has been widely adopted by subsequent models. It is worth noting that this type of model typically applies LayerNorm to the state or its output to stabilize gradients and the distribution of the memory stored in the state. Titans \cite{behrouz2024titans} introduces a first-order momentum term to represent long-term memory and momentary surprise. Lattice \cite{karami2025lattice} introduced a mechanism designed for linear models that can compress information into a limited number of memory slots efficiently and can update exclusively with information that is orthogonal to its current state, ensuring that only new and non-redundant data is written into memory and reducing the memory interference. Miras \cite{behrouz2025s} presents a general framework for explaining the role of standard model architectural choices, including: (1) associative memory architecture, (2) attentional bias objective, (3) retention gate, and (4) memory learning algorithm, and introduces three novel sequence model variants based on the guidance of this framework, outperforming traditional Transformers and other modern linear recurrent models on language modeling and recall intensive tasks. Atlas \cite{behrouz2025atlas} adopts higher-order feature mappings for memory capacity enhancing, and proposes the Omega rule and Muon optimizer \cite{jordan2024muon} for memory updating, which can be demonstrated effective in language modeling downstream tasks. LaCT \cite{zhang2025test} further replaces the loss function with a dot product operation and employs large chunks of gradient descent to improve computational efficiency.

\subsection{Unified Linear Sequence Modeling}
\label{subsec:unified}

While linear attention, linear RNNs, state space models, and test-time-training RNNs have traditionally followed separate lines of development, recent studies~\cite{qin2024unlocking,sun2025linear} have begun to integrate these methods within a unified theoretical framework. In this paper, we present a consolidated view of these approaches, focusing on their memory update rules and optimization strategies.

\begin{table}[htbp]
\centering
\small
\vspace{-1em}
\caption{A comparative overview of various linear sequence modeling approaches in terms of their memory update rules and optimization objectives. For reference, we also include the principles of softmax attention and sliding-window attention mechanisms. Broadly speaking, linear attention, linear RNNs, and state-space models follow a convergent developmental trajectory, evolving through mutual influence. This trajectory reflects a shift from data-independent to data-dependent gating mechanisms, and from L1-based to L2-based optimization objectives. In contrast, the development of TTT models has been primarily guided by advances in modern optimization algorithms.}
\vspace{0.5em}
\renewcommand{\arraystretch}{1.6} 
\begin{adjustbox}{width=1\columnwidth, center}
\renewcommand{\multirowsetup}{\centering}
\setlength{\tabcolsep}{.1pt}
\begin{threeparttable}
\begin{tabular}{c c c c}
\toprule
   \textbf{Type}  & \textbf{Model }  & \textbf{Memorizing with {\color{blue}Gate}} & \textbf{Optimization Objective $\mathcal{L}_\text{min}$} \\
   \midrule
   Softmax Attn & SA \cite{vaswani2017attention} & 
   $\bm{S}_t = \bm{S}_{t-1}.\operatorname{append}(\bm{k}_t, \bm{v}_t)$ &
   $ \sum_{i=1}^t\exp(\bm{q}_t^\intercal\bm{k}_i)\left\|\bm{v}-\bm{v}_i\right\|^2 $ \cite{wang2025test}\\
   & SWA \cite{beltagy2020longformer}& 
   $\bm{S}_t = \bm{S}_{t-1}.\operatorname{append}(\bm{k}_t, \bm{v}_t).\operatorname{drop}(\bm{k}_{t-M}, \bm{v}_{t-M})$ &
   $ \sum_{i=t-M}^t\exp(\bm{q}_t^\intercal\bm{k}_i)\left\|\bm{v}-\bm{v}_i\right\|^2$ \\
   \midrule
   Linear Attn & LA \cite{katharopoulos2020transformers}& 
   $\bm{S}_t = \bm{S}_{t-1}+\bm{v}_t\bm{k}_t^\intercal$ &
   $\left\langle\bm{S}_t \bm{k}_t, \bm{v}_t\right\rangle$\\
   & Lightning Attn \cite{qin2024various}&
   $\bm{S}_t = {\color{blue}\alpha}\bm{S}_{t-1}+\bm{v}_t\bm{k}_t^\intercal$ & 
   $\left\langle\bm{S}_t \bm{k}_t, \bm{v}_t\right\rangle+\frac{{\color{blue}\bm{\alpha}}}{2}\left\|\bm{S}_t\right\|_2^2$\\
   & RetNet \cite{sun2023retentive} &
   $\bm{S}_t = {\color{blue}\alpha}\bm{S}_{t-1}+\bm{v}_t\bm{k}_t^\intercal$ & 
   $\left\langle\bm{S}_t \bm{k}_t, \bm{v}_t\right\rangle+\frac{{\color{blue}\bm{\alpha}}}{2}\left\|\bm{S}_t\right\|_2^2$\\
   & GLA \cite{yang2023gated}& 
   $\bm{S}_t = \bm{S}_{t-1}{\color{blue}\operatorname{diag}(\bm{\alpha}_t)}+\bm{v}_t\bm{k}_t^\intercal$ & 
   $\left\langle\bm{S}_t \bm{k}_t, \bm{v}_t\right\rangle+\frac{{\color{blue}\bm{\alpha}_t}}{2}\left\|\bm{S}_t\right\|_2^2$\\
   & MetaLA \cite{chou2024metala} & 
   $\bm{S}_t = \bm{S}_{t-1}{\color{blue}\operatorname{diag}(\bm{\alpha}_t)}+\bm{v}_t(\bm{1}-{\color{blue}\bm{\alpha}_t})^\intercal$ & 
   $\left\langle\bm{S}_t (\bm{1}-{\color{blue}\bm{\alpha}_t}), \bm{v}_t\right\rangle+\frac{{\color{blue}\operatorname{diag}(\bm{\alpha}_t)}}{2}\left\|\bm{S}_t\right\|_2^2$ \\
   & GSA \cite{zhang2024gated} & 
   $\bm{K}_t = \bm{K}_{t-1}{\color{blue}\operatorname{diag}(\bm{\alpha}_t)}+\bm{k}_t(\bm{1}-{\color{blue}\bm{\alpha}_t})^\intercal\quad (\text{same as}~\bm{v})$ & $\left\langle\bm{K}_t (\bm{1}-{\color{blue}\bm{\alpha}_t}), \bm{k}_t\right\rangle+\frac{{\color{blue}\operatorname{diag}(\bm{\alpha}_t)}}{2}\left\|\bm{K}_t\right\|_2^2$ \\
   & DeltaNet \cite{yang2024parallelizing}&  
   $\bm{S}_t = \bm{S}_{t-1}{\color{blue}\left(\bm{I}-{\color{blue}\beta}_t\bm{k}_t\bm{k}_t^\intercal\right)}+{\color{blue}\beta}_t\bm{v}_t\bm{k}_t^\intercal$& 
   $ {\color{blue}\beta_t}\left\|\bm{v}_t-\bm{S}_t \bm{k}_t\right\|^2 $\\
   & Gated-DeltaNet \cite{yang2024gated} &  
   $\bm{S}_t = \bm{S}_{t-1}{\color{blue}\alpha_t^{\sim 1}\left(\bm{I}-{\color{blue}\beta}_t\bm{k}_t\bm{k}_t^\intercal\right)}+{\color{blue}\beta}_t\bm{v}_t\bm{k}_t^\intercal$& 
   $ {\color{blue}\beta_t}\left\|\frac{1}{{\color{blue}\alpha_t}}\bm{v}_t-{\color{blue}\alpha_t}\bm{S}_t \bm{k}_t\right\|^2+\frac{\color{blue}\alpha_t}{2}\left\|\bm{S}_t\right\|_2^2 $\\
   & Delta-Product \cite{siems2025deltaproduct} &
   $\bm{S}_t = \bm{S}_{t-1}{\color{blue}\prod_{i=1}^n\left(\bm{I}-{\color{blue}\beta}_{ti}\bm{k}_{ti}\bm{k}_{ti}^\intercal\right)}+\sum_{i=1}^n{\color{blue}\prod_{i=1}^n\left(\bm{I}-{\color{blue}\beta}_{ti}\bm{k}_{ti}\bm{k}_{ti}^\intercal\right)\beta_{ti}}\bm{v}_{ti}\bm{k}_{ti}^\intercal$ & 
   $ \sum_{i=1}^n{\color{blue}\beta_{ti}}\left\|\bm{v}_{ti}-\bm{S}_{ti} \bm{k}_{ti}\right\|^2 $\\
   & MesaNet \cite{von2025mesanet} & 
   $\bm{S}_t = {\color{blue}\alpha_t}\bm{S}_{t-1}+{\color{blue}\beta_t}\bm{v}_t\bm{k}_t^\intercal,~\bm{H}_t = {\color{blue}\alpha_t}\bm{H}_{t-1}+{\color{blue}\beta_t}\bm{k}_t\bm{k}_t^\intercal$ &
   $ \sum_{i=1}^t{\color{blue}\beta_{t}}\left\|\bm{v}_{t}-\bm{S}_{t} \bm{k}_{t}\right\|^2 $ \\
   \midrule
   Linear RNN 
   & HGRN2 \cite{qin2024hgrn2}& 
   $\bm{S}_t = \bm{S}_{t-1}{\color{blue}\operatorname{diag}(\bm{\alpha}_t)}+\bm{v}_t{\color{blue}(\bm{1-\alpha}_t})^\intercal$ &   
   $\left\langle\bm{S}_t {\color{blue}(\bm{1-\alpha}_t}), \bm{v}_t\right\rangle+\frac{{\color{blue}\bm{\alpha}_t}}{2}\left\|\bm{S}_t\right\|_2^2$\\
   & RWKV6 \cite{peng2024eagle}& 
   $\bm{S}_t = \bm{S}_{t-1}{\color{blue}\operatorname{diag}(\bm{\alpha}_t)}+\bm{v}_t\bm{k}_t^\intercal$ & 
   $\left\langle\bm{S}_t \bm{k}_t, \bm{v}_t\right\rangle+\frac{{\color{blue}\bm{\alpha}_t}}{2}\left\|\bm{S}_t\right\|_2^2$\\
   & RWKV7 \cite{peng2025rwkv}&  
   $\bm{S}_t = \bm{S}_{t-1}{\color{blue}(\operatorname{diag}(\bm{\alpha}_t)-\beta_t\hat{\bm{k}}_t\hat{\bm{k}}_t^\intercal)}+\bm{v}_t\tilde{\bm{k}}_t^\intercal$& 
   $ {\color{blue}\beta_t}\left\|\frac{1}{{\color{blue}\beta_t}}\bm{v}_t-\bm{S}_t \bm{k}_t\right\|^2+\frac{\color{blue}\alpha_t}{2}\left\|\bm{S}_t\right\|_2^2 $\\
   \midrule
   SSM 
   & S4 \cite{gu2021efficiently} & 
   $\bm{S}_t = \bm{S}_{t-1}{\color{blue}\operatorname{diag}(\bm{\alpha})}+\bm{v}_t\bm{b}^\intercal \in \mathbb{C}$ & 
   $\left\langle\bm{S}_t \bm{b}, \bm{v}_t\right\rangle+\frac{{\color{blue}\bm{\alpha}}}{2}\left\|\bm{S}_t\right\|_2^2$ \\
   & Mamba \cite{gu2023mamba} &
   $\bm{S}_t = \bm{S}_{t-1}{\color{blue}\operatorname{diag}(\bm{\alpha})}+{\color{blue}\beta_t^{\sim 0}}\bm{v}_t\bm{k}_t^\intercal$ & 
   ${\color{blue}\beta_t}\left\langle\bm{S}_t \bm{k}_t, \bm{v}_t\right\rangle+\frac{{\color{blue}\bm{\alpha}_t}}{2}\left\|\bm{S}_t\right\|_2^2$\\
   & Mamba2 \cite{dao2024transformers}&
   $\bm{S}_t = {\color{blue}\alpha_t^{\sim 1}}\bm{S}_{t-1}+{\color{blue}\beta_t^{\sim 0}}\bm{v}_t\bm{k}_t^\intercal$ & 
   ${\color{blue}\beta_t}\left\langle\bm{S}_t \bm{k}_t, \bm{v}_t\right\rangle+\frac{{\color{blue}\alpha_t}}{2}\left\|\bm{S}_t\right\|_2^2$\\
   & Longhorn \cite{liu2024longhorn} & $\bm{S}_t = \bm{S}_{t-1}{\color{blue}\left(\bm{I}
-\frac{\beta_t}{1+\beta_t\bm{k}_t^\intercal\bm{k}_t}\bm{k}_t\bm{k}_t^\intercal\right)}+{\color{blue}\beta_t}\bm{v}_t\bm{k}_t^\intercal$ &
   $ {\color{blue}\beta_t}\left\|\bm{v}_t-\bm{S}_t \bm{k}_t\right\|^2 $ \\
   & Comba \cite{hu2025comba} & 
   $\bm{S}_t = \bm{S}_{t-1}{\color{blue}\left(\alpha_t^{\sim 1}
-\beta_t^{\downarrow}\bm{k}_t\bm{k}_t^\intercal\right)}+{\color{blue}\beta_t^{^{\uparrow}}}\bm{v}_t\bm{k}_t^\intercal$& 
   $ {\color{blue}\beta_t}\left\|\bm{v}_t-\bm{S}_t \bm{k}_t \right\|^2+\frac{\color{blue}\alpha_t}{2}\left\|\bm{S}_t\right\|_2^2 + \left\langle\bm{q}_t, {\color{blue}d}\bm{k}_t\right\rangle $\\
   \midrule
   TTT & TTT-MLP \cite{sun2024learning}&  
   $\bm{S}_t(\cdot)=\bm{S}_{t-B}(\cdot)-\sum_{i=1}^{B}{\color{blue}\beta_i}\nabla_S\mathcal{L}\left(\bm{S}_{t-1},\bm{k}_t,\bm{v}_t \right)$& 
   $ {\color{blue}\beta_i}\left\|\bm{v}_i-\psi(\bm{S}_j(\bm{k}_i))\right\|^2 $\\
   & MIRAS \cite{behrouz2025s} &  
   $\bm{S}_t={\color{blue}\alpha_t}\bm{S}_{t-1}-{\color{blue}\beta_t} \nabla_S\mathcal{L}\left(g,\bm{M}_{t-1},\bm{k}_t,\bm{v}_t \right)$& 
   $ {\color{blue}\beta_t}\left\|\bm{v}_t-g(\psi(\bm{S}_t), \bm{k}_t)\right\|_p^p +\frac{{\color{blue}\bm{\alpha}_t}}{2}\left\|\bm{S}_t\right\|_2^2$\\
   & Titans \cite{behrouz2024titans} & $\bm{M}_t = {\color{blue}\gamma_t}\bm{M}_{t-1}+\bm{S}_t,~\bm{S}_t={\color{blue}\alpha_t}\bm{S}_{t-1}-{\color{blue}\beta_t}\nabla_{M}\mathcal{L}\left(\bm{M}_{t-1},\bm{k}_t,\bm{v}_t \right)$ &
   $ {\color{blue}\beta_t}\left\|\bm{v}_t-\psi(\bm{S}_t), \bm{k}_t\right\|_2^2 +\frac{{\color{blue}\bm{\alpha}_t}}{2}\left\|\bm{S}_t\right\|_2^2$\\
    & Atlas \cite{behrouz2025atlas}  & $\bm{M}_t = {\color{blue}\gamma_t}\bm{M}_{t-1}-{\color{blue}\eta_t}\operatorname{Muon}(\bm{S}_t),~\bm{S}_t={\color{blue}\alpha_t}\bm{S}_{t-1}-{\color{blue}\beta_t}\nabla_{M}\mathcal{L}\left(\bm{M}_{t-1},\bm{k}_t,\bm{v}_t \right)$ &
    $ {\color{blue}\beta_t}\left\|\bm{v}_t-\psi(\bm{S}_t), \bm{k}_t\right\|_2^2 +\frac{{\color{blue}\bm{\alpha}_t}}{2}\left\|\bm{S}_t\right\|_2^2$ \\
  \bottomrule
\end{tabular}
\vspace{0.5em}
\end{threeparttable}
\end{adjustbox}
\vspace{-1em}
\label{tab: models}
\end{table}

\subsubsection{A Memory Perspective}
In this part, we adopt the conceptual framework proposed in Comba \cite{hu2025comba}, wherein "linear" specifically refers to models whose dynamics with respect to key-value associative memory are linear.

\textbf{Linear Update Rule.}
In contrast to autoregressive Transformers \cite{vaswani2017attention}, which preserve all contextual information explicitly in a key-value (KV) cache, linear RNNs distill higher-level representations into a fixed-size hidden state to enhance generalization. From a metaphysical perspective, this transition can be viewed as a shift from conductive to inductive attention \cite{hu2025comba,zancato2024b}, reflecting the notion that the compression is the intelligence \cite{huang2024compression}. Structurally, they bear similarities to energy-based models \cite{lecun2006tutorial} such as Hopfield networks \cite{mceliece1987capacity,farhat1985optical} and neural systems employing Hebbian learning principles \cite{chakraverty2019hebbian}. Early instances, including Linformer \cite{wang2020linformer}, S4 \cite{gu2021efficiently}, and RetNet \cite{sun2023retentive}, did not incorporate sufficiently adaptive, data-driven memory management, which limited their performance relative to models utilizing softmax attention. More recent approaches, exemplified by Mamba \cite{gu2023mamba} and GLA \cite{yang2023gated}, have mitigated this limitation by adopting dynamic, projection-based gating mechanisms, thereby achieving notable gains. From a formal perspective, these architectures can be regarded as linear register systems equipped with key-value associative memory: information is written through learnable forgetting and input gates (denoted as $\bm{\alpha}$ and $\bm{\beta}$) and subsequently accessed by query-based retrieval:
\begin{equation}
    \bm{S}_{t}=(\bm{\alpha}_t,\bm{\beta}_t) @{(\bm{S}_{t-1}, \bm{k}_t^\intercal\bm{v}_t)}^\intercal~~\textit{(Write)},\quad
    \bm{o}_{t}=\bm{S}_{t}\bm{q}_{t}~~\textit{(Read)}
\end{equation}

\textbf{Bilinear Update Rule.}
From the perspective of neural memory systems \cite{gershman2025key}, achieving effective memory regulation remains a fundamental challenge. In contrast to the Hebbian learning principle \cite{chakraverty2019hebbian}, which depends on reinforcement-driven updates, the Delta learning rule \cite{prados1989neural} emphasizes supervised control of memory traces. These systems are linear with respect to state and input individually, but nonlinear overall due to the product term (e.g., $\bm{Sk}$). They are regarded as a special class of nonlinear systems that preserve controllability \cite{bruni1974bilinear,zhao2016gramian,wang2023expectation,pardalos2010optimization}. For a general formulation:
\begin{equation}
    \bm{S}_{t}=(\bm{\alpha}_t,\bm{\beta}_t) @{(\bm{S}_{t-1}, \bm{k}_t^\intercal\bm{v}_t)}^\intercal + \gamma_t\bm{S}_{t-1}\bm{k}_{t}~~{{\textit{(Write)}}},\quad
    \bm{o}_{t}=\bm{S}_{t}\bm{q}_{t}~~{{\textit{(Read)}}}
\end{equation}
Where $\bm{Sk}$ is the bilinear term in models such as Comba \cite{hu2025comba}, (Gated)-DeltaNet \cite{yang2024parallelizing,yang2024gated}, and RWKV7 \cite{peng2025rwkv}. Alternatively, under the constraint that the spectral radius of the state transition matrix remains less than 1, it can also adopt an interaction between $\bm{S}$ and $\bm{v}$, similar to that in Lattice \cite{karami2025lattice}. Under the bilinear update rule, these models can be computed efficiently on GPUs through chunk-wise parallelism.

\textbf{Nonlinear Update Rule.}
Apart from early models such as LSTM \cite{hochreiter1997long} and GRU \cite{cho2014learning}, the series of models discussed in $\S$\ref{subsec:ttt} (e.g., TTT) can be regarded as modern versions of nonlinear RNNs. These models employ nonlinear operations for state memory (such as nonlinear activations in a two-layer MLP memory), resulting in inherently nonlinear dynamics. While this provides theoretically stronger expressiveness, these models are constrained by block-wise parallel computation and can only be updated via minibatch gradient descent, which leads to very low hardware utilization (less than 5\%). A potential solution is to follow LaCT \cite{zhang2025test} by adopting large-batch gradient descent, combined with sliding-window attention for intra-batch information learning as a hybrid architecture.

\subsubsection{An Optimizer Perspective}
Another perspective for unified linear sequence modeling is from an optimizer angle, which was initially proposed in TTT~\cite{sun2024learning}, Titans~\cite{behrouz2024titans} and Test-time Regression \cite{wang2025test}. We further classify the methods according to different forms of objective loss as below.

\textbf{Local L1 Loss.}
Early models like linear attention \cite{wang2020linformer} used an L1 loss to optimize the model in a single step, but this update scheme is unbounded, making it easy for $\bm{S}$ to suffer from memory conflicts. Subsequently, models such as RetNet \cite{sun2023retentive}, Mamba \cite{gu2023mamba}, and GLA \cite{yang2023gated} introduced memory gating mechanisms, which can be seen as a form of L2 regularization.

\begin{equation}
\hat{\bm{S}}_t=\arg \min _{\bm{S}}\left[ \beta_t\left|\bm{v}_t-\bm{S}_t \bm{k}_t\right| + \frac{1}{2} \operatorname{Tr}\left(\bm{S}_t^{\top} \Lambda \bm{S}_t\right) \right]
\end{equation}

\textbf{Local L2 Loss.}
However, an L1 loss typically cannot directly enforce $\bm{v}=\bm{S}\bm{k}$, and thus cannot achieve precise key-value associative memory retrieval. A better choice is the squared loss, which provides stronger memory management capabilities. Such models also include an L2 regularization term on the memory states.
\begin{equation}
\hat{\bm{S}}_t=\arg \min _{\bm{S}} \frac{1}{2} \left[\beta_t\left\|\bm{v}_t-\bm{S}_t \bm{k}_t\right\|^2 +  \operatorname{Tr}\left(\bm{S}_t^{\top} \Lambda \bm{S}_t\right)\right]
\end{equation}


\textbf{Multi-step L2 Loss.}
Models such as Delta‑Product \cite{siems2025deltaproduct} apply a sequence of Householder transformations within a single timestep, effectively treating the latent update as a composition of structured orthogonal operators. This allows the training objective to be interpreted as a form of multi-step L2 loss, where the model is supervised not just on immediate transitions but on future latent states over multiple horizons.

In this formulation, a single-step L2 loss encourages the model to approximate simple reflections, which is the most basic form of orthogonal transformations \cite{hu2025comba}. In contrast, a multi-step L2 loss allows the model to approximate general orthogonal matrices through compositions of reflections, thereby capturing richer linear dynamics such as rotations, permutations, or shears. This property significantly improves performance on state-tracking tasks like S5 \cite{merrill2024illusion}. Specifically, Delta‑Product formalizes the following objective:
\begin{equation}
\hat{\bm{S}}_t=\arg \min _{\bm{S}} \frac{1}{2} \sum_{i=1}^{n}\left[\beta_{ti}\left\|\bm{v}_{ti}-\bm{S}_{ti} \bm{k}_{ti}\right\|^2 \right]\quad \textit{(~Step by Step~)}
\end{equation}

\textbf{Global L2 Loss.}
Inspired by Recursive Least Squares (RLS), MesaNet \cite{von2025mesanet} defines a global L2 optimization for memory state, enabling the model to take the global optimum into account at each update step.
\begin{equation}
\hat{\bm{S}}_t=\arg \min _{\bm{S}} \frac{1}{2} \left[ \sum_{i=1}^t \left(\gamma_i\left\|\bm{v}_i-\bm{S} \bm{k}_i\right\|^2 \right) + \operatorname{Tr}\left(\bm{S}^{\top} \Lambda \bm{S}\right)\right]
\end{equation}

By deriving the analytical solution of this formula, two linear recursions can be used for fast computation. To further improve computational efficiency, gradient conjugation is applied for correction at the model output instead of directly computing the matrix inverse.

\subsection{Linearization}
\label{subsec:linearization}

\begin{figure}[t]
    \centering
    \includegraphics[width=0.75\linewidth]{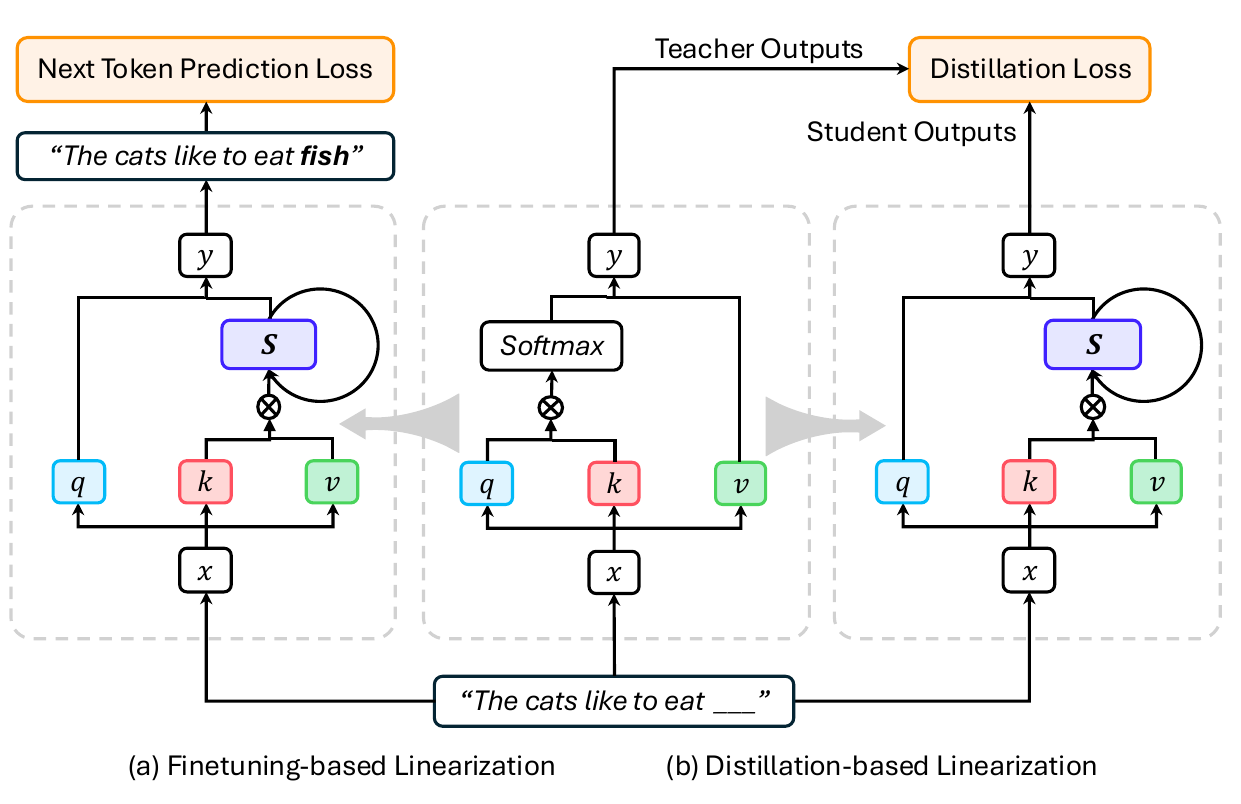}
    \caption{Mechanism Comparison of Finetuning-based and Distillation-based Linearization Procedures.}
    \label{fig:linearization}
\end{figure}

The Transformer-based architecture used in modern LLMs suffers from quadratic computational complexity and glowing memory usage. In contrast, the linear recurrent model architecture benefits from linear computational complexity and constant memory usage, solving the inefficiency challenge of the Transformer architecture. Despite this, the pretraining process for LLMs from inception necessitates substantial computational and financial resources, which remains a significant barrier to their widespread adoption and real-world implementation. In this case, \textbf{linearization} is a more ideal choice, whose purpose is to convert the pretrained transformer-based LLM architecture into a linear recurrent structure using less training and fine-tuning costs, and restore the capabilities of the original model on natural language understanding and generation scenarios.


\textbf{Finetuning-based Linearization.} Finetuning-based linearization directly replaces standard softmax attention with linear sequence modeling in a pre-trained transformer, then finetune the modified model for architecture adaption without  relying on knowledge from external models. Transformer-to-RNN (T2R)~\cite{kasai2021finetuning} introduces a method to convert a pretrained transformer to an RNN with linear-time computation and constant memory complexity. T2R follows a swap-then-finetune procedure to align the attention computation of a pretrained transformer, then finetunes the converted RNN for model architecture adaption. Mao~\cite{mao2022fine} investigates various update rule configurations to finetune pretrained autoregressive Transformers into RNNs and propose a simple element-wise decay fast weights update rule with no feature map for linearizing transformers. DiJiang \cite{chen2024dijiang} leverages linear attention with the weighted Quasi-Monte Carlo method for efficient sampling. Based on the Discrete Cosine Transform (DCT) kernelization process, DiJiang significantly reduce the training cost of the transformation of a pre-trained vanilla Transformer into a linear complexity model. SUPRA \cite{mercat2024linearizing} introduces a linearization technique for converting large-scale pretrained softmax-attention transformers into RNNs by replacing the softmax operation with GroupNorm and using a small MLP for queries and keys feature mapping. Liger \cite{lan2025liger} proposes a novel linearization technique for converting Transformer-based LLM to gated recurrent structures by fully reuse original pretrained model weights without introducing any extra learnable parameters, thus simplifying the linearization process to a single-stage end-to-end training without relying on distillation and significantly reducing the linearization cost.

\textbf{Distillation-based Linearization.} Distillation-based linearization utilizes knowledge distillation \cite{hinton2015distilling} to transfer the capabilities of a pre-trained transformer with standard softmax attention as a teacher model to a student model with linear sequence modeling. LoLCATs \cite{zhang2024lolcats} proposes a simple two-step linearization process via Low-rank Linear Conversion via Attention Transfer, significantly improves linearizing performance, training efficiency, and scalability with low-rank adaption (LoRA). Llamba \cite{bick2025llamba} demonstrates the effectiveness and scalability of cross-architecture distillation based on Llama-3.x series using MOHAWK \cite{bick2024transformers}. LightTransfer \cite{zhang2025lighttransfer} transforms standard transformers into hybrid model architecture by identifing lazy layers which are and replacing their full attention layers with efficient streaming attention, demonstrating throughput efficiency and effectiveness on long-context tasks with minimal performance loss. MOHAWK \cite{bick2024transformers} proposes a distillation procedure for converting Transformer to Mamba. By incorporating three-stage progressive training: 1) matrix orientation, 2) hidden state alignment, and 3) weight transfer and knowledge distillation, MOHAWK can distill the Transformer architecture by matching different degrees of granularity in the SSM.  MambaInLlama \cite{wang2024mamba} also adopts a similar multi-stage training procedure to obtain Mamba model initialized from large Transformer model by combining cross-architecture distillation and hardware-aware speculative decoding algorithm, further accelerate the inference speed of Mamba and its hybrid models. Lizard \cite{van2025lizard}, similar as Liger \cite{lan2025liger}, also considers introducing gating mechanism into linearization process combining with sliding window attention with meta memory, and designed a hardware-efficient factorization algorithm that incorporates log-gating values into the query and key projections for training stability. LoLA \cite{mcdermott2025lola} integrates sparse caching technique into linear attention, further improving language modeling and long-context ability after model architecture linearization. Yeonju et al. \cite{ro2025onthefly} proposes a dual-state linear attention (DSLA) for maintaining long-term historical information and tracking short-term recency-thereby, and introduced an online adaptive distillation system for linearizing quadratic complexity self-attention layers to linear-cost DSLA layers.

\textbf{Linearization in the Era of RL Scaling.} As the development of R1-like large reasoning models \cite{guo2025deepseek,qu2025survey,li2025system,chen2025towards} have been demonstrated their supreme advantages on reasoning tasks, one significant question is prompted: \textit{Can we boost linear recurrent models' reasoning performance by scaling test-time computation through long CoT?}

A promising direction involves leveraging existing transformer-based reasoning models to create \textbf{linearized} LLMs—models that retain strong reasoning capabilities while benefiting from the efficiency of linear sequence modeling. This transformation is typically achieved through architectural linearization followed by SFT and RL to enhance test-time reasoning performance. Daniele et al.~\cite{paliotta2025thinking} explore this by distilling both pure and hybrid Mamba models from large-scale LLaMA models, investigating their reasoning scalability at test time. Similarly, M1~\cite{wang2025m1} introduces a hybrid large reasoning model built on the Mamba architecture, achieved by distilling and fine-tuning a linearized transformer. Their results highlight the potential of linear recurrent architectures in supporting efficient and scalable reasoning.

\subsection{Hardware-efficient Implementation}
\label{subsec:hardware-linear}

One of the key reasons for the rapid development and adoption of linear sequence modeling methods is the availability of hardware-efficient implementations. Many approaches provide optimized code using tools such as Triton or CUDA to achieve efficient computation on Modern GPUs. In the following, we review and summarize these hardware-efficient implementation methods.

\begin{figure}[t]
\centering
    \includegraphics[width=0.9\linewidth]{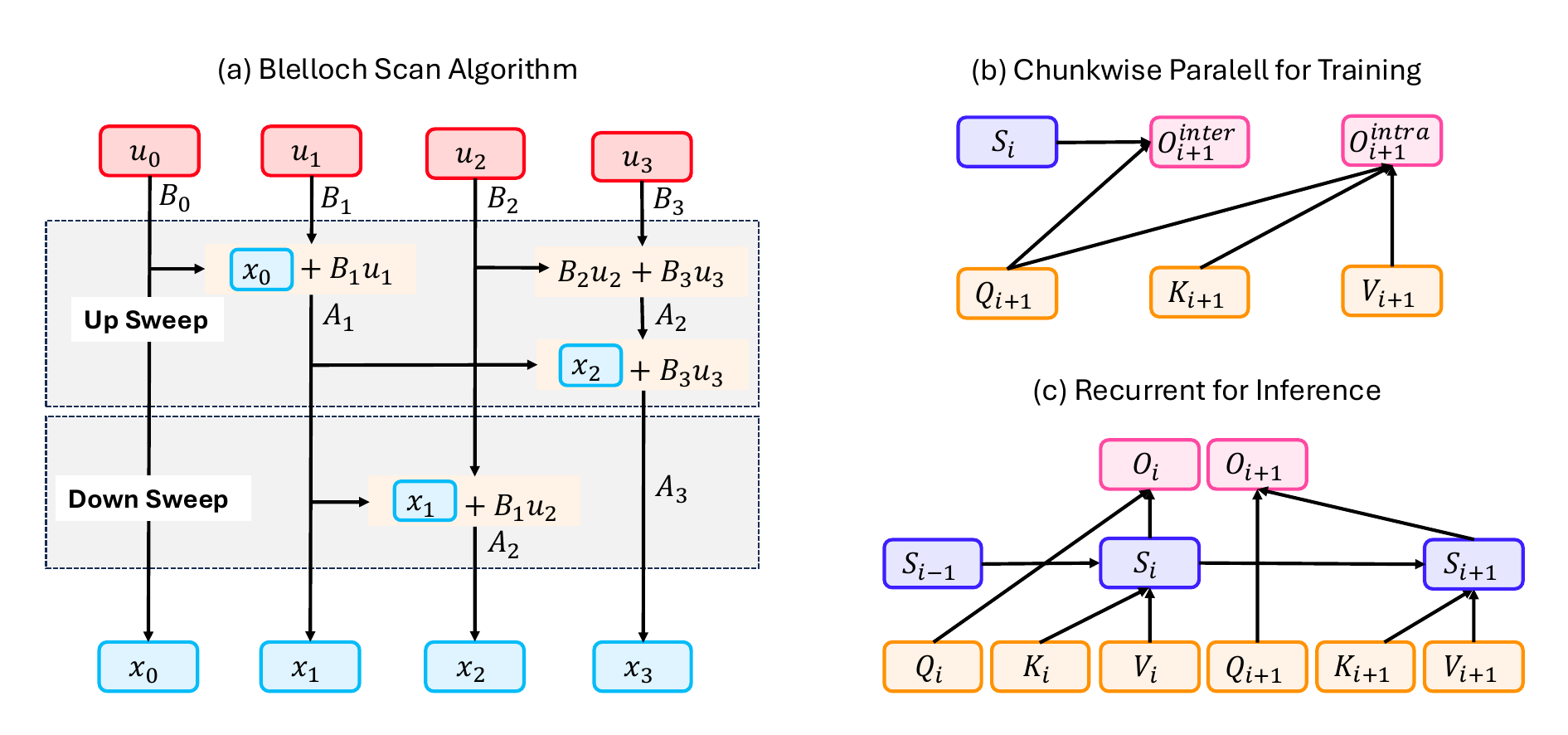}
    \caption{Hardware-efficient Implementation Algorithms for Linear Sequence Modeling. (a) In the Blelloch parallel scan, the upward sweep computes and stores the outputs at even indices; the downward sweep then reuses these stored values to compute the outputs at odd indices.
(b) To better exploit tensor‑core matmul acceleration during training and prefilling, linear‑recurrence models can adopt intra‑block parallel computation combined with inter‑block recursion.
(c) During inference, the recurrent formulation supports decoding with constant memory and $\mathcal{O}(1)$ per‑step computational complexity.}
    \label{fig:hardware-efficient}
\end{figure}

\textbf{Fast Recurrent.}
Early linear sequence modeling methods often leverage the properties of structured matrices for acceleration. For example, state-space models like S4 \cite{gu2021efficiently} can be written in the form of convolutions to enable fast computation. Models like Mamba \cite{gu2023mamba} use the Blelloch scan algorithm \cite{blelloch1990prefix,smith2022simplified} to achieve fast recursion. Specifically, the Blelloch scan operates as a tree-like structure. During the computation, the even positions in the sequence are first processed via an upward scan, which serves as an intermediate step. Then, during the downward scan, the odd positions of the sequence are computed. This computational approach has recently inspired works such as multi-scale state-space models \cite{hu2024attractor} and log-linear attention mechanisms \cite{guo2025log,yau2025sequential}.

\textbf{Chunk-wise Parallel.}
Although Linear RNNs achieve a favorable pretraining time complexity of $\mathcal{O}(Nd^2)$, they are often slower than softmax attention, which has a complexity of $\mathcal{O}(N^2d)$, on shorter sequences. This is primarily due to the fact that current hardware is highly optimized for \texttt{matmul} operations, which limits the efficiency of linear recurrence and necessitates additional training optimizations. Recent methods inspired by FlashAttention \cite{dao2022flashattention}, including Lightning Attention \cite{qin2024lightning,qin2024various,li2025minimax}, GLA \cite{yang2023gated}, and Mamba2 \cite{dao2024transformers}, introduce inter-chunk recurrence combined with intra-chunk parallelism to fully leverage matrix compute throughput. These techniques optimize computation graphs and algorithms, ensuring efficient parallel execution across data chunks, which significantly enhances model performance on modern hardware. A basic formulation using chunk size $C$ can be expressed as:
\begin{equation}
\bm{S}_{[t+1]} = 
   \bm{S}_{[t]}+\bm{V}_{[t]}^{\intercal}\bm{K}_{[t]}~\in\mathbb{R}^{D\times D},\quad
    \bm{O}_{[t]} = 
   \bm{Q}_{[t]}
    \bm{S}_{[t]}^\intercal + (\bm{Q}_{[t]} \bm{K}_{[t]}^{\intercal} \odot \operatorname{Mask}_{[t]} )\bm{V}_{[t]}~\in\mathbb{R}^{C\times D}
    \label{eq: seq-parallel}
\end{equation}

The open-source framework \textit{Flash Linear Attention}~\cite{yang2024fla} offers a collection of chunk-wise parallel Triton kernels specifically designed for linear sequence modeling, with a strong emphasis on hardware efficiency. These implementations not only accelerate the computation of linear attention mechanisms but also enable easy integration into existing model training and inference pipelines. By making such optimized kernels publicly available, this framework lowers the barrier for researchers and practitioners to experiment with and deploy efficient linear attention variants, promoting further innovation within the community.

\section{Sparse Sequence Modeling}
\label{sec:sparse}

\begin{figure*}[t]
    \centering
    \includegraphics[width=\linewidth]{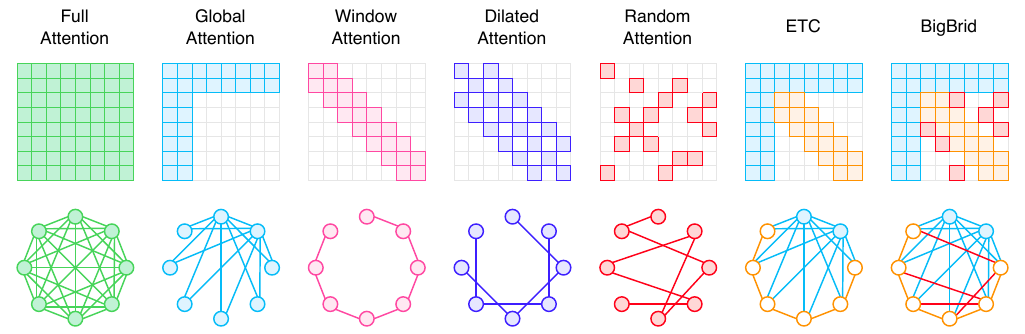}
    \caption{\textbf{Example Patterns of Static Sparse Attention.} ETC and BigBird are representative examples of mixed static sparse attention mechanisms, which combine multiple fixed sparsity patterns within a single attention framework.}
    \label{fig:sparse_attention}
\end{figure*}

Sparse sequence modeling is a paradigm designed to enhance the efficiency of processing sequential data by limiting interactions between elements to a strategically chosen subset. A typical example is sparse attention mechanism in transformer models, which addresses the computational bottlenecks of traditional full attention methods while aiming to preserve modeling performance.

\subsection{Static Sparse Attention}
\label{subsec:static sa}

Unlike full self-attention, which scales quadratically with sequence length, static sparse attention reduces computational complexity by restricting each token to attend to a predefined, fixed subset of other tokens. These static patterns are designed before training and remain unchanged during inference, making them highly efficient and easy to deploy. Common structural partterns include global, window, strided, dilated, random and blockwise sparsity. Such approaches have been widely adopted in natural language, vision, and multimodal models due to their strong inductive biases and scalability~\cite{nawrot2025sparse}.

Early work on Sparse Transformer~\citep{child2019generating} pioneered this idea by introducing fixed strided and dilated attention patterns that ensure each token connects to both nearby and distant tokens in a deterministic manner. This structure reduces computation to near-linear complexity while preserving the model’s ability to capture long-range dependencies, establishing a foundation for later sparse models. Building on this foundation, Star-Transformer~\citep{guo2019star} proposed a radial topology with a central relay node that connects to all tokens, while peripheral tokens form a ring with local connections. This architecture maintains global communication with linear-time attention and performs well on tasks requiring global structure, making it both simple and efficient. BlockBERT~\citep{qiu2019blockwise} adapts BERT for longer sequences using blockwise sparse attention. It divides input into fixed-size blocks, allowing dense intra-block attention and sparse inter-block communication through selected key tokens. This enables effective long-document modeling while lowering GPU memory requirements, proving useful for document classification and QA tasks. 

To enhance static sparse models with richer contextual representations, Longformer~\citep{beltagy2020longformer} extends this idea with a combination of sliding window attention and a small set of global tokens. The window handles local context, while global tokens enable information flow across distant positions. This hybrid setup allows linear complexity and proves effective in tasks like summarization and long-context QA. GMAT~\citep{gupta2020gmat} injects global memory tokens that act as information hubs, enabling interactions across segments of long sequences. Compatible with other sparse structures like Longformer and Sparse Transformer, GMAT demonstrates improved performance in long-context understanding tasks. A similar dual-attention structure is introduced in ETC~\citep{ainslie2020etc}, which separates tokens into local and global streams. Local tokens attend within sliding windows, while global tokens attend to the full sequence, allowing hierarchical representations. ETC also incorporates segment-aware attention and relative position encodings, boosting performance in document-level comprehension. BigBird~\citep{zaheer2020big} generalizes static sparse patterns by integrating local, global, and random connections, forming a sparse attention graph with small-world properties. This design ensures all tokens remain connected through short paths, and provides theoretical guarantees of universal approximation. BigBird scales effectively to long sequences in both encoder-only and encoder-decoder setups, supporting tasks across NLP and bioinformatics. LongT5~\citep{guo2021longt5} builds upon T5 for long-text generation using transient global attention. It summarizes local contexts into temporary global tokens, which adaptively aggregate information layer by layer. This allows encoder-decoder models to scale to longer inputs without relying on dense attention, improving performance in summarization and long-form QA. Pushing the boundary further, LongNet~\citep{ding2023longnet} introduces exponentially dilated attention, where each layer increases the attention span in powers of two. This hierarchical scheme enables efficient modeling of sequences up to 1 billion tokens with logarithmic complexity.

Beyond text, static sparse attention has also proven effective in vision. Axial Attention~\citep{ho2019axial} factorizes 2D attention into independent operations along height and width axes, dramatically reducing complexity while maintaining expressiveness. This technique underpins many high-resolution vision transformers and has seen wide adoption in image and video models. In multimodal settings, sparse spatiotemporal attention mechanisms, such as those used in Open-Sora~\citep{opensora}, decouple attention along spatial and temporal dimensions for efficient video modeling. By applying windowed attention in space and strided or pooled attention over time, these methods significantly reduce compute cost while preserving temporal coherence, enabling scalable video understanding and generation in large multimodal models.

\subsection{Dynamic Sparse Attention}
\label{subsec:dynamic sa}

Unlike static sparse attention, dynamic sparse mechanisms determine attention patterns adaptively based on the input content. These models aim to approximate the expressiveness of full attention by focusing computation on a dynamically selected subset of token interactions, thereby retaining task-relevant information over long contexts while minimizing computational overhead. The evolution of these methods can be seen as a progress from heuristic grouping techniques to more sophisticated, learnable retrieval systems.

Early approaches focused on grouping or clustering semantically similar tokens to restrict the scope of attention. Transformer-XL~\cite{dai2019transformer}, proposes to create a recurrent memory by reusing hidden states from previous segments, breaking the fixed-length barrier. Compressive Transformer~\cite{rae2019compressive}, refines this memory by actively compressing older states into a more efficient long-term store. Adaptive Attention Span~\cite{sukhbaatar2019adaptive} was proposed to tackle the problem from an efficiency angle, allowing each attention head to learn its optimal context size, which dramatically extended the manageable sequence length without a proportional increase in computational cost. Furthermore, Reformer~\citep{kitaev2020reformer} pioneered using locality-sensitive hashing (LSH) to bucket tokens, allowing each query to attend only to keys within the same hash bucket. This created content-dependent sparsity with near-linear complexity. Similarly, the Routing Transformer~\citep{roy2021efficient} applies online k-means clustering to partition tokens, with attention confined to these dynamically formed clusters. A related idea, Sparse Sinkhorn Attention~\citep{tay2020sparse}, learns a differentiable permutation to reorder tokens, enabling efficient block-wise attention on soft-sorted, semantically coherent segments.

These heuristic-based grouping methods were later generalized under the unified framework of {Attention with Bounded-Memory Control (ABC)}~\citep{peng2021abc}. This work frames the context as a memory of a fixed, constant size and demonstrates that many prior methods are specific instances of managing this memory. Its key innovation, {ABC-MLP}, moves beyond fixed heuristics by introducing a learned, contextualized controller to manage the memory, achieving a better trade-off between efficiency and accuracy. An alternative to compressing the immediate context is to augment the model with an external memory and use retrieval mechanisms for sparsity. {Memorizing Transformers}~\citep{wu2022memorizing} implements this by using a k-nearest-neighbor (kNN) index to retrieve relevant past token representations, enabling sparse access to long-term memory. 
{CoLT5}~\citep{ainslie2023colt5} introduced a conditional routing approach where lightweight routers select a small subset of "important" tokens for global attention, while most tokens undergo cheaper local attention. {Unlimiformer}~\citep{bertsch2024unlimiformer} refined this concept for cross-attention, allowing pre-trained models to attend over vastly longer inputs by retrieving relevant keys from an index without any weight modification. The focus then shifted towards learning explicit routing and gating mechanisms to prune the attention matrix directly. 

The latest advancements aim not only for theoretical efficiency but also for practical, hardware-aligned speedups and architectural specialization. {Native Sparse Attention (NSA)}~\citep{yuan2025native} is designed for this purpose, employing a hardware-aligned hierarchical strategy that combines coarse-grained token compression for global context with fine-grained selection for local precision, delivering significant real-world speedups. Concurrently, {Mixture-of-Sparse-Attention (MoSA)}~\citep{pikekos2025mixture} adapts the popular Mixture-of-Experts (MoE) paradigm to dynamic sparse attention. Each head acts as an "expert" that dynamically selects a small subset of tokens, and the computational savings are reinvested to train a larger number of specialized heads, achieving superior performance within the same FLOP budget.

\subsection{Training-free Sparse Attention}
\label{subsec: training-free sa}

While many sparse attention methods are designed for training, a significant body of work focuses on accelerating inference, which is composed of two distinct phases: the computationally-intensive prefill of processing the initial prompt, and the memory-bandwidth-bound decoding stage of generating subsequent tokens. Different techniques have emerged to tackle the specific bottlenecks in each phase.

\textbf{Accelerating Prefill Stage.}
Optimizing the prefill stage involves reducing the massive computation of the initial self-attention pass over a long prompt. Early approaches impose structured sparsity based on offline analysis. For instance, LongLoRA~\cite{chen2023longlora} fine-tunes a model for longer contexts by using shifted sparse attention (S2-Attn). Instead of full attention, S2-Attn performs attention within local token groups and then shifts the groups to allow information to flow across the entire context, approximating global attention with significantly less computational cost. {MInference}~\citep{jiang2024minference} observes that attention maps often conform to a few prototype shapes (e.g., diagonal, vertical stripes) and leverages specialized GPU kernels for these fixed patterns. Similarly, {MoA (Mixture of Attention)}~\citep{fu2024moa} uses gradient-based profiling to assign a static, heterogeneous sliding-window size to each attention head, effectively creating a fixed sparse mask that reduces computation without retraining.

A more adaptive approach involves learning the sparsity pattern dynamically based on the input content. {SeerAttention}~\citep{gao2024seerattention} exemplifies this by augmenting each attention layer with a lightweight gating module. After a brief self-distillation phase, this gate learns to predict which blocks of the attention matrix are most important for a given input, generating a dynamic mask that prunes irrelevant computations on-the-fly. This allows the model to achieve significant prefill speedups while maintaining high accuracy by adapting the sparsity pattern to the specific context. The follow-up work SeerAttention-R~\cite{gao2025seerattention} builds on self-distilled attention sparsity and introduces key changes for efficient auto-regressive decoding. It removes sequence-level query pooling and adopts a GQA-style shared sparsity pattern for better hardware efficiency. SeerAttention-R can be applied to any pretrained Transformer by inserting a learnable gate into the attention layer, without modifying the original model weights.

\textbf{Accelerating Decoding Stage.}
In the auto-regressive decoding stage, the primary bottleneck is not computation but the memory bandwidth required to load the ever-growing Key-Value (KV) cache at each step. Research in this area focuses on intelligently pruning this cache to retain only the most critical information. SpAtten~\cite{wang2021spatten} proposes a sparse attention architecture that improves efficiency through cascade token and head pruning, removing unimportant elements layer by layer using a Top-K engine. It also introduces progressive quantization, loading low-precision bits only when needed. Together, these methods reduce memory and computation without harming accuracy, enabling faster long-context inference. Another simple yet powerful heuristic is to exploit the "attention sink" phenomenon, as demonstrated by {StreamingLLM}~\citep{xiao2023efficient}. This work found that initial tokens consistently attract high attention throughout generation and are crucial for maintaining stability. By caching only these sink tokens alongside a sliding window of recent tokens, models can handle infinitely long streams with a small, fixed-size cache. Other methods adopt dynamic eviction policies. {TOVA}~\citep{oren2024tova} continuously evicts the token with the lowest attention score to make room for new ones, while {H2O}~\citep{zhang2024h2o} formulates eviction as a submodular optimization problem to retain a set of "heavy hitter" tokens that have the greatest influence on the attention output. RetrievalAttention~\cite{liu2024retrievalattention} accelerates inference by building an approximate nearest neighbor index of the KV cache in CPU memory. During generation, it uses an attention-aware vector search to retrieve only a small subset of the most relevant KV pairs for the current query, thus avoiding the computational and memory costs of the full attention mechanism.

More structured approaches introduce sparsity at a coarser granularity. {FastGen}~\citep{ge2023fastgen} profiles each attention head to classify its behavior (e.g., local vs. global) and applies a tailored eviction policy to each, compressing the KV cache more aggressively for heads with localized attention patterns. Taking this further, query-aware methods select relevant context blocks on-the-fly. {Quest}~\citep{tang2024quest} divides the KV cache into pages and uses a lightweight scoring function to retrieve only the most relevant pages for the current query. Similarly, {LongHeads}~\citep{lu2024longheads} re-purposes multi-head attention by allowing each head to independently select and attend to a small number of context chunks, effectively parallelizing the search for relevant information across the sequence. DuoAttention~\cite{xiao2024duoattention} separates attention heads into retrieval heads for long-term context and streaming heads for local context. Only retrieval heads maintain full KV caches, while streaming heads use sliding windows, reducing memory and latency. A data-driven method is used to identify retrieval heads, enabling efficient long-context decoding with minimal performance loss. 
ShadowKV~\cite{sun2024shadowkv} offloads the KV cache of a large target model to the CPU, while a smaller draft model maintains a compact "shadow" KV cache on the GPU. The draft model generates tokens using speculative decoding, and the target model only retrieves the necessary KV data from the CPU to verify the draft, minimizing slow memory access. LServe~\cite{yang2025lserve} introduces a unified sparse attention design for efficient long-sequence serving. It combines block-level token skipping with query-driven KV pruning, retaining only relevant memory pages. By converting half of the heads into lightweight streaming heads, LServe accelerates both prefill and decoding stages while preserving accuracy.
PQCache~\cite{zhang2025pqcache} compresses the KV cache using Product Quantization (PQ) to group and quantize vector dimensions into low-bit representations. To maintain accuracy, it introduces a lightweight, learnable "dequantization helper" module that is trained to reconstruct the high-fidelity vectors from their compressed form during inference. XAttention~\cite{xu2025xattention} proposes an efficient block-sparse attention framework using the sum of antidiagonals as a lightweight block importance metric. This enables accurate pruning with minimal overhead, achieving up to 13.5× speedup while maintaining accuracy across long-context benchmarks.

\subsection{Hardware-efficient Implementation}
\label{subsec:hardware sa}
Efficient hardware implementation plays a key role in scaling sparse attention to support longer sequences without compromising runtime or memory efficiency. 
This section reviews recent approaches aimed at enhancing the hardware efficiency of sparse attention computation.

Except exact attention optimization, FlashAttention-1~\cite{dao2022flashattention} also incorporates block-sparse FlashAttention by introducing structured sparsity to further reduce memory access overhead and accelerate computation for long sequences. By applying a predefined block-wise sparsity mask, it skips unnecessary attention blocks during computation:
\begin{equation}
\bm{S} = \bm{QK}^T \in \mathbb{R}^{N \times N}, \quad
\bm{P} = \mathrm{softmax}(\bm{S} \odot \tilde{\bm{M}}) \in \mathbb{R}^{N \times N}, \quad
\bm{O} = \bm{PV} \in \mathbb{R}^{N \times d}
\end{equation}
where $\tilde{\bm{M}}$ denotes the block mask and aligns with the block structure of $\bm{Q}$ and $\bm{KV}$ in the FlashAttention, specifically expressed as follows:
\begin{equation}
    S_{kl}=
    \left\{
    \begin{aligned}
        S_{kl},\quad & if\ \tilde{\bm{M}}_{kl}= \bm{M}_{ij}=1 \\
        -inf,\quad & if\ \tilde{\bm{M}}_{kl}=\bm{M}_{ij}=0
    \end{aligned}
    \right.
    \ \
    ,\ \ i = \lfloor k/B_r \rfloor, \ \ j = \lfloor l/B_c \rfloor, \ \ \bm{M} \in \{0, 1\}^{N/B_r \times N/B_c}
\end{equation}
where $B_r$ and $B_c$ denote the block sizes of $\bm{Q}$ and $\bm{KV}$, respectively. Empirically, block-sparse FlashAttention achieves $2-4\times$ speedups over dense FlashAttention and scales transformers to sequence lengths up to $64\text{K}$. FlashAttention-2~\cite{dao2023flashattention} extends FlashAttention-1’s block-sparse attention to support structured sparsity patterns such as local, dilated, and general block-sparse attention.

Recent work NSA~\cite{yuan2025native} integrates three different sparse attention mechanisms and improves hardware efficiency through three strategies. First, \textit{Group-Centric Data Loading} ensures that all query vectors $\bm{Q} \in \mathbb{R}^{[h, d_k]}$ in a GQA group and their corresponding sparse key/value indices $\bm{I}_t$ are loaded together in each inner loop. Second, \textit{Shared KV Fetching} minimizes memory access by sequentially loading continuous key/value blocks $\bm{K} \in \mathbb{R}^{[B_k, d_k]}$, $\bm{V} \in \mathbb{R}^{[B_k, d_v]}$ into SRAM, where $B_k$ is a kernel-aligned block size. Finally, by leveraging the uniform inner-loop length across blocks, the \textit{Outer Loop on Grid} utilizes Triton’s grid scheduler for efficient parallelization of query and output computation. This design gives 9.0×/6.0× speedups in forward/backward passes for 64k sequences compared to FlashAttention-2, showing strong performance for long-context tasks in GQA/MQA. 
Furthermore, MoBA~\cite{lu2025moba} implements dynamic sparse attention efficiently through blockwise variable-length computation. By partitioning sequences into fixed-size chunks and encoding block selections via segment length tensors, it transforms irregular sparsity into hardware-friendly memory access patterns. This varlen formulation maintains FlashAttention's optimization benefits while enabling adaptive routing, achieving 16× speedup for 10M-token sequences at $95\%$ sparsity. 

Besides block sparse attention, other sparse patterns also benefit from efficient kernels. Longformer’s~\cite{beltagy2020longformer} CUDA kernel implements optimized sparse attention with three key patterns: sliding window attention for local context, dilated window attention for wider receptive fields, and task-specific global attention for special tokens. SeerAttention~\cite{gao2024seerattention} introduces a CUDA kernel supporting dynamic sparsity shapes like A-shape, vertical-slash, and diagonal blocks. It integrates FlashAttention’s tiling scheme to skip inactive blocks, reducing computation and memory overhead. 

\section{Efficient Full Attention}
\label{sec:softmax}
\subsection{IO-Aware Attention}
\label{subsec:io-aware attn}
Transformer self‑attention scales quadratically in both compute and
memory with the sequence length~$N$, which quickly becomes the dominant
bottleneck in large‑context language models.
A series of \emph{FlashAttention} kernels
eliminates most of the memory traffic while preserving
exact softmax attention, delivering
large wall‑clock speed‑ups on modern GPUs. 

Given matrices $\bm{Q},\bm{K},\bm{V}\in\mathbb{R}^{N\times d}$, the standard softmax attention computation on GPU can be written as:
\begin{enumerate}[leftmargin=*]
\item Load $\bm{Q}$ and $\bm{K}$ by blocks from HBM, compute $\bm{S}=\bm{Q}\bm{K}^{\mathsf{T}}$, write $\bm{S}$ to HBM.
\item Read $\bm{S}$ from HBM, compute $\bm{P}=\operatorname{softmax}(\bm{S})$, write $\bm{P}$ to HBM.
\item Load $\bm{P}$ and $\bm{V}$ by blocks from HBM, compute $\bm{O}=\bm{P}\bm{V}$, write $\bm{O}$ to HBM.
\end{enumerate}
It suffers from several issues: 1) \textbf{High memory usage.} When the hidden dimension is large, it becomes difficult to load both $\bm{Q}$ and $\bm{K}$ into SRAM simultaneously for fast computation. 2) \textbf{Excessive HBM and SRAM access.} This leads to a major performance bottleneck during training. Specifically, standard softmax attention computation:
\begin{enumerate}[leftmargin=*]
    \item Requires reading $\bm{Q}$ and $\bm{K}$ from HBM twice, and writing the intermediate result $\bm{S}$ once. Total: 3 memory operations.
    \item Reads $\bm{S}$ once and writes $\bm{O}$ once. Total: 2 memory operations.
    \item Reads $\bm{P}$ and $\bm{V}$ twice, and writes $\bm{O}$ once. Total: 3 memory operations.
\end{enumerate}


\subsubsection{FlashAttention-1}
Exist algorithms aim to reduce attention computation complexity, but often do not result in significant speed-ups because GPU runtimes are mainly limited by global memory traffic, not computation intensity. FlashAttention-1~\cite{dao2022flashattention} overcomes this by optimizing memory transfers between global and on-chip memory. This optimization greatly reduces memory traffic and memory usage, resulting in faster computations and enabling longer sequence lengths without needing approximations.
FlashAttention-1 improves standard attention computation by dividing the sequence into smaller query and key-value tiles. The sequence is processed in blocks to minimize memory access. The query and key-value pairs are loaded into shared memory in one operation, and the score block is computed efficiently using matrix multiplication on specialized hardware. A partial softmax operation is applied across the tiles, maintaining running maximums and sums for numerical stability. The final step accumulates partial outputs with the computed probabilities.

The key contributions of FlashAttention-1 can be summarized as:

\textbf{Online Softmax.}  
FlashAttention-1 reduces memory usage by computing softmax normalization incrementally, keeping track of the running maximum and cumulative weights. This avoids storing all intermediate scores and ensures efficient memory usage.

\textbf{Fused Attention Computation.}  
Instead of performing separate kernel invocations for computing attention scores, softmax, and weighted sums, FlashAttention-1 combines these steps into a single kernel. This reduces redundant memory transfers and takes full advantage of GPU parallelism, improving computation efficiency.

\textbf{Recomputation in Backward Pass.}  
During backpropagation, FlashAttention-1 saves memory by recomputing attention weights instead of storing all intermediate results. It uses previously saved normalization statistics to recompute these values, allowing the method to handle longer sequences and larger batch sizes without exceeding memory limits.


\subsubsection{FlashAttention-2}
\label{subsec:fa-v2}

\begin{wrapfigure}{r}{0.32\textwidth}
  \begin{center}
  \vspace{-5em}
    \includegraphics[width=0.3\textwidth]{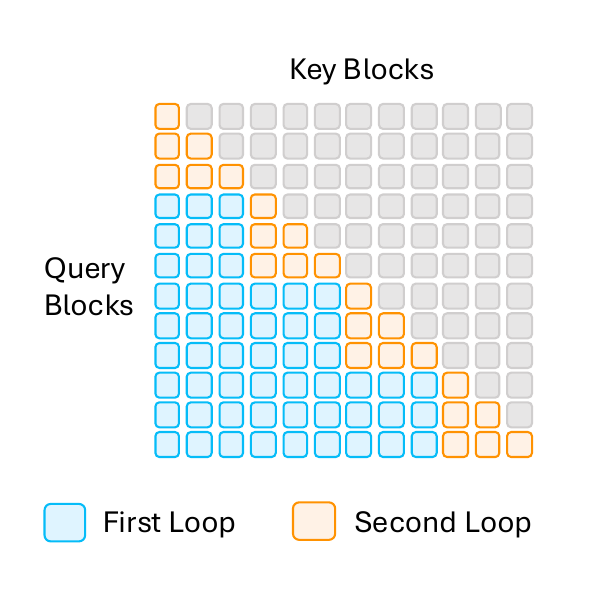}
  \end{center}
  \vspace{-1.75em}
  \caption{{Attention Map Computation in FlashAttention-2.}}
  \label{fig: flashattention2}
  \vspace{-3em}
\end{wrapfigure}

FlashAttention-1 significantly outperforms standard attention methods, but its forward pass still achieves only a fraction of the device's theoretical peak throughput. The backward pass performs even worse, reaching just a small portion of the peak throughput. The bottleneck arises due to suboptimal work partitioning across thread blocks and warps, as well as numerous non-matmul operations, which result in low occupancy and unnecessary memory accesses. 

FlashAttention-2~\cite{dao2023flashattention} addresses these issues and optimizes performance from the following aspects:

\textbf{More Matmul Operations.} FlashAttention-2 reduces softmax bookkeeping by storing the Intermediate terms and fusing epilogue operations directly within the Tensor-Core pipeline.

\textbf{Query-Outer, Key-Value-Inner Loop Structure.}  
In FlashAttention-2, the structure of the computation is optimized by processing the query and key-value pairs in a more efficient loop format, improving parallelism and reducing overhead.

\textbf{Row-wise Computation.}  
For each query block, attention is computed with every key and value block in the same row and the results are summed to produce the output; with causal attention, this is usually done in two passes: first, the outputs for all blocks strictly below the main diagonal are computed without any masking, and then a mask is applied to compute the lower-triangular part of each diagonal block:

\begin{equation}
\bm{A}_{[i],[j]} \propto\left\{\begin{array}{ll}
\exp \left(\bm{Q}_{[i]}^{\top} \bm{K}_{[j]}\right), & \text { if } i<j \\
\exp \left(\operatorname{lower}\left(\bm{Q}_{[i]}^{\top} \bm{K}_{[i]}\right)\right), & \text { if } i=j
\end{array} \right.
\end{equation}


\subsubsection{FlashAttention-3}
\label{subsec:fa-v3}

FlashAttention-3 \cite{shah2024flashattention} brings substantial improvements tailored for Hopper-class GPUs, integrating two pivotal hardware features: the Tensor Memory Accelerator (TMA) and the WGMMA Tensor-Core instructions. These innovations optimize performance by enabling warp-asynchronous memory operations and enhancing the efficiency of matrix multiplications, which are critical for accelerating attention mechanisms. 

The key contributions FlashAttention-3 can be summarized as:

\textbf{Producer-Consumer Asynchrony.}
This technique restructures the processing pipeline by assigning warp groups to distinct roles—\emph{producers} and \emph{consumers}. The producers are tasked with prefetching data from memory using TMA, while the consumers focus on the computationally intensive tasks of matrix multiplication and softmax operations. This separation of responsibilities allows for a two-stage pipeline that concurrently handles memory transfers and computation. By overlapping these operations, FlashAttention-3 minimizes idle time, improving resource utilization and increasing throughput. The result is more efficient hardware utilization, especially in high-latency stages of computation such as memory access.

\textbf{Interleaved Matmul+Softmax.}
In this method, blocks are processed in a double-buffered fashion, where the softmax operation for a block is computed while the next block's scores are calculated using WGMMA. This interleaving of the matrix multiplication and softmax operations ensures that the GPU's compute resources are kept fully occupied, reducing the time spent waiting between these two stages. This technique exploits the parallelism of modern GPUs to maintain high throughput and reduce latency, thus significantly accelerating the attention computation pipeline.

\textbf{Block-Wise FP8 Quantization with Incoherent Processing.}
In FlashAttention-3, all numerical operations, including GEMM and softmax, are performed using FP8 precision, which reduces memory footprint and improves computational efficiency. However, to mitigate the precision loss typically associated with FP8, each block independently selects a per-tile scaling factor, allowing it to adjust for cumulative errors that would otherwise degrade performance. This approach significantly reduces numerical errors compared to traditional FP8 kernels, ensuring that the results remain robust while maintaining the memory and computation benefits of reduced precision.





\begin{figure*}[t]
    \centering
    \includegraphics[width=\linewidth]{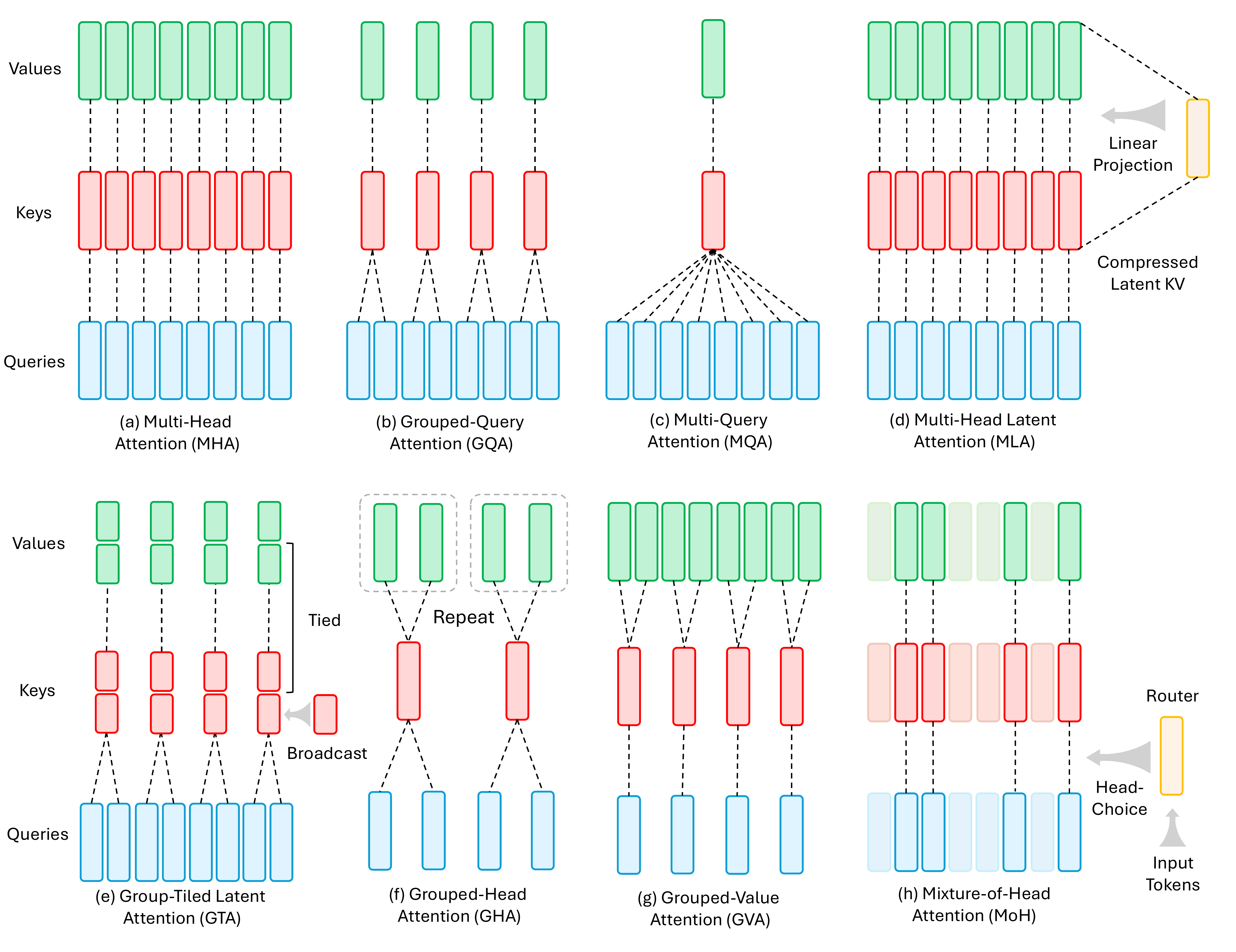}
    \caption{Mechanism Comparison of Primary Grouped Attention Methods.}
    \label{fig:grouped_attn}
\end{figure*}

\subsection{Grouped Attention}
\label{subsec:grouped attn}

{Grouped attention} techniques, including Multi-Query Attention (MQA), Grouped-Query Attention (GQA), and Multi-Head Latent Attention (MLA), have been widely adopted in the training of large language models (LLMs). These methods have been extensively validated for their ability to reduce key-value (KV) cache sizes during inference, leading to improved memory efficiency without compromising model performance.

Specifically, MQA~\cite{shazeer2019fast} was designed to address the significant memory bandwidth bottleneck encountered during the incremental decoding process in autoregressive models. By allowing multiple query heads to share a single key and value head, MQA dramatically reduces the size of the KV cache, leading to substantial improvements in inference speed. However, this efficiency sometimes comes at the cost of model quality.
To mitigate the performance degradation observed in MQA while retaining its speed advantages, Grouped-Query Attention (GQA) was proposed~\cite{ainslie2023gqa}. GQA serves as a middle ground between traditional Multi-Head Attention (MHA) and MQA. It works by dividing query heads into groups, with each group sharing a single key and value head. This approach offers a favorable trade-off, achieving quality comparable to MHA while maintaining inference speeds close to those of MQA. A key innovation in the GQA paper is the concept of "uptraining," a cost-effective method for converting existing MHA models into GQA or MQA models with only a small fraction of the original pre-training compute~\cite{ainslie2023gqa}.

Building on these efforts to optimize attention mechanisms, Multi-head Latent Attention (MLA) was introduced as part of the DeepSeek-V2~\cite{liu2024deepseekv2} and V3 models~\cite{liu2024deepseekv3}. MLA further enhances inference efficiency by compressing the KV cache into a low-rank latent vector. This technique significantly reduces the KV cache size, even more so than MQA, while maintaining strong performance~\cite{liu2024deepseekv2}. MLA's design decouples the positional information from the compressed keys and values, which is a key difference from standard attention mechanisms and allows for its aggressive compression.

Group Tied Attention (GTA) and Group Latent Attention (GLA)~\cite{zadouri2025hardware} are proposed as hardware-friendly and memory-efficient attention mechanisms aimed at addressing key inefficiencies in LLM inference, particularly those related to the KV cache. GTA builds upon the foundation of GQA by tying the Key and Value representations into a single shared projection, which is then reused across small groups of query heads. This design not only reduces the KV cache size by approximately 2× but also improves arithmetic intensity by minimizing memory movement relative to compute. A critical innovation in GTA lies in its careful handling of Rotary Position Embeddings (RoPE): only half of the tied KV state is used for the non-RoPE part of the key, while a separate smaller head computes the RoPE-associated component, which is broadcast across the group. Experimental results show that GTA preserves or even improves perplexity and task performance compared to GQA, offering a drop-in replacement with both memory and computational efficiency gains.

GLA extends the idea of latent compression in MLA by improving its compatibility with tensor parallelism. While MLA compresses the KV cache into a single latent representation, this compression limits its ability to parallelize efficiently, often requiring cache duplication across devices. GLA resolves this by sharding the latent KV cache into multiple segments, such as two halves, and assigning separate groups of query heads to each shard. Local attention is independently computed within each group and later merged, eliminating the need for full cache replication and enabling better performance in distributed and imbalanced workloads. In addition to maintaining or improving model quality over MLA, GLA demonstrates substantial speedups during inference. Specifically, its custom kernel achieves up to 2× speedup over FlashMLA in speculative decoding. Together, GTA and GLA provide highly efficient, scalable alternatives to existing attention variants, offering strong trade-offs between memory usage, compute efficiency, and parallelization capability.

\subsection{Mixture of Attention}
\label{subsec:mixture of attn}







The idea of combining multiple attention strategies into a unified attention layer began with MoA (Mixture-of-Attention)~\cite{fu2024moa}, which proposed assigning different sparse attention patterns to each head and layer within a Transformer. By automatically selecting head-wise sparsity from a pool of patterns, MoA significantly extends the effective context length while improving throughput and reducing memory consumption. This early work emphasized the benefits of heterogeneous attention structures, highlighting that not all tokens or heads require uniform computation. Expanding on this notion, SwitchHead~\cite{csordas2024switchhead} treated attention heads as experts and introduced routing mechanisms to activate only a subset of them per token, reducing attention computation by up to 8× without sacrificing performance. In parallel, MoH (Mixture-of-Heads)~\cite{jin2024moh} adopted a similar perspective, but instead of discrete routing, it used soft selection: each token computes a weighted combination of attention heads, enabling partial pruning while improving or maintaining accuracy. These works collectively mark a shift toward dynamic head-level specialization, bringing MoE-style routing into the self-attention mechanism.

Building on this head-wise mixture foundation, LLaMA-MoE v2~\cite{Qu2024LLaMAMoEVE} scaled the concept to full LLMs by sparsifying both the attention and feedforward layers through structured expert partitioning. The model achieves performance on par with dense LLaMA variants across multiple domains, demonstrating that MoE-based sparsity is viable for large-scale, instruction-tuned transformers. Further pushing flexibility, MoBA (Mixture of Block Attention)~\cite{lu2025moba} introduced block-level routing where each chunk of tokens can choose between full or sparse attention dynamically. Rather than relying on predefined patterns, MoBA allows the model to adaptively allocate computation, improving long-context performance and aligning efficiency with task demands. These advancements show the trend of increasing granularity and adaptivity in MoE routing, from heads to full blocks, making attention computation more efficient and task-aware.

Extending MoE principles beyond traditional attention layers, MoM (Mixture-of-Memories)~\cite{du2025mom} applies expert routing to memories with linear sequence modeling. Instead of maintaining a single shared state, MoM uses multiple sparsely activated memory slots, each acting as an expert updated via token-wise routing, enhancing long-term recall and reducing interference. Finally, MoSA (Mixture of Sparse Attention)~\cite{pikekos2025mixture} brings MoE routing to a fine-grained level by allowing each attention head to dynamically select the top-$k$ relevant tokens based on content. This results in a token-wise sparse structure that outperforms dense attention under equal compute budgets while significantly reducing memory usage. Across these works, the MoA paradigm evolves from fixed sparsity patterns to fully dynamic, learnable routing strategies at the head, block, memory, and token levels, offering a unified path toward more scalable, efficient, and adaptable attention mechanisms in modern large models.

\subsection{Quantized Attention}
\label{subsec:quantized attn}

To address the trade-off between computational efficiency and task accuracy in low-precision quantization, numerous hardware acceleration platforms and quantization techniques have been developed in recent years. Quantized Attention mechanisms have been proposed and studied to improve computational efficiency and reduce memory requirements, while maintaining as much of the original model accuracy as possible.

\textbf{Post-training Quantized Attention.}
Post-training quantization (PTQ) methods convert the Transformer attention operators to low-bit arithmetic without any retraining. For example, SageAttention \cite{zhang2024sageattention} quantizes the \( QK^T \) product to INT8 format (with per-channel \(\bm{smoothing}\) of outliers) while keeping the \( ({\rm softmax}(QK^T),V) \) matmul in FP16. This mixed-precision approach (INT8 for the first matmul, FP16 for the second matmul) doubles the matmul computation speed on GPUs with negligible accuracy loss. SageAttention’s Triton kernel even fuses RoPE \cite{su2024roformer} and quantization, using NVIDIA Tensor Cores to reach high efficiency. Similarly, INT-FlashAttention \cite{chen2024int} builds a FlashAttention-compatible INT8 kernel: it quantizes \( Q, K, V \) and the softmax input into 8-bit and performs the entire attention in INT8. INT-FlashAttention achieves faster inference and smaller quantization error than the FP16 baseline, with only minor quality degradation. Q-BERT \cite{shen2020q} uses Hessian-based analysis to quantize attention to very low bits; e.g., uniform 4-bit Q-BERT shows only 0.5\% accuracy drop on SQuAD versus over 11\% for naive quantization. In summary, PTQ attention methods quantize the main matmuls (often \( QK^T \) or \( ({\rm softmax}(QK^T),V) \)) and introduce scaling or smoothing to preserve fidelity, enabling plug-and-play INT8 inference.

\textbf{Quantization-Aware Training for Attention.}
Quantization-aware training (QAT) embeds low precision into training so the model learns to cope with reduced bit-width. For example, Q8BERT \cite{zafrir2019q8bert} finetunes BERT under 8-bit constraints (8-bit weights and activations) and can compress BERT by 4× with minimal loss. I-BERT \cite{kim2021bert} goes further by training the model so that every operation (including GELU and softmax) can be performed in integer arithmetic: it learns simple integer approximations for the non-linearities, allowing a RoBERTa model to run end-to-end in INT8 with near-identical accuracy. FullyQT \cite{prato2019fully} quantizes all matrix multiplies, including softmax inputs/outputs, during training, and even approximates the softmax exponent via bit-shifts. It shows that an 8-bit Transformer can match FP32 BLEU scores when properly trained.

\textbf{Mixed-Precision Attention.}
These schemes mix precisions within the attention computation to balance speed and accuracy. The SageAttention example again illustrates this: it uses INT8 for the \( QK^T \) multiplication but keeps the \( V \)-matmul in FP16. More generally, one might quantize \( Q \) and \( K \) aggressively (e.g., 8-bit) but perform the softmax/\( V \) multiplication in higher precision so that the final output remains accurate. TurboAttention \cite{kang2024turboattention}’s FlashQ is also a form of hybrid attention: it applies separate quantization to each head (one scale per head) so that most matmuls run in low precision, while still using sufficient precision to compress the KV cache. Other schemes similarly choose mixed precision by layer or token: for example, some layers might be run in INT8 while others (or certain operations like softmax normalization) remain FP16. These mixed-precision designs leverage the fact that modern GPUs can execute INT8 and FP16 ops on different tensor cores; by assigning the most sensitive sub-operations to a higher precision, they recover accuracy without sacrificing the overall throughput boost of quantization.


\textbf{INT8 Fused Attention Kernels.}
Beyond algorithmic tricks, specialized INT8 attention kernels have been developed. INT-FlashAttention \cite{chen2024int} is one example: it provides a fused CUDA kernel that takes INT8 \( Q, K, V \) and softmax inputs, performs all matmuls and softmax in 8-bit, and outputs FP16 or FP32 results. Similarly, SageAttention’s implementation fuses quantization and FlashAttention-style tiling in a Triton kernel, exploiting INT8 TensorCore MMA operations for both matmuls. HACK \cite{zhang2025hack}’s attention prefill is another fused kernel: it quantizes \( Q, K, V \) on-the-fly and feeds them into the attention computation without separate quantize/dequantize steps.

\textbf{Ultra-Low-Bit (Sub-4-Bit) Attention.}
The most aggressive quantizers push below 4 bits. SageAttention2 \cite{zhang2024sageattention2} does this by using warp-wise INT4 for the \( QK^T \) and FP8 for the \( V \)-matmul, combined with outlier smoothing on \( Q \) and \( V \). SageAttention3 \cite{zhang2025sageattention3} goes further: it uses FP4 “microscaling” for both matmuls. Specifically, it quantizes blocks of the matrices to FP4 with per-block scale factors and per-token normalization, which contains quantization error and enables $5\times$ speedup. On the training side, BitDistiller \cite{du2024bitdistiller} is a QAT framework for sub-4-bit LLMs: it employs asymmetric quantization and a self-distillation KL objective so that 3-bit and 2-bit models can approach full-precision performance. Even earlier, Q-BERT showed that carefully assigning 2-3 bits per layer (with some layers at higher precision) can yield reasonable accuracy; for example, uniform 2-bit Q-BERT retained 70\% accuracy on SQuAD versus 5\% for naive 2-bit \cite{shen2020q}.

\section{Sparse Mixture-of-Experts}
\label{sec:moe}

Sparsely-activated Mixture-of-Experts (MoE) is a cost-effective approach to enlarge the model capacity while maintaining the computational consumption~\cite{shazeer2017outrageously,lepikhin2020gshard,fedus2022switch}.
Unlike conventional model ensembling, MoE is comprised of a gate and many specialized experts.
The gate is a router that accepts input tokens and adaptively selects the most relevant experts, which brings better robustness~\cite{Chen2024MoERBenchTB}.
Besides, the gate induces specialized experts and the routing preference may be used for dataset-level representations~\cite{li2024your,zhu2024dynamic}.
Through such a sparse activation mechanism, MoE could be significantly scaled with greater model capacity, leading to better task performance~\cite{fedus2022review,pmlr-v162-du22c,pmlr-v162-clark22a,cai2025survey,mu2025comprehensive}.
Based on these advantages, the MoE paradigm has been widely applied in existing LLMs~\cite{dai2024deepseekmoe,gupta2024dbrx,xue2024openmoe,wei2024skywork,sun2024hunyuan,jiang2024mixtral,lu2024twin,yang2025qwen3,muennighoff2024olmoe}.

This section introduces \textit{gate} and \textit{experts}, which are two core components of the MoE paradigm.
Besides, we also introduce modern strategies to build MoE from dense models, which alleviate significant cost to train MoE models from scratch.

\subsection{Routing Mechanisms}
\label{subsec: routing}

The gate is a crucial component to bring sparsity in MoE models.
For a batch of input token representations $\bm{X} \in \mathbb{R}^{T \times D}$, the gate function $G$ determines the probabilities of dispatching token $x_i$ to each expert $e_i$:
\begin{gather}
    P = G(\bm{X}), \enspace P \in \mathbb{R}^{T \times N}, \quad
    G(\bm{X}) = \operatorname{Softmax}(\bm{X}\bm{W}_g + \bm{b}_g) 
    \label{eq:moe-gate}
\end{gather}
where $N$ is the number of experts, $\bm{W}_g \in \mathbb{R}^{D \times N}$ and $\bm{b}_g \in \mathbb{R}^{N}$ are trainable parameters in each MoE layer.

As shown in Eq.~(\ref{eq:moe-gate}), the most common case of gate function $G(\cdot)$ is the softmax function after a linear projection, and the calculation needs to be cast into FP32 for numerical stability especially when the model parameter is in lower precision formats.
There are some variants of $G$ to replace the linear projection with a feed-forward network (FFN) with nonlinear activations~\cite{zhu-etal-2024-llama} for better performance, or substituting softmax with sigmoid~\cite{liu2024deepseekv3,liu2025muon} or cosine similarity function~\cite{NEURIPS2022_df4f371f} to avoid representation collapse and expert competition problems~\cite{nguyen2024sigmoid}.

After obtaining the routing probabilities, MoE introduces sparsity by only selecting the top-$k$ experts.
It is also a common practice to normalize the top-$k$ routing probabilities and make the summation to be 1.0 for better convergence~\cite{jiang2024mixtral}.
\begin{equation}
    \bm{Y} = \sum_{i}^{k}\text{Top-}k\left(G(\bm{X})_i\right)\cdot E_i(\bm{X})\label{eq:moe-aggregation}
\end{equation}

\begin{figure}[t]
    \centering
    \includegraphics[width=0.7\linewidth]{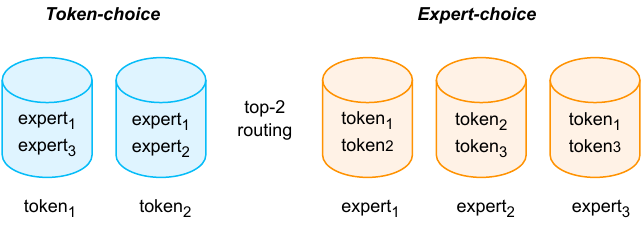}
    \caption{MoE Routing Strategies.}
    \label{fig:moe-routing-strategies}
\end{figure}

\textbf{Routing Strategies.}
Routing mechanisms in modern LLMs are usually \textbf{token-choice}-based, while a token would selects corresponding $k$ experts to calculate new representations, as illustrated in Figure~\ref{fig:moe-routing-strategies}.
A significant pitfall of this token-choice strategy is its tendency to lead poor load balancing across experts.
This often results in some experts being over-utilized and others under-utilized, leading to wasted capacity and potentially unprocessed tokens due to expert capacity limits (i.e. token dropping).

Besides token-choice, each expert could also selects top-$k$ tokens in an \textbf{expert-choice} manner.
Expert-choice~\cite{zhou2022mixture} routing simply changes the softmax operation from the $N$ dimension to $T$ dimension in Eq.~(\ref{eq:moe-gate}).
Furthermore, the top-$k$ operation in Eq.~(\ref{eq:moe-aggregation}) would select $k$ tokens for each expert, leading to perfect load balancing.
Expert-choice is effective in encoder-based models, such as T5 encoder~\cite{raffel2020exploring} and ViT~\cite{NEURIPS2021_48237d9f,Dosovitskiy2020AnII}.
However, for autoregressive language modeling, experts could not see the whole sequence at each step, and an expert may accepts too many tokens to fill all its capacity, remaining limit space for future tokens.
Therefore, modern LLMs with expert-choice has to pair an additional estimator or a loss constraint to make capacities for future tokens~\cite{muennighoff2024olmoe}.

BASE (Balanced Assignment of Experts) Layer~\cite{Lewis2021BASELS} discards the auxiliary load balancing loss in token-choice by formulating token routing as a linear assignment problem, reaching an equal number of tokens for each expert during training.
To avoid the potential future token leakage problem that is similar to expert-choice in autoregressive language modeling, a simpler greedy assignment is used when inference.

Instead of training a gate function with learnable parameters, Hash Layer~\cite{Roller2021HashLF} proposes to use a fixed predefined hash function (lookup table) to route tokens into specific experts.
To build a balanced hash function, the authors first analyze the training corpus and assign the most frequent token in the vocabulary into the emptiest bucket until all tokens are assigned.
Based on this setup, Hash Layer achieves balanced expert loads and surpasses BASE Layer.
However, due to the nature of Zipf's law in word distribution, perfect balance is nearly impossible.

\textbf{Adaptive Top-$k$ Routing.}
Besides top-$k$ routing with fixed $k$ values, recent studies also focus on dynamic routing to adaptive $k$ experts, where the number of selected experts is determined by the task difficulties or computational requirements.
Based on this mechanism, models could be both efficient and effective on complex tasks.
There are three types of adaptive routing: \textit{(1)} differentiable activation; \textit{(2)} expert activation estimation; and \textit{(3)} zero computational experts.

Differentiable activation replaces the traditional top-$k$ \& softmax router with differentiable activation functions, such as ReLU and sigmoid.
ReMoE~\cite{wang2024remoe} utilizes ReLU as the router's activation function, and experts with zero activation scores would be discarded to process the input token.
Since the router is totally differentiable, it could be optimized adaptively based on the task complexities.
However, training from scratch may induce all experts to be activated.
Therefore, ReMoE applies a regularization to ensure MoE could reach certain sparsity levels.
BlockFFN~\cite{song2025blockffn} also uses ReLU to make adaptive routing.
Instead of regularization to obtain target sparsity, it introduces chunk-level sparsity training objectives to encourage the model to be sparse in both the token-level and the chunk-level.
DynMoE~\cite{Guo2024DynamicMO} regards the expert selection process as a multi-label classification problem where each expert is an individual class.
For an input token, DynMoE calculates the cosine similarities with all expert representations and select experts with a predefined threshold.


Expert activation estimation reforms the top-$k$ routing to be directly adaptive with dynamic $k$ estimation.
MoE-Dynamic~\cite{Huang2024HarderTN} selects experts by a predefined confidence score $p$, and the gate would incorporate experts in a descending order of their selection probabilities until the cumulative score of the chosen experts exceeds $p$.
Therefore, MoE-Dynamic converts the fixed-size top-$k$ routing to dynamic top-$p$ routing, thereby dynamically allocating more computational resources to more complex inputs while conserving them for simpler ones.
Ada-K~\cite{Yue2024AdaKRB} dynamically adjusts the number of activated experts for each token in MoE-based LLMs through a learnable allocator.
This allocator module is similar to the gate and takes the input token and outputs a probability distribution over the number of experts, from which an optimal count is sampled.

Zero computational experts introduces heterogeneous experts with zero or few computational costs as placeholders.
MoE++~\cite{Jin2024MoEAM} keeps the number of activated experts $k$ unchanged, and devises zero computational experts.
By introducing heterogeneous load balancing across normal experts and these zero computational experts, some tokens would skip the computation process and therefore improve the overall efficiency.
AdaMoE~\cite{Zeng2024AdaMoETR} achieves token-adaptive routing by introducing \textit{null} experts alongside the standard experts in MoE, where these null experts perform no computations.
Similar to MoE++, the router still selects a fixed number of $k$ experts, where it is larger than the $k$ in a comparable vanilla MoE.
This allows different tokens to dynamically use a different number of true experts based on their needs.
For instance, semantically rich tokens might engage more true experts, while less significant tokens might be routed to more null experts, thus achieving the adaptive routing.

\textbf{Load Balancing.}
\textit{Gate load} is the number of tokens routed to each expert.
As discussed in \textbf{Routing Strategies}, token-choice routing is prone to be unbalanced due to biased gates, where an expert may be selected by many tokens and other experts may process less tokens.
This phenomenon would make MoE collapses where some experts are not sufficiently trained.
Furthermore, unbalanced routing would result in slow training because experts with less tokens may \textit{wait} for other experts to finish processing.
Given the fact that modern language models prefer token-choice routing for effective autoregressive token prediction, it is crucial for MoE to reach balancing.

A common practice for addressing unbalanced routing is to add an auxiliary load balancing loss $\mathcal{L}_{\text{aux}}$ as a soft balance constraint, and the final optimization object of LLM training would be the summation of a cross-entropy loss $\mathcal{L}_{\text{ce}}$ for language modeling and the auxiliary loss $\mathcal{L}_{\text{aux}}$ for load balancing.
Shazeer et al.~\cite{shazeer2017outrageously} proposes to apply the coefficient of variance (CV) of gate importance scores (i.e. routing probabilities) as an auxiliary loss, where a more balanced training would lead to lower CV scores.
For a stricter balance constraint, the gate load $D(\bm{X})$ is also estimated for the back-propagation compatibility:
\begin{gather}
    \mathcal{L}_{\text{aux}} = \alpha \operatorname{CV}(\sum G(\bm{X}))^2 + \beta \operatorname{CV}(\sum D(\bm{X}))^2,
\end{gather}
where $\alpha$ and $\beta$ are hyper-parameters.

GShard~\cite{lepikhin2020gshard} provides a simplified version of auxiliary loss that considers both the gate load and the gate importance scores:
\begin{gather}
    \mathcal{L}_{\text{aux}} = \frac{1}{N} \sum_{i=1}^{N} D_i(\bm{X}) \cdot G_i(\bm{X})
\end{gather}
This design is prevalent in MoE and has become a default setting in many popular LLMs~\cite{dai2024deepseekmoe,jiang2024mixtral}.

Although such auxiliary losses are effective for reaching load balancing, it may violate the optimization direction due to the interference gradient.
To push the language modeling's performance upper bound, Wang et al.~\cite{Wang2024AuxiliaryLossFreeLB} introduces a method to discard the auxiliary loss while maintaining the load balancing.
The implementation involves applying an expert-wise bias to the routing scores of each expert before the top-$k$ routing decision is made.
This bias is not static and it is dynamically updated for each expert based on its recent load.
This iterative process of token routing and bias updating allows the model to maintain a balanced distribution of expert load consistently throughout training by directly adjusting gating scores rather than introducing an auxiliary loss term, thereby avoiding interference with the primary language modeling objective.

Qiu et al.~\cite{Qiu2025DemonsIT} addresses the issue of inhibited expert specialization in MoE models when using traditional load balancing loss (LBL) calculated at the micro-batch level.
The authors argue that micro-batch load balancing loss forces tokens, even domain-specific ones, to be uniformly routed to all experts within each small sequence, hindering the ability of experts to specialize.
To solve this, they propose a global-batch load balancing loss by introducing an extra communication step to synchronize the expert load $D(\bm{X})$ across all parallel micro-batches that form a global batch.
The load balancing loss is then calculated using these globally aggregated gate load.
This approach aims to loosen the strict constraint of micro-batch LBL, encouraging load balance at the broader corpus level, thereby fostering better domain specialization among experts and improving overall model performance in both pretraining perplexity and downstream tasks.

\subsection{Expert Architectures}
\label{subsec: expert}

Traditional MoE introduces sparsity in FFNs and the experts are composed of multiple FFN blocks~\cite{fedus2022switch}.
Although FFN experts are effective in most cases, there are still rooms for improving the efficiency, robustness, and training stability~\cite{Zoph2022STMoEDS}.
Besides, experts are specialized in a pretrained MoE model, where each expert may process tokens in certain field or have similar patterns.
This also innovates new Parameter-Efficient Fine-Tuning (PEFT) paradigms to only finetune task specific experts for lower computational cost~\cite{wang2024let}.

\begin{figure}[t]
    \centering
    \includegraphics[width=0.7\linewidth]{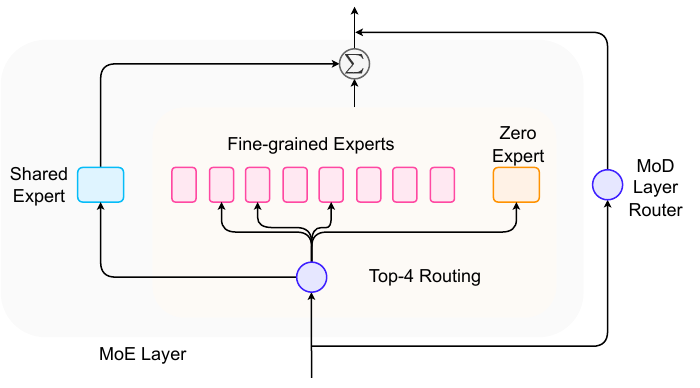}
    \caption{MoE Expert Architectures.}
    \label{fig:moe-expert-arch}
\end{figure}

As shown in Figure~\ref{fig:moe-expert-arch}, there are multiple expert architectures in MoE layers, including fine-grained experts, shared experts, MoD experts, and other heterogeneous experts (e.g. zero experts).
In this section, we introduce different types of expert architectures in FFN blocks.
For attention experts, please refer to \S~\ref{subsec:mixture of attn}.

\textbf{Fine-grained Experts.}
For an MoE layer with $N$ experts and each expert has a intermediate size of $M$, there are $NM$ neurons in total.
While keeping the number of total parameters unchanged, the fine-grained MoE has smaller $M$ values, leading to larger $N$ (i.e. more small experts).
Fine-grained experts bring more combinations and better performance with a clear scaling property~\cite{muennighoff2024olmoe}.
For instance, a 16-top-2 setting only has $\binom{16}{2}=120$ expert selection combinations, while 64-top-8 has $\binom{64}{8}=4.4\times10^9$ combinations.
Such a fine-grained setting has been verified to be effective in modern LLMs like DeepSeekMoE~\cite{dai2024deepseekmoe}, Qwen-MoE~\cite{yang2025qwen3}, and OLMoE~\cite{muennighoff2024olmoe}.
However, more experts would bring additional overhead in routing and bring new challenges in efficient training/inference framework designs~\cite{Krajewski2024ScalingLF}.

\textbf{Shared Experts.}
Shared experts (or residual experts) are fixed experts where tokens are always routed to.
DeepSpeed-MoE~\cite{Rajbhandari2022DeepSpeedMoEAM} utilizes another linear projection and softmax to make a weighted sum of the shared experts and the vanilla experts.
Qwen2-MoE~\cite{Yang2024Qwen2TR,Yang2024Qwen25TR} applies a similar linear projection with sigmoid to calcualte the contribution weight of shared experts and sum two types of experts' outputs as the final output.
DeepSeekMoE~\cite{dai2024deepseekmoe} directly adds the shared experts' and the vanilla experts' outputs without complex re-weighting.
These implementations all show performance improvements in the language modeling perplexities or downstream task performance.

\textbf{Mixture-of-Depths.}
Besides the above horizontally distributed experts, experts could also be vertically distributed from the layer perspective.
Mixture-of-Depths (MoD)~\cite{Raposo2024MixtureofDepthsDA} regards transformer layers as experts and select top-$k$ tokens for each layer with a corresponding router.
Thus it significantly reduces the computational cost by consuming less tokens on each layer.
Different from LayerSkip~\cite{Elhoushi2024LayerSkipEE}, tokens dropped from former layers may be processed by latter layers.
MoD is orthogonal with MoE and could be combined with MoE as Mixture-of-Depths-and-Experts (MoDE).

\textbf{Other Special Experts.}
Besides the conventional modules that could be constructed as experts, there are also special variants that integrate the idea of sparse MoE.
SoftMoE~\cite{Puigcerver2023FromST} mixes tokens into soft slots, applies experts to process such slots, and utilizes output projectio to reveal token representations.
MoE++~\cite{Jin2024MoEAM} introduces zero, copy, and constant experts in MoE to adaptively control the top-$k$ routing and the overall computational cost.
ModuleFormer~\cite{Shen2023ModuleFormerLM} provides a scheduled expert expansion method to add new specialized experts while maintaining the original knowledge.
Inspired by the parameter-efficient fine-tuning methods, experts could also be LoRA weights for lightweight design~\cite{Dou2024LoRAMoEAW,Luo2024MoELoRACL,Wu2024MixtureOL,fan2025make}, and the number of experts could be scaled to 1 million with a specially designed expert retrieval strategy~\cite{He2024MixtureOA}.


\subsection{MoE Conversion}
\label{subsec:moe_conversion}

\begin{figure}[t]
    \centering
    \includegraphics[width=0.75\textwidth]{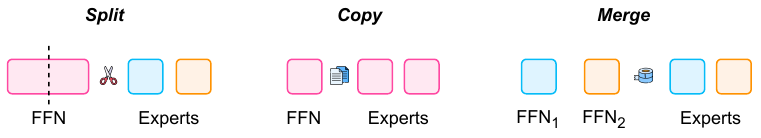}
    \caption{MoE Conversion Strategies.}
    \label{fig:moe-conversion}
\end{figure}

Training MoE models from scratch costs a mass of computational resources.
Therefore, it is worth constructing MoE models from existing dense models to efficiently scale the model capacity or reduce the inference cost.
As shown in Figure~\ref{fig:moe-conversion}, there are three main strategies:
\textit{(1)} Splitting existing FFN to construct experts (MoEfication~\cite{Zhang2021MoEficationTF});
\textit{(2)} Copying FFNs and aggregate as an expert (Sparse Upcycling~\cite{Komatsuzaki2022SparseUT,zhang2024clip});
\textit{(3)} Merging two existing dense models as an MoE model (BTX~\cite{Sukhbaatar2024BranchTrainMiXME}).
Here we introduce these strategies in two aspects: from existing ``dense to MoE'', and ``sparse model routing'' to aggregate two dense models as an MoE model.

\textbf{Dense to MoE.}
MoEBERT~\cite{Zuo2022MoEBERTFB} builds MoE by splitting encoder-based BERT-family models and proposes to use importance scores~\cite{Molchanov2019ImportanceEF} for guiding the model adaptation.
MoEfication~\cite{Zhang2021MoEficationTF} construct MoE by splitting T5 models from a co-activation graph and the results are very close to the original T5 model.
LLaMA-MoE~\cite{zhu-etal-2024-llama,Qu2024LLaMAMoEVE} partition FFNs in LLaMA~\cite{Touvron2023Llama2O,Dubey2024TheL3} to build MoE models with continual pretraining and supervised fine-tuning.
Besides splitting dense FFNs to sparse experts, Komatsuzaki et al.~\cite{Komatsuzaki2022SparseUT} propose to scale the model capacity by copying current FFNs and initialize new gates for each MoE layer, namely sparse upcycling.
For vertical moe conversion, DLO~\cite{Tan2024DLODL} expands layers and build MoD experts.
MoDification~\cite{Zhang2024MoDificationMO} converts layers in an existing dense model to MoD-style experts with a threshold-$p$ gate, where tokens exceed a predefined threshold are processed.

\textbf{Sparse Model Routing.}
Besides module-based MoE, models could be merged or aggregated to build MoE-like models for better comprehending user instructions and generating responses with specialized experts.
Branch-Train-Merge (BTM)~\cite{Li2022BranchTrainMergeEP} proposes to train several expert language models (LMs) with domain specific datasets.
During inference, these expert models are weighted averaged by the input query's domain category and a new model would generate the response.
Instead of model merging, Branch-Train-Mix (BTX)~\cite{Sukhbaatar2024BranchTrainMiXME} directly aggregates expert LMs' FFNs into MoE layers and train a new gate for each layer to dynamically select specialized experts.


\section{Hybrid Architectures}
\label{sec:hybrid}
As the core component of the Transformer, softmax attention has demonstrated its superior performance across a wide range of tasks. However, its quadratic computational complexity, coupled with the growing KV-cache overhead, has become a critical barrier to efficiency in long-sequence processing. In contrast, linear models have achieved significant breakthroughs in computational efficiency. Nevertheless, extensive studies have shown that these linear models often fall short of standard softmax attention in tasks requiring precise recall, sparse information retrieval, and long sequence modeling. In response to these challenges, researchers have been developing hybrid models to achieve a better balance between efficiency and performance.

\begin{figure*}[t]
    \centering
    \includegraphics[width=0.9\linewidth]{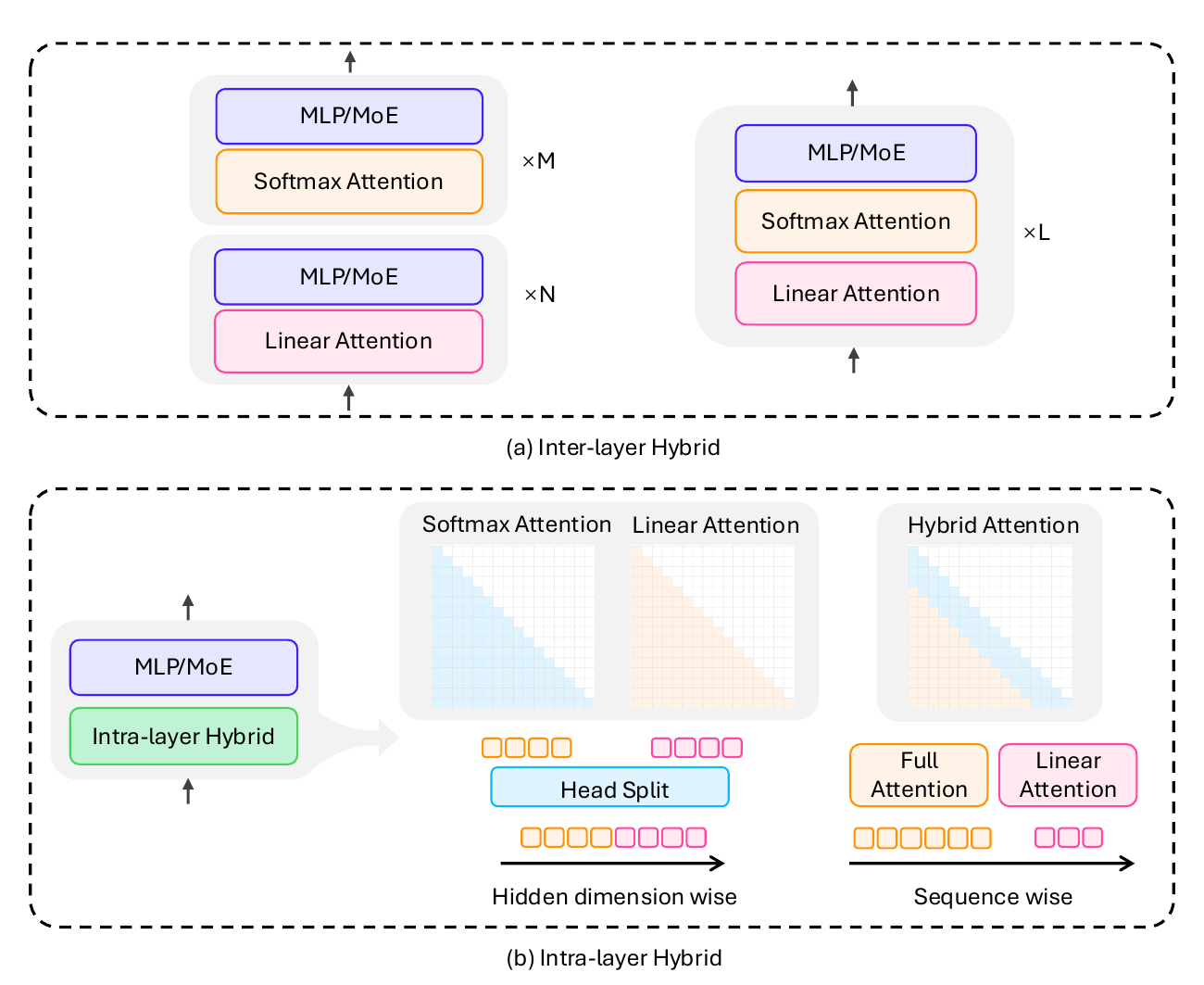}
    \caption{\text{Hybrid Model Architectures.} (a) illustrates the classical paradigm of inter-layer hybrid approach. (b) demonstrates the classical paradigm of intra-layer hybrid approach. The left side of (b) represents a pattern similar to Hymba~\cite{hymba}, which employs head-wise partitioning with either softmax attention or linear attention. The right side, depicts a pattern analogous to LoLCATs~\cite{zhang2024lolcats}, featuring sequence-wise partitioning where local regions utilize softmax attention while distant regions employ linear attention.}
    \label{fig:hybrid_model}
\end{figure*}

The core challenge of hybrid models lies in efficiently and effectively integrating the efficiency benefits of linear models and the expressive power of standard softmax attention. As shown in Figure~\ref{fig:hybrid_model}, current hybrid models can be categorized into two main categories based on the integration approach: 
\begin{itemize}[leftmargin=*]
    \item \textbf{Inter-layer Hybrid} employs a hierarchical alternating approach, where softmax attention layers are inserted at specific intervals between consecutive linear sequence modeling layers. This approach ensures that the majority of layers retain linear computational complexity while still benefiting from the high representational capability of softmax attention layers. 
    \item \textbf{Intra-layer Hybrid} achieves fine-grained fusion within individual layers by strategically blending linear sequence modeling layers with softmax attention transformer layers. 
\end{itemize}

\subsection{Inter-layer Hybrid}
\label{sec:hybrid_inter}
The inter-layer hybrid model typically combines softmax attention layers and linear sequence modeling layers in a predefined proportion. Due to its practical effectiveness and straightforward implementation, it has become a dominant approach in the development of hybrid architectures.

Many notable hybrid models have been built on the Mamba~\cite{gu2023mamba} architecture. 
Zamba~\cite{zamba} is a 7B-parameter hybrid language model built primarily on the Mamba with periodic insertions of a single global shared self-attention block. Zamba demonstrates competitive accuracy on general language understanding benchmarks, approaching leading models like Mistral~\cite{jiang2023mistral} despite using significantly fewer training tokens. 
Compared to Zamba, Zamba2~\cite{glorioso2024zamba2} introduces key improvements including a switch to Mamba2~\cite{dao2024transformers}, two alternating shared attention blocks and non-shared LoRAs. These changes lead to better performance and efficiency. Jamba~\cite{jamba} introduces a more expressive and scalable 52B hybrid architecture. 
Jamba adopts a hybrid architecture combining Mamba, standard attention, and MoE, interleaved at a 7:1 ratio. This design enables Jamba's strong in-context learning, and efficient inference, achieving $3\times$ higher throughput than Mixtral and supporting 256K context with only 4GB KV cache. 
Samba~\cite{samba} combines Mamba and Sliding Window Attention. Samba avoids full attention and adopts linear-time mechanisms, enabling it to extrapolate up to 1 million tokens in zero-shot and recall up to 256K tokens with minimal fine-tuning.
Mamba-in-Llama~\cite{wang2024mamba} distills pretrained Transformers into Mamba, and proposes a hybrid version of standard attention and Mamba. The method leverages weight reuse from Transformer attention layers to initialize Mamba blocks, followed by a multi-stage distillation pipeline to preserve the capabilities of standard attention layer.
Hunyuan-TurboS~\cite{liu2025hunyuan} is a 56B activated (560B total) hybrid model that combines Mamba2, standard attention, and MoE layers to balance long-sequence efficiency and contextual reasoning. It uses an interleaved Attention-Mamba2-FFN/Mamba2-FFN block structure supports a 256K context length. 
Zebra-Llama~\cite{yang2025zebra} proposes an efficient hybrid architecture combining Mamba2 and MLA~\cite{liu2024deepseekv2} to achieve Transformer-level performance with minimal overhead. The method employs SVD-based initialization and layer distillation to transfer knowledge from pre-trained models, along with a sensitivity-aware layer replacement strategy for optimal efficiency.

In addition to Mamba, numerous works combine standard attention with linear attention, linear RNNs, and sliding window attention. RWKV-X~\cite{rwkv-x} strategically interleaves sparse attention MoBA~\cite{lu2025moba} within RWKV-7~\cite{peng2025rwkv} layers, with about $25\%$ of layers using sparse attention, achieving strong performance across both short- and long-context tasks. It demonstrates superior decoding stability and lower latency than full-attention models at long sequence lengths. 
YOCO~\cite{sun2024yoco} combines sliding-window attention with standard attention. It features shared KV cache across layers, where the self-decoder generates a single KV cache that is reused by the cross-decoder layers. YOCO reduces memory usage and improves efficiency, especially for long-context tasks, by caching only once for all layers.
RecurrentGemma~\cite{botev2024recurrentgemma} combines Griffin~\cite{de2024griffin} with sliding window attention, avoiding global attention entirely. This design enables RecurrentGemma to match the performance of similarly sized transformer models like Gemma, while offering significantly better memory and inference efficiency.
MiniMax-01~\cite{li2025minimax} is a 456B-parameter MoE hybrid Model that integrates Lightning Attention with standard softmax attention to handle ultra-long sequences up to 4M tokens efficiently.  
LaCT~\cite{zhang2025test} combines large-chunk TTT with local window attention to efficiently model long sequences while optimizing hardware utilization. This hybrid design enables linear-complexity global context modeling with quadratic costs only locally, supporting tasks like novel view synthesis with 1M tokens and autoregressive video diffusion.

\subsection{Intra-layer Hybrid}
\label{sec:hybrid_intra}
Intra-layer hybrid architectures combine linear and standard attention within a single layer. Typical designs include: (1) head-wise split, where different heads use either linear or standard attention and (2) sequence-wise split, applying different attention types to different input segments.

\textbf{Head-wise split.} Hymba~\cite{hymba} represents a classic head-wise hybrid architecture, where the attention heads are partitioned into two subsets: one employing Mamba and the other utilizing standard softmax attention. Additionally, it incorporates innovations such as learnable meta tokens for memory initialization, sliding window attention with selective global attention, and cross-layer KV cache sharing. 1.5B Hymba outperforms Llama-3.2-3B across reasoning/recall tasks while achieving $3.49\times$ higher throughput and $19\times$ smaller cache size. 
WuNeng~\cite{wuneng} combines RWKV-7 state-driven mechanisms with Transformer attention, balancing high-resolution recall and efficiency through cross-head interaction. WuNeng outperforms LLaMA and Hymba, achieving better results in complex reasoning and long-context tasks. 

\textbf{Sequence-wise split.} LoLCATs~\cite{zhang2024lolcats} is a sequence-level hybrid model that replaces softmax attention with a combination of linear attention and SWA. It uses linear attention for all earlier tokens and local softmax for the most recent tokens, capturing local dependencies. The model is trained in two steps: (1) attention transfer to match softmax outputs, and (2) LoRA to fine-tune the replaced layers. With less than $0.2\%$ parameter updates and just 0.04B tokens used, LoLCATs successfully linearizes models up to 405B parameters, outperforming previous hybrid and subquadratic models in both efficiency and quality.
Liger~\cite{lan2025liger} transforms pretrained Transformer-based LLMs into gated linear recurrent structures by reusing key projection weights to build gate matrix—avoiding added parameters and preserving efficiency. Liger blends Linear Attention with Sliding Window Attention, enabling softmax-like expressivity with linear-time inference and constant memory. Supporting models from 1B to 8B parameters, Liger recovers up to $93\%$ of original model performance using only 0.02B fine-tuning tokens via LoRA. 
TransMamba~\cite{transmamba} is a sequence-level hybrid model that dynamically combines standard softmax attention and Mamba state space modeling using shared parameters and a learned TransPoint to switch between them. Early tokens are processed with standard softmax attention for precision, while later tokens use Mamba for efficient long-range modeling, facilitated by a lossless Memory Converter. Supporting models up to 1.5B parameters, TransMamba achieves up to $25\%$ faster training than Transformers and excels in long-context tasks. 
LoLA~\cite{mcdermott2025lola} is developed to overcome the limitations of conventional linear attention models in long-context tasks. By integrating three complementary memory systems, low-rank linear attention for efficient global token storage, sliding window attention for precise local context modeling, and a sparse global cache for retaining high-fidelity representations of interference-prone key-value pairs.




\section{Diffusion Large Language Models}
\label{sec:diffusion}

In the preceding sections, we discussed various Transformer-based efficient architectures for LLMs, focusing primarily on attention mechanisms and MoE modules. These approaches improve efficiency by optimizing computation within the standard autoregressive (AR) framework. However, despite their advantages, they remain constrained by the sequential nature of AR generation, where tokens are produced one at a time. This token-wise dependency requires as many forward passes as the sequence length, representing a fundamental bottleneck for inference latency.

In this section, we introduce recent advances in diffusion large language models (Diffusion LLMs)~\cite{yu2025discrete,sahoo2024simpleeffectivemaskeddiffusion}. 
In contrast to autoregressive models, Diffusion LLMs generate text by progressively denoising a sequence from a noisy or masked state to a coherent output. 
This fundamental shift in generation enables several unique advantages. Most notably, Diffusion LLMs support parallel decoding, allowing multiple tokens to be produced at each refinement step, which significantly reduces inference latency by avoiding sequential token generation. 
Moreover, the formulation of text generation within Diffusion LLMs as a denoising or infilling process over a fixed-length canvas inherently provides superior controllability.
This allows the model to better adhere to specific output constraints, such as length, format, or structure, challenges that are difficult to address with standard autoregressive methods. 
Finally, Diffusion LLMs utilize bidirectional attention, enabling the model to access and revise context across the entire sequence at every step. This global view helps mitigate issues such as the reversal curse, which arise from the unidirectional nature of autoregressive decoding.

\begin{figure*}
    \centering
    \includegraphics[width=0.7\linewidth]{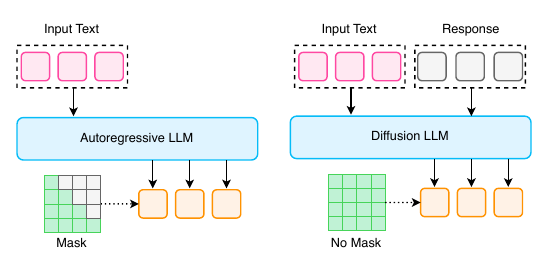}
    \caption{Mechanism Comparison of Autoregressive Models and Diffusion LLMs.}
    \label{fig:structures}
\end{figure*}

\subsection{Non-Autoregressive Diffusion LLM}
\label{subsec:diff_non-auto}

LLaDA~\cite{nie2025largelanguagediffusionmodels} is a non-autoregressive Diffusion LLM with 8 billion parameters, trained from scratch. 
It models the data distribution through a forward process that progressively masks tokens and a reverse denoising process that simultaneously predicts all masked tokens per step. 
Empirically, LLaDA exhibits strong scalability and achieves performance on par with leading AR models such as Llama3-8B across diverse benchmarks, including in-context learning and instruction-following after supervised fine-tuning (SFT). 
Furthermore, it overcomes known failure modes of AR models, such as the ``reversal curse'', attributed to their unidirectional processing.

In this section, we formalize the LLaDA's probabilistic framework for better understanding. 
The model distribution \( p_\theta(x_0) \) is characterized by a predefined forward corruption process and a learned reverse generative process. The forward process systematically transforms a clean sequence \( x_0 \) into a corrupted intermediate \( x_t \) by masking tokens with an independent probability dictated by a timestep \( t \in [0, 1] \). 
The reverse process is trained to recover the original data by iteratively denoising \( x_t \), progressively predicting the masked tokens as \( t \) is annealed from 1 to 0.

Central to this framework is a parametric mask predictor, \( p_\theta(\cdot|x_t) \), which operates on a corrupted sequence \( x_t \) to jointly infer the entire set of masked (\textrm{M}) tokens. 
Its parameters \( \theta \) are optimized by minimizing the following cross-entropy objective, computed exclusively over the masked positions:
\begin{align}
\label{eq:objective}
   \mathcal{L}(\theta)  \triangleq   -  \mathbb{E}_{t, x_0,  x_t} \left[\frac{1}{t} \sum_{ i = 1 }^L \textbf{1}[x_t^i = \textrm{M}] \log p_{\theta}(x_0^i|x_t) \right] , 
\end{align}
where \( x_0 \) is a sample drawn from the data distribution, \( t \) is a timestep uniformly sampled from the interval \( [0, 1] \), and \( x_t \) denotes the resulting corrupted sequence.

Upon training, sampling from \( p_\theta(x_0) \) is performed by simulating the reverse process. 
Crucially, the training objective \( \mathcal{L}(\theta) \) is a variational upper bound on the model's negative log-likelihood:
\begin{align}
\label{eq:bound}
    - \mathbb{E}_{p_{\textrm{data}}(x_0)} \left[\log p_\theta(x_0) \right]  \le  \mathcal{L}(\theta),
\end{align} 
making it a principled objective for density estimation.

Unlike masked language models such as BERT~\cite{devlin2019bert} that employ a fixed masking ratio, LLaDA's use of a random ratio is fundamental to its design. 
This distinction, theoretically grounded by the negative log-likelihood bound in Eq.~\eqref{eq:bound}, establishes LLaDA as a true generative model. 
This principled foundation enables emergent capabilities like \emph{in-context learning}, positioning LLaDA as a viable, non-autoregressive alternative to mainstream LLMs.


Whether Diffusion LLMs can match the reasoning prowess of Reinforcement Learning (RL)-enhanced AR models has been a significant open problem. The d1 framework~\cite{zhao2025d1scalingreasoningdiffusion} provides a compelling affirmative answer. It adapts pre-trained masked Diffusion LLMs for complex reasoning via a two-stage process: first, SFT on reasoning traces, followed by a novel RL algorithm, diffu-GRPO. 
This policy-gradient algorithm is tailored for the non-autoregressive nature of Diffusion LLMs, featuring an efficient one-step log-probability estimation regularized via random masking prompt. 
Applied to the LLaDA-8B-Instruct model, d1 achieves substantial gains on mathematical and logical reasoning benchmarks, surpassing not only the base model but also rigorous SFT- and RL-only ablations. This work provides strong evidence that Diffusion LLMs can be potent reasoners when augmented with RL, establishing them as a competitive architecture in a domain previously dominated by AR models.


\subsection{Bridging Diffusion LLM and Autoregressive}
\label{subsec:diff_semi-auto}

Diffusion LLMs offer significant advantages over their AR counterparts, namely parallelized generation and greater controllability. Nevertheless, they face challenges in likelihood modeling and are inherently constrained to fixed-length outputs.
A significant thrust in recent Diffusion LLM research is to combine the strengths of both AR and diffusion methods.
BD3-LMs \cite{arriola2025blockdiffusioninterpolatingautoregressive} interpolates between discrete denoising diffusion and AR models by defining an autoregressive distribution over blocks of tokens, while performing diffusion within each block. 
By synergizing diffusion and autoregressive paradigms, this hybrid approach remedies the fixed-length constraint of the former and the high inference latency of the latter. Specifically, BD3-LMs integrate KV caching from AR models with parallel in-block token sampling, thereby enabling flexible-length generation and substantially accelerating inference.
They further propose an efficient training algorithm to tackle the key challenge of gradient variance estimation in Diffusion LLMs, complemented by data-driven noise schedules engineered to minimize this variance.


Formally, BD3-LMs operates on a sequence $x = (x_1, \ldots, x_L)$ by partitioning it into $B$ non-overlapping blocks of length $L'$, denoted $x = (x^{(1)}, \ldots, x^{(B)})$. The model defines the joint log-likelihood through an autoregressive factorization over these blocks:
\begin{equation}
\log p_\theta(x) = \sum_{b=1}^{B} \log p_\theta(x^{(b)} \mid x^{(<b)}),
\end{equation}
where each conditional distribution $p_\theta(x^{(b)} \mid x^{(<b)})$ is modeled by a discrete diffusion process. This process learns a denoising model $p_\theta(x^{(b)} \mid x^{(b)}_t, x^{(<b)})$ to reverse a fixed forward noising process $q(\cdot \mid \cdot)$, according to the reverse-time Markov chain:
\begin{align}
p_\theta(x^{(b)}_s \mid x^{(b)}_t, x^{(<b)}) = \sum_{x^{(b)}} q(x^{(b)}_s \mid x^{(b)}_t, x^{(b)}) \, p_\theta(x^{(b)} \mid x^{(b)}_t, x^{(<b)})
\end{align}
The denoising function is parameterized by a single transformer, $f_\theta$, which employs a block-causal attention mask to enforce the autoregressive dependency between blocks. For a given corrupted block $x^{(b)}_t$ and its causal context $x^{(<b)}$, the model predicts the original clean block $\hat{x}^{(b)}_0$:
\begin{equation}
f_\theta\left(x^{(b)}_t, x^{(<b)}\right) \rightarrow \hat{x}^{(b)}_0
\end{equation}
Inference proceeds autoregressively across blocks while leveraging parallel decoding within each block. This structure naturally accommodates KV caching for all preceding blocks $x^{(<b)}$, significantly improving efficiency.

The training objective is a variational upper bound on the negative log-likelihood, constructed by aggregating the diffusion loss from each individual block:

\begin{equation}
- \log p_\theta(x) \leq \mathcal{L}_{\mathrm{BD}}(x; \theta) := \sum_{b=1}^{B} \mathcal{L}(x^{(b)}, x^{(<b)}; \theta),
\end{equation}
where each term $\mathcal{L}(x^{(b)}, x^{(<b)}; \theta)$ is a standard objective for discrete diffusion, adaptable to various formulations such as continuous-time or masking-based processes.










From another perspective, DiffuLLaMA~\cite{gong2025scalingdiffusionlanguagemodels} pioneers an alternative approach that leverages the abundance of pre-trained AR models. Their method adapts foundational AR models, such as GPT-2 and LLaMA (spanning 127M to 7B parameters), into text diffusion architectures named DiffuGPT and DiffuLLaMA. Grounded in a demonstrated connection between AR and diffusion objectives, this conversion is achieved via a simple continual pre-training strategy. The resulting models outperform prior Diffusion LLMs and achieve performance competitive with their AR counterparts, despite requiring significantly less training data than training from scratch. This adaptation also enables the Diffusion LLMs to inherit capabilities like in-context learning and instruction following, and they naturally excel at non-sequential tasks such as "fill-in-the-middle" without prompt reordering, leveraging the inherent nature of the diffusion process.



\subsection{Extending Diffusion LLM to Multimodality}
\label{subsec:dllm multimodal}

Recent advancements have extended Diffusion LLMs into the more complex multimodal domain \cite{liu2023visual,bai2025qwen2,su2025openthinkimg,su2025thinking,chen2025advancing}. 
These multimodal Diffusion LLMs are engineered to operate on joint textual and visual data by incorporating a vision encoder and a projection layer, which maps extracted visual features into the language model's latent embedding space.

In this emerging area, several distinct approaches have emerged. 
LLaDA-V \cite{you2025lladavlargelanguagediffusion} introduces a purely diffusion-based Multimodal LLM architecture, utilizing a vision encoder and an MLP connector that maps the resulting visual representations into the language model's latent space.
This strategy enables effective multimodal alignment, primarily achieved through visual instruction tuning. The model demonstrates notable multimodal performance and substantial data scalability.
Similarly, UniDisc \cite{swerdlow2025unified} advocates for a unified generative paradigm across both textual and visual domains, using purely discrete diffusion. 
It builds upon a shared architecture that jointly tokenizes text and images into a common vocabulary and uses full self-attention. UniDisc demonstrates strong joint multimodal inpainting and zero-shot editing capabilities without explicit optimization for these tasks.
In a similar vein, LaViDa \cite{li2025lavida} equips discrete diffusion models with a vision encoder but specifically focuses on overcoming practical challenges in adapting diffusion models for vision-language tasks. It introduces novel techniques such as ``complementary masking'' to improve training data efficiency by ensuring all output tokens contribute to learning, ``Prefix-Diffusion LLM decoding'' to enable KV caching for efficient inference with long multimodal prompts, and ``timestep shifting" to enhance sample quality, particularly at reduced generation steps, thereby offering the unique advantages of diffusion models like speed-quality tradeoff and controllability.

To address the inherent instabilities and suboptimal performance associated with purely discrete diffusion training in multimodal settings, Dimple \cite{yu2025dimplediscretediffusionmultimodal} proposes an innovative hybrid ``Autoregressive-then-Diffusion'' training paradigm. 
This approach first utilizes an autoregressive phase for robust vision-language alignment and instruction following. 
It is then succeeded by a transition to a diffusion-based masked language modeling stage, devised to reinstate parallel decoding capabilities.
For inference, Dimple introduces ``Confident Decoding'' for dynamic modulation of tokens generated per step, and ``Structure Priors", enabling fine-grained control over response length and format. 
Consequently, it achieves performance levels on par with prominent autoregressive baselines like LLaVA-NEXT \cite{li2024llavanext-ablations}.
Further pushing the boundaries of unified modeling, MMaDA \cite{yang2025mmadamultimodallargediffusion} pioneers a novel category of multimodal diffusion foundation models, distinguished by a common probabilistic framework and a modality-agnostic design. 
This architecture inherently eliminates the requirement for discrete modality-specific components and is synergistically enhanced by a ``mixed long chain-of-thought (CoT)'' fine-tuning methodology. 
This methodology establishes a consistent CoT structure across diverse modalities, thereby enabling cold-start RL. Moreover, MMaDA introduces UniGRPO, a novel unified RL algorithm optimized for diffusion models, which exhibits remarkable generalization across tasks, including textual reasoning, multimodal comprehension, and even text-to-image synthesis.



\section{Applications to Other Modalities}
\label{sec:application}
Originally developed and popularized for their success in the text domain, efficient architectures are now being broadly adapted to other domains. This section surveys the transfer of these powerful paradigms, namely linear-time sequence models like SSMs and RWKV-like architectures, and sparse computation strategies like MoE, to non-textual data. We explore their impact across several key modalities, beginning with their extensive applications in computer vision (\S\ref{subsec:vision}). We then examine their growing influence in the audio domain (\S\ref{subsec:audio}). Finally, we discuss their crucial role in multimodal learning (\S\ref{subsec:multimodality}), where efficiency is paramount for integrating diverse data streams. This expansion beyond text demonstrates the versatility of these models, unlocking new capabilities and enabling scaling to previously intractable problem sizes across the broader landscape of artificial intelligence.

\begin{table}[ht]
\scriptsize
\centering
\caption{Overview of Applications of Efficient Architectures Across Modalities and Categories.}
\label{tab:application}
\begin{tabular}{ccp{11.3cm}}
\toprule
\textbf{Modality} & \textbf{Category} & \textbf{Approach} \\
\midrule
\multirow{8}{*}{\raisebox{-18ex}[0pt][0pt]{Vision}}
& \raisebox{-1ex}[0pt][0pt]{Classification} & InsectMamba~\cite{wang2025insectmamba}, V-MoE~\cite{NEURIPS2021_48237d9f}, MoE-CNN~\cite{zhang2023robust}, pMoE~\cite{chowdhury2023patch}, Res-vmamba~\cite{chen2024res}, Mammil~\cite{fang2024mammil}, Spectralmamb~\cite{yao2024spectralmamba}, Memorymamba~\cite{wang2024memorymamba} \\\cmidrule{2-3}
& \raisebox{-1ex}[0pt][0pt]{Detection} & Vig~\cite{liao2025vig}, Vision-rwkv~\cite{duan2024vision}, Tutel~\cite{hwang2023tutel}, Mamba yolo~\cite{wang2024mambayolo}, Voxel mamba~\cite{zhang2024voxel}, Mim-istd~\cite{chen2024mim}, Htd-mamba~\cite{shen2025htd}, Soar~\cite{verma2024soar} \\\cmidrule{2-3}
& \raisebox{-1ex}[0pt][0pt]{Segmentation} & RWKV-SAM~\cite{yuan2024mamba}, Segman~\cite{fu2024segman}, VM-Unet~\cite{chen2024vision}, Pyramidmamba~\cite{wang2024pyramidmamba}, Vig~\cite{liao2025vig}, Vision-rwkv~\cite{duan2024vision} \\\cmidrule{2-3}
& \raisebox{-2.5ex}[0pt][0pt]{Enhancement \& Restoration} & Restore-rwkv~\cite{yang2024restore}, Dvmsr~\cite{lei2024dvmsr}, Q-mambar~\cite{chen2025q}, RWKV-IR~\cite{du2024exploring}, Pixmamba~\cite{lin2024pixmamba}, Watermamba~\cite{guan2024watermamba}, Retinexmamba~\cite{bai2024retinexmamba}, Llemamba~\cite{zhang2024llemamba}, Fouriermamba~\cite{li2024fouriermamba}, Vmambair~\cite{shi2025vmambair}, Serpent~\cite{sepehri2024serpent}, Matir~\cite{wen2025matir}, Cu-mamba~\cite{deng2024cu}, Lfmamba~\cite{lu2024lfmamba} \\\cmidrule{2-3}
& \raisebox{-1ex}[0pt][0pt]{Generation} & DiS~\cite{fei2024scalable}, Dim~\cite{teng2024dim}, DiM~\cite{mo2024scaling}, Zigma~\cite{hu2024zigma}, AiM~\cite{li2024scalable}, Maskmamba~\cite{chen2024maskmamba}, Dimba~\cite{fei2024dimba}, DiM-3D~\cite{mo2024efficient}, Gamba~\cite{shen2025gamba}, Diffusion-rwkv~\cite{fei2024diffusion}, Sdit~\cite{yang2024sdit} \\\cmidrule{2-3}
& \raisebox{-2.5ex}[0pt][0pt]{Medicine} & U-mamba~\cite{ma2024u}, Vm-unet~\cite{ruan2024vm}, Rwkv-unet~\cite{jiang2025rwkv}, Segmamba~\cite{xing2024segmamba}, Zig-rir~\cite{chen2025zig}, Bsbp-rwkv~\cite{zhou2024bsbp}, Mambamil~\cite{yang2024mambamil}, Mmr-mamba~\cite{zou2024mmr}, Mambamir~\cite{huang2024mambamir}, Vmambamorph~\cite{wang2024vmambamorph}, I2i-mamba~\cite{atli2024i2i}, Delta-wkv~\cite{lu2025delta} \\\cmidrule{2-3}
& \raisebox{-1ex}[0pt][0pt]{Autonomous Driving} & Mambabev~\cite{you2024mambabev}, Occrwkv~\cite{wang2024occrwkv}, H-mba~\cite{chen2025h}, Trajectory mamba~\cite{huang2025trajectory}, Drama~\cite{yuan2024drama}, Salm$^2$~\cite{zhao2025salm2} \\\cmidrule{2-3}
& \raisebox{-1.5ex}[0pt][0pt]{Remote Sensing} & RS-Mamba~\cite{zhao2024rs}, Samba~\cite{zhu2024samba}, Changemamba~\cite{chen2024changemamba}, Rs3mamba~\cite{ma2024rs3mamba}, Hsimamba~\cite{yang2024hsimamba}, Pan-mamba~\cite{he2024pan}, LE-Mamba~\cite{cao2024novel}, Rsdehamba~\cite{zhou2024rsdehamba}, FMSR~\cite{xiao2024frequency}, Rscama~\cite{liu2024rscama} \\
\midrule
\multirow{2}{*}{\raisebox{-2.5ex}[0pt][0pt]{Audio}}
& \raisebox{-1ex}[0pt][0pt]{Understanding} & Audio mamba~\cite{lin2024audio, erol2024audio_mamba, yadav2024audio}, Mamca~\cite{zhang2024MAMC}, Rawbmamba~\cite{chen2024rawbmamba}, BiMamba~\cite{zhang2024mamba}, Ssamba~\cite{shams2024ssamba}, Dual-path mamba~\cite{jiang2024dual}, Spmamba~\cite{li2024spmamba}, VAD~\cite{zuo2023advancing} \\\cmidrule{2-3}
& Enhancement \& Generation & SEMamba~\cite{chao2024investigation}, Tramba~\cite{sui2024tramba}, SaShiMi~\cite{goel2022its_raw}, Music-Diff~\cite{liu2024perturbing}, oSpatialNet-Mamba~\cite{quan2024multichannel_speech_enhancement} \\
\midrule
\multirow{2}{*}{\raisebox{-3ex}[0pt][0pt]{Multimodality}}
& \raisebox{-2.5ex}[0pt][0pt]{Understanding} & MaTAV~\cite{li2024mamba}, Avs-mamba~\cite{gong2025avs}, Av-mamba~\cite{huang2024av}, VisualRWKV-UHD~\cite{li2024visualrwkv}, Rwkv-clip~\cite{gu2024rwkv}, Lavida~\cite{li2025lavida}, LIMoE~\cite{mustafa2022multimodal}, Uni-MoE~\cite{li2025uni}, VL-MoE~\cite{shen2023scaling}, Moe-llava~\cite{lin2024moe}, MoCLE~\cite{gou2023mixture}, Llava-mole~\cite{chen2024llava}, PaCE~\cite{li2023pace}, Llada-v~\cite{you2025lladavlargelanguagediffusion}, Dimple~\cite{yu2025dimplediscretediffusionmultimodal} \\\cmidrule{2-3}
& Unified & Fragkiadaki~\cite{swerdlow2025unified}, Mmada~\cite{yang2025mmadamultimodallargediffusion} \\
\bottomrule
\end{tabular}
\label{tab:mamba_methods}
\end{table}

\subsection{Vision}
\label{subsec:vision}
While ViTs excel in many tasks, their quadratic computational complexity has spurred the development of more efficient architectures. This section reviews the burgeoning application of linear-time sequence models, such as SSMs, and sparse computation strategies like MoE. We explore their impact across the vision landscape, beginning with foundational tasks including classification, detection, and segmentation (\S\ref{sec:classification}). We then cover their role in image enhancement, restoration, and generation (\S\ref{sec:enhancement}). Finally, we highlight their growing importance in domain-specific applications such as medicine, autonomous driving, and remote sensing (\S\ref{sec:domain}). Collectively, these advancements mark a shift towards models that achieve high performance while remaining computationally tractable, enabling new capabilities at scale.

\subsubsection{Image Classification, Detection, and Segmentation}
\label{sec:classification}
In image classification, recent efforts have focused on adapting linear-time sequence models, particularly Mamba-based SSMs, to various vision tasks. A key trend is the creation of hybrid architectures that integrate SSMs with established modules, such as CNNs or residual connections, to effectively capture both local features and long-range dependencies~\cite{wang2025insectmamba, chen2024res, yue2024medmamba}. To overcome the inherent 1D nature of SSMs for 2D image processing, researchers have proposed innovative data scanning strategies, such as multi-path scanning for remote sensing images~\cite{chen2024rsmamba} and topology-aware scanning for graph-structured whole-slide images in medical pathology~\cite{fang2024mammil}. These models are also being tailored for highly specialized domains by addressing unique data challenges, including the high dimensionality of hyperspectral images~\cite{yao2024spectralmamba} and data scarcity in industrial defect recognition~\cite{wang2024memorymamba}. In a parallel pursuit of efficiency, sparse MoE models have been successfully applied to vision. The V-MoE demonstrated that sparsely activating ``expert'' sub-networks can scale Vision Transformers to billions of parameters while reducing computational costs~\cite{NEURIPS2021_48237d9f}. Subsequent work has further matured this paradigm by enhancing its adversarial robustness~\cite{zhang2023robust} and providing theoretical guarantees for its sample efficiency~\cite{chowdhury2023patch}.

Building on these foundational applications, the same principles are being extended to object detection, where linear-time models replace or augment traditional backbones to enhance efficiency and capability. A prominent trend is the direct integration of SSMs into established detector frameworks, such as in Mamba-YOLO, which leverages an SSM-based backbone to achieve real-time performance without pre-training~\cite{wang2024mambayolo}. The adaptability of SSMs is particularly evident in specialized and challenging domains. For instance, Voxel-Mamba introduces a novel group-free paradigm for 3D object detection from point clouds by processing entire voxel spaces as a single sequence~\cite{zhang2024voxel}. In multimodal and niche applications, models like Fusion-Mamba pioneer cross-modal fusion in the hidden state space for RGB-Infrared detection~\cite{dong2024fusion}, while others employ hierarchical structures to efficiently detect small or specialized targets~\cite{chen2024mim, shen2025htd, verma2024soar}. Beyond SSMs, other general-purpose linear-complexity backbones, such as Gated Linear Attention (ViG) and RWKV-based models (Vision-RWKV), are also emerging, demonstrating competitive performance with significantly reduced computational overhead~\cite{liao2025vig, duan2024vision}. Complementing these architectural innovations, systems-level optimizations like Tutel are proving critical for enabling the efficient, large-scale training of sparse MoE models, paving the way for future scalable detection systems~\cite{hwang2023tutel}.

The demand for efficiency is further amplified in semantic segmentation, a dense prediction task that requires processing high-resolution imagery while maintaining a global receptive field. Here again, linear-time sequence models like Mamba and RWKV have emerged as strong backbones. A key strategy involves creating hybrid encoder-decoder architectures that pair the global context-capturing ability of these models with local mechanisms, such as convolutional blocks or local attention, to preserve fine-grained details for accurate boundary delineation~\cite{yuan2024mamba, fu2024segman}. This approach is also adapted for specialized domains, from lightweight crack segmentation~\cite{chen2024vision} to remote sensing, where innovations in the decoder use SSMs for efficient pyramid feature fusion~\cite{wang2024pyramidmamba}. Confirming their versatility, general-purpose backbones like ViG and Vision-RWKV also prove highly effective for dense prediction, offering significant speed and memory advantages over traditional Transformers without sacrificing performance~\cite{liao2025vig, duan2024vision}.

\subsubsection{Image Enhancement, Restoration, and Generation}
\label{sec:enhancement}
Efficient architectures are transforming image enhancement and restoration by modeling long-range dependencies with linear complexity. Mamba and RWKV-based models, often embedded in U-Net frameworks, are being tailored for specific tasks. In low-light and underwater enhancement, for example, they are often combined with physical principles like Retinex theory~\cite{lin2024pixmamba, guan2024watermamba, bai2024retinexmamba, zhang2024llemamba}. For broader restoration challenges such as dehazing and super-resolution, key innovations focus on adapting these models to 2D data. These include sophisticated scanning strategies, hybrid Mamba-Transformer designs, and even modeling dependencies across feature channels~\cite{li2024fouriermamba, shi2025vmambair, sepehri2024serpent, wen2025matir, deng2024cu}. The efficiency of these models is particularly advantageous for high-dimensional data like 4D light fields and medical images~\cite{gao2024mamba, lu2024lfmamba, yang2024restore}. To ensure practical deployment, significant research also focuses on techniques like knowledge distillation, low-bit quantization, and the creation of balanced benchmark datasets~\cite{lei2024dvmsr, chen2025q, du2024exploring}.

Building on their success in restoration, these architectures are now being applied to the even more computationally demanding field of generative modeling. Here, a dominant trend is replacing expensive Transformer backbones in diffusion models with scalable architectures like Mamba~\cite{fei2024scalable, teng2024dim, mo2024scaling}. A central challenge is adapting the 1D nature of SSMs for 2D image processing. This has spurred innovations in scanning patterns, such as zigzag scans and bidirectional processing, to better capture spatial context~\cite{hu2024zigma}. The architectural shift extends beyond diffusion, with Mamba also being adapted for autoregressive generation to achieve substantial inference speed-ups~\cite{li2024scalable}. To leverage complementary strengths, hybrid models combining Mamba and Transformers are also being explored for tasks like masked image modeling and text-to-image synthesis~\cite{chen2024maskmamba, fei2024dimba}. This new class of efficient backbones is also enabling advances in high-dimensional outputs like 3D shape and scene generation~\cite{mo2024efficient, shen2025gamba}. Finally, other linear-time models like RWKV are emerging as a viable alternative, further diversifying the architectural landscape of generative AI~\cite{fei2024diffusion, yang2024sdit}.

\subsubsection{Domain-specific Applications}
\label{sec:domain}
\paragraph{Medicine.}
The medical imaging field has rapidly adopted efficient sequence models like Mamba and RWKV, leveraging their ability to model long-range dependencies in high-resolution data. In semantic segmentation, a dominant trend is the integration of these models into U-Net-like architectures to combine global context awareness with local feature extraction. This has been successfully applied to a wide range of tasks, from 2D biomedical and skin lesion segmentation~\cite{ma2024u, ruan2024vm, jiang2025rwkv} to challenging 3D volumetric data~\cite{xing2024segmamba}, with some approaches using innovative nested or boundary-preserving RWKV structures for enhanced efficiency and precision~\cite{chen2025zig, zhou2024bsbp}. This architectural flexibility also extends to classification, where these models excel in the data-limited settings common to medicine~\cite{nasiri2024vision} and are particularly powerful in computational pathology for modeling interactions across vast sequences of patches in whole-slide images~\cite{yang2024mambamil, ji2023rnn}. Furthermore, these models are proving indispensable for complex image reconstruction, restoration, and synthesis tasks. They are being used for multi-contrast MRI reconstruction~\cite{zou2024mmr}, joint reconstruction and uncertainty estimation~\cite{huang2024mambamir}, 3D deformable registration~\cite{wang2024vmambamorph}, and multi-modal image synthesis~\cite{atli2024i2i}, often outperforming established methods. The RWKV architecture and its variants have shown similar promise in general medical image restoration and super-resolution~\cite{yang2024restore, lu2025delta}. The versatility of these models is even enabling novel applications beyond static images, such as motion-guided tracking of endoscopic instruments in dynamic environments~\cite{zhang2024motion2}.

\paragraph{Autonomous Driving.}
In autonomous driving, efficient sequence modeling methods are being applied across the entire pipeline, from perception to prediction and planning. In the critical Bird's-Eye View (BEV) space, these models serve as powerful backbones for tasks like 3D object detection~\cite{you2024mambabev} and semantic occupancy prediction~\cite{wang2024occrwkv}, where they efficiently process temporal and spatial information with linear complexity. Their capabilities extend to other fundamental perception tasks, including multi-modal video understanding for risk detection~\cite{chen2025h} and self-supervised depth estimation~\cite{meng2025enhancing}. Beyond perception, these models are enhancing downstream modules by replacing computationally intensive attention mechanisms. For instance, they are used to build highly efficient trajectory forecasting models~\cite{huang2025trajectory} and to enable lightweight, end-to-end motion planners that fuse multi-modal sensor data~\cite{yuan2024drama}. The extreme efficiency of these architectures also enables novel in-cabin applications, such as ultra-lightweight models for real-time driver attention monitoring~\cite{zhao2025salm2}.

\paragraph{Remote Sensing.}
The field of remote sensing, characterized by very high-resolution (VHR) and hyperspectral imagery, has become a fertile ground for efficient sequence modeling methods due to their linear complexity. For dense prediction tasks like semantic segmentation and change detection, a key innovation has been the development of specialized scanning mechanisms, such as omnidirectional scans, to effectively capture the complex spatial layouts of ground features from a global perspective~\cite{zhao2024rs, zhu2024samba, chen2024changemamba, ma2024rs3mamba}. These models are also being tailored for specific data modalities, most notably for hyperspectral image classification and fusion (pansharpening), where they excel at modeling intricate spectral dependencies~\cite{yang2024hsimamba, he2024pan, cao2024novel}. Similarly, in restoration tasks like dehazing and super-resolution, hybrid approaches that combine SSMs with convolutional or frequency-domain modules are proving effective for enhancing both spatial and spectral quality~\cite{zhou2024rsdehamba, xiao2024frequency}. The application of these architectures extends beyond pixel-level tasks to higher-level semantic understanding, such as generating descriptive captions for detected changes between images, demonstrating their versatility across the remote sensing pipeline~\cite{liu2024rscama}.

\subsection{Audio}
\label{subsec:audio}
The audio domain has widely adopted efficient sequence models as a powerful alternative to self-attention, leveraging their linear complexity for processing long audio signals. These architectures, particularly Mamba and its variants, are rapidly establishing new baselines across a range of fundamental audio tasks. They have been successfully applied to audio tagging~\cite{lin2024audio}, classification~\cite{erol2024audio_mamba}, and specialized tasks like automatic modulation classification~\cite{zhang2024MAMC}. A key innovation for audio has been the development of bidirectional Mamba models, which are crucial for capturing the non-causal context inherent in many audio signals and have proven effective for tasks like deepfake detection and general speech processing~\cite{chen2024rawbmamba, zhang2024mamba}. Furthermore, these models serve as a powerful backbone for self-supervised pre-training, learning robust general-purpose audio representations that outperform prior Transformer-based methods on a wide array of downstream tasks~\cite{yadav2024audio, shams2024ssamba}.

Beyond general audio understanding, these models are setting new performance records in complex signal processing domains. In speech separation, a prominent trend is to replace the recurrent or attention modules in state-of-the-art frameworks like DPRNN and TF-GridNet with Mamba blocks. This strategy has led to new SOTA results on benchmark datasets while significantly reducing computational complexity~\cite{jiang2024dual, li2024spmamba}. Similarly, in speech enhancement, Mamba-based systems have achieved top performance on standard benchmarks~\cite{chao2024investigation}. Hybrid architectures are also emerging, such as combining Mamba with Transformers to create highly efficient enhancement models tailored for resource-constrained mobile and wearable platforms~\cite{sui2024tramba}.

The inherent recurrent nature of these models makes them exceptionally well-suited for generation and streaming applications. State Space Models were early pioneers in generating raw audio waveforms directly, outperforming classic autoregressive models in both quality and speed~\cite{goel2022its_raw}, and they continue to be relevant in advanced tasks like symbolic music generation~\cite{liu2024perturbing}. For real-time processing, both Mamba and RWKV have demonstrated strong capabilities. They are being used to build low-latency streaming systems for multi-channel speech enhancement~\cite{quan2024multichannel_speech_enhancement}, voice activity detection (VAD)~\cite{zuo2023advancing}, and automatic speech recognition (ASR), where RWKV-based transducers match or exceed the performance of traditional chunk-based models with significantly less memory usage~\cite{an2023exploring}.

\subsection{Multimodality}
\label{subsec:multimodality}
In multi-modality, efficient architectures are critical for aligning diverse data streams and scaling large models. Linear-time sequence models like Mamba and RWKV have proven effective for this purpose. Mamba, for instance, is used to build sophisticated alignment and fusion modules. In conversational emotion recognition, a Mamba-aligner synchronizes text, audio, and video features~\cite{li2024mamba}. For audio-visual segmentation and question answering, specialized modules like a Temporal Mamba Block and a Cross-Modality Mamba enable efficient temporal modeling and selective cross-modal attention, respectively~\cite{gong2025avs, huang2024av}. Similarly, the RWKV architecture is being adapted for robust vision-language learning. Models like VisualRWKV-HD use techniques such as lossless downsampling to process high-resolution images without increasing sequence length, while RWKV-CLIP integrates the efficient RWKV into a contrastive learning framework, achieving strong performance in zero-shot tasks~\cite{li2024visualrwkv, gu2024rwkv}.

In the generative space, an emerging trend is the use of diffusion models as an alternative to autoregressive systems for multimodal tasks. Researchers have developed large-scale diffusion language models like LaViDa and MMaDA, which are combined with visual encoders to achieve strong performance in multimodal understanding and generation~\cite{li2025lavida, yang2025mmadamultimodallargediffusion}. A key focus is on visual instruction tuning, where models like LLaDA-V demonstrate that a pure diffusion framework can be competitive with top-tier autoregressive models~\cite{you2025lladavlargelanguagediffusion}. A notable sub-trend is the development of discrete diffusion models, such as Dimple and UniDisc. These models map all modalities to discrete token sequences, enabling parallel decoding and novel capabilities like unified cross-modal editing~\cite{yu2025dimplediscretediffusionmultimodal, swerdlow2025unified}.

To scale these large multimodal models efficiently, sparse MoE has become a dominant paradigm. MoE architectures are used to build foundational models from scratch, such as LIMoE for contrastive learning and Uni-MoE for unified systems that handle text, image, audio, and video within a single framework~\cite{mustafa2022multimodal, li2025uni, shen2023scaling}. Sparsity is also a key strategy for efficient fine-tuning. For example, MoE-LLaVA introduces a ``MoE-Tuning'' strategy to increase model capacity with constant computational cost~\cite{lin2024moe}. Furthermore, a mixture of specialized LoRA experts is being used to mitigate task and data conflicts during instruction tuning, where different experts can focus on specific instruction clusters or data domains~\cite{gou2023mixture, chen2024llava}. Finally, the concept of compositional experts, as seen in PaCE, breaks down complex tasks like multimodal dialogue into sub-skills handled by different expert modules, which are trained progressively~\cite{li2023pace}.

\section{Conclusion and Future Directions}
\label{sec: future}

In this survey, we have reviewed the key architectural innovations and optimization strategies developed to overcome the efficiency bottlenecks of Transformer‑based models. We highlighted how the quadratic cost of self‑attention and the growth of FFN layers drive up both computation and memory demands, especially in long‑sequence, multimodal, and multi‑step reasoning scenarios. We categorized recent solutions into seven main areas: linear sequence modeling, sparse sequence modeling, efficient full attention, sparse mixture of experts, hybrid architectures, diffusion LLMs, and cross‑modal applications. For each category, we examine the core ideas and underlying technical details, summarize representative works, and analyze the strengths and limitations of them. By organizing these approaches systematically, we aim to provide a clear picture of the current landscape and the common challenges they address.

Looking forward, we identify several promising directions for future exploration:

\paragraph{Efficient Architectures Design.}
As models continue to grow in scale and are expected to operate across a wide range of environments, from cloud to edge, there is a pressing need to rethink core design principles. The following research directions highlight key architectural innovations driving these goals forward:

\begin{itemize}[leftmargin=*]

  \item \textbf{Algorithm-System-Hardware Co‑Design:}  
  Jointly co-designing algorithm, system and hardware can improve efficiency for linear, sparse, or full attention, especially on edge devices and specialized chips.
  
  \item \textbf{Adaptive Attention Mechanisms:}  
  Attention modules that dynamically adjust sparsity or computation based on input or hardware conditions can better balance efficiency and flexibility.

  \item \textbf{Enhanced MoE Routing:}  
  Smarter MoE routing can improve expert utilization, reduce communication overhead, and lower latency during inference.

  \item \textbf{Efficient Large Models with Far More Parameters:}  
  Scaling models to even larger sizes requires innovations in memory layout, sparse activation, and communication-efficient designs.

  \item \textbf{Hierarchical Memory Architectures:}  
  Multi‑tiered memory modules (local, short‑term, long‑term) integrated into the model to efficiently store and retrieve past computation results and world knowledge.

  \item \textbf{Efficient Small Models on Edge Devices:}  
  Designing efficient small-scale LLMs or VLMs for edge deployment calls for quantization, pruning, and compact architecture design.

  \item \textbf{Non‑Autoregressive Diffusion LLMs:}  
  Diffusion-based LLMs offer parallel generation and faster inference, with potential to match autoregressive quality in tasks like dialogue and summarization.

\end{itemize}

\paragraph{Applications of Efficient Architectures.}
Beyond improving core architectural efficiency, an equally important frontier lies in applying these advancements to broaden the functional capabilities of language and multimodal models. As models are increasingly expected to operate in real-time, dynamic, and multimodal environments, new design priorities emerge, ranging from infinite context handling and agentic behavior to lifelong learning and multimodal reasoning. The following directions outline practical applications of efficient design:

\begin{itemize}[leftmargin=*]

  \item \textbf{Infinite Long Context:}  
  Efficient models facilitate handling of extremely long or even unbounded contexts, enhancing RAG, agents, reasoning, and multimodal tasks over extended inputs.

  \item \textbf{Efficient Agentic LLMs:}  
  Models optimized for efficiency enable real-time tool usage, planning, and multimodal reasoning, supporting agile agent behaviors with minimal latency in interactive applications.

  \item \textbf{Efficient Large Reasoning Models:}  
  Efficient reasoning models reduce redundant computation and leverage lightweight logic or memory components, improving scalability in tasks.

  \item \textbf{Efficient Vision-Language-Action (VLA) Models:}  
  Efficient multi-modal fusion and rapid visual reasoning empower VLA models to perform real-time control in robotics and interactive systems.

  \item \textbf{Efficient Omni-modal Models:}  
  Unified efficient models seamlessly process diverse modalities, including text, vision, audio, and 3D data.

  \item \textbf{Efficient Unified Multimodal Models for Understanding and Generation:}  
  Combining multimodal perception with generation supports more coherent and context-aware outputs in applications.

  \item \textbf{Continual Adaptation and Lifelong Learning:}  
  Architectures that support on‑the‑fly adaptation to new data streams without catastrophic forgetting, enabling LLMs to evolve continually in long-term changing environments.

\end{itemize}



\bibliographystyle{unsrt}
\bibliography{reference}

\begin{thebibliography}{100}

\bibitem{zhao2023survey}
Wayne~Xin Zhao, Kun Zhou, Junyi Li, Tianyi Tang, Xiaolei Wang, Yupeng Hou, Yingqian Min, Beichen Zhang, Junjie Zhang, Zican Dong, et~al.
\newblock A survey of large language models.
\newblock {\em arXiv preprint arXiv:2303.18223}, 1(2), 2023.

\bibitem{brown2020language}
Tom Brown, Benjamin Mann, Nick Ryder, Melanie Subbiah, Jared~D Kaplan, Prafulla Dhariwal, Arvind Neelakantan, Pranav Shyam, Girish Sastry, Amanda Askell, et~al.
\newblock Language models are few-shot learners.
\newblock {\em Advances in neural information processing systems}, 33:1877--1901, 2020.

\bibitem{raffel2020exploring}
Colin Raffel, Noam Shazeer, Adam Roberts, Katherine Lee, Sharan Narang, Michael Matena, Yanqi Zhou, Wei Li, and Peter~J Liu.
\newblock Exploring the limits of transfer learning with a unified text-to-text transformer.
\newblock {\em Journal of machine learning research}, 21(140):1--67, 2020.

\bibitem{feng2020codebert}
Zhangyin Feng, Daya Guo, Duyu Tang, Nan Duan, Xiaocheng Feng, Ming Gong, Linjun Shou, Bing Qin, Ting Liu, Daxin Jiang, et~al.
\newblock Codebert: A pre-trained model for programming and natural languages.
\newblock {\em arXiv preprint arXiv:2002.08155}, 2020.

\bibitem{li2022competition}
Yujia Li, David Choi, Junyoung Chung, Nate Kushman, Julian Schrittwieser, R{\'e}mi Leblond, Tom Eccles, James Keeling, Felix Gimeno, Agustin Dal~Lago, et~al.
\newblock Competition-level code generation with alphacode.
\newblock {\em Science}, 378(6624):1092--1097, 2022.

\bibitem{fried2022incoder}
Daniel Fried, Armen Aghajanyan, Jessy Lin, Sida Wang, Eric Wallace, Freda Shi, Ruiqi Zhong, Wen-tau Yih, Luke Zettlemoyer, and Mike Lewis.
\newblock Incoder: A generative model for code infilling and synthesis.
\newblock {\em arXiv preprint arXiv:2204.05999}, 2022.

\bibitem{lewis2020retrieval}
Patrick Lewis, Ethan Perez, Aleksandra Piktus, Fabio Petroni, Vladimir Karpukhin, Naman Goyal, Heinrich K{\"u}ttler, Mike Lewis, Wen-tau Yih, Tim Rockt{\"a}schel, et~al.
\newblock Retrieval-augmented generation for knowledge-intensive nlp tasks.
\newblock {\em Advances in neural information processing systems}, 33:9459--9474, 2020.

\bibitem{guu2020retrieval}
Kelvin Guu, Kenton Lee, Zora Tung, Panupong Pasupat, and Mingwei Chang.
\newblock Retrieval augmented language model pre-training.
\newblock In {\em International conference on machine learning}, pages 3929--3938. PMLR, 2020.

\bibitem{ranathunga2023neural}
Surangika Ranathunga, En-Shiun~Annie Lee, Marjana Prifti~Skenduli, Ravi Shekhar, Mehreen Alam, and Rishemjit Kaur.
\newblock Neural machine translation for low-resource languages: A survey.
\newblock {\em ACM Computing Surveys}, 55(11):1--37, 2023.

\bibitem{radford2018improving}
Alec Radford, Karthik Narasimhan, Tim Salimans, Ilya Sutskever, et~al.
\newblock Improving language understanding by generative pre-training, 2018.

\bibitem{radford2019language}
Alec Radford, Jeffrey Wu, Rewon Child, David Luan, Dario Amodei, Ilya Sutskever, et~al.
\newblock Language models are unsupervised multitask learners.
\newblock {\em OpenAI blog}, 1(8):9, 2019.

\bibitem{achiam2023gpt}
Josh Achiam, Steven Adler, Sandhini Agarwal, Lama Ahmad, Ilge Akkaya, Florencia~Leoni Aleman, Diogo Almeida, Janko Altenschmidt, Sam Altman, Shyamal Anadkat, et~al.
\newblock Gpt-4 technical report.
\newblock {\em arXiv preprint arXiv:2303.08774}, 2023.

\bibitem{hurst2024gpt}
Aaron Hurst, Adam Lerer, Adam~P Goucher, Adam Perelman, Aditya Ramesh, Aidan Clark, AJ~Ostrow, Akila Welihinda, Alan Hayes, Alec Radford, et~al.
\newblock Gpt-4o system card.
\newblock {\em arXiv preprint arXiv:2410.21276}, 2024.

\bibitem{jaech2024openai}
Aaron Jaech, Adam Kalai, Adam Lerer, Adam Richardson, Ahmed El-Kishky, Aiden Low, Alec Helyar, Aleksander Madry, Alex Beutel, Alex Carney, et~al.
\newblock Openai o1 system card.
\newblock {\em arXiv preprint arXiv:2412.16720}, 2024.

\bibitem{OpenAI_GPT-o3_SystemCard}
{OpenAI}.
\newblock Openai o3 and o4-mini system card.
\newblock Technical report, OpenAI, April 2025.

\bibitem{gpt-oss}
{OpenAI}.
\newblock gpt-oss-120b \& gpt-oss-20b model card.
\newblock Technical report, OpenAI, August 2025.

\bibitem{gpt5}
OpenAI.
\newblock Introducing gpt-5, August 2025.
\newblock Blog post, August 7, 2025.

\bibitem{claude1}
Anthropic.
\newblock Introducing claude, March 2023.
\newblock Blog post, March 14, 2023.

\bibitem{claude21}
Anthropic.
\newblock Introducing claude 2.1, November 2023.
\newblock Blog post, November 21, 2023.

\bibitem{claude37}
Anthropic.
\newblock Claude 3.7 sonnet and claude code, February 2025.
\newblock Blog post, February 25, 2025.

\bibitem{claude4}
Anthropic.
\newblock Introducing claude 4, May 2025.
\newblock Blog post, May 23, 2025.

\bibitem{claude41}
Anthropic.
\newblock Claude opus 4.1, August 2025.
\newblock Blog post, August 6, 2025.

\bibitem{team2023gemini}
Gemini Team, Rohan Anil, Sebastian Borgeaud, Jean-Baptiste Alayrac, Jiahui Yu, Radu Soricut, Johan Schalkwyk, Andrew~M Dai, Anja Hauth, Katie Millican, et~al.
\newblock Gemini: a family of highly capable multimodal models.
\newblock {\em arXiv preprint arXiv:2312.11805}, 2023.

\bibitem{team2024gemini}
Gemini Team, Petko Georgiev, Ving~Ian Lei, Ryan Burnell, Libin Bai, Anmol Gulati, Garrett Tanzer, Damien Vincent, Zhufeng Pan, Shibo Wang, et~al.
\newblock Gemini 1.5: Unlocking multimodal understanding across millions of tokens of context.
\newblock {\em arXiv preprint arXiv:2403.05530}, 2024.

\bibitem{Gemini2025}
{Gemini Team, Google}.
\newblock Gemini 2.5: Pushing the frontier with advanced reasoning, multimodality, long context, and next generation agentic capabilities, 2025.

\bibitem{bi2024deepseek}
Xiao Bi, Deli Chen, Guanting Chen, Shanhuang Chen, Damai Dai, Chengqi Deng, Honghui Ding, Kai Dong, Qiushi Du, Zhe Fu, et~al.
\newblock Deepseek llm: Scaling open-source language models with longtermism.
\newblock {\em arXiv preprint arXiv:2401.02954}, 2024.

\bibitem{liu2024deepseekv2}
Aixin Liu, Bei Feng, Bin Wang, Bingxuan Wang, Bo~Liu, Chenggang Zhao, Chengqi Dengr, Chong Ruan, Damai Dai, Daya Guo, et~al.
\newblock Deepseek-v2: A strong, economical, and efficient mixture-of-experts language model.
\newblock {\em arXiv preprint arXiv:2405.04434}, 2024.

\bibitem{liu2024deepseekv3}
Aixin Liu, Bei Feng, Bing Xue, Bingxuan Wang, Bochao Wu, Chengda Lu, Chenggang Zhao, Chengqi Deng, Chenyu Zhang, Chong Ruan, et~al.
\newblock Deepseek-v3 technical report.
\newblock {\em arXiv preprint arXiv:2412.19437}, 2024.

\bibitem{guo2025deepseek}
Daya Guo, Dejian Yang, Haowei Zhang, Junxiao Song, Ruoyu Zhang, Runxin Xu, Qihao Zhu, Shirong Ma, Peiyi Wang, Xiao Bi, et~al.
\newblock Deepseek-r1: Incentivizing reasoning capability in llms via reinforcement learning.
\newblock {\em arXiv preprint arXiv:2501.12948}, 2025.

\bibitem{bai2023qwen}
Jinze Bai, Shuai Bai, Yunfei Chu, Zeyu Cui, Kai Dang, Xiaodong Deng, Yang Fan, Wenbin Ge, Yu~Han, Fei Huang, et~al.
\newblock Qwen technical report.
\newblock {\em arXiv preprint arXiv:2309.16609}, 2023.

\bibitem{team2024qwen2}
Qwen Team.
\newblock Qwen2 technical report.
\newblock {\em arXiv preprint arXiv:2412.15115}, 2024.

\bibitem{Yang2024Qwen25TR}
Qwen, An~Yang, Baosong Yang, Beichen Zhang, Binyuan Hui, Bo~Zheng, Bowen Yu, Chengyuan Li, Dayiheng Liu, Fei Huang, Guanting Dong, Haoran Wei, Huan Lin, Jian Yang, Jianhong Tu, Jianwei Zhang, Jianxin Yang, Jiaxin Yang, Jingren Zhou, Junyang Lin, Kai Dang, Keming Lu, Keqin Bao, Kexin Yang, Le~Yu, Mei Li, Mingfeng Xue, Pei Zhang, Qin Zhu, Rui Men, Runji Lin, Tianhao Li, Tingyu Xia, Xingzhang Ren, Xuancheng Ren, Yang Fan, Yang Su, Yi-Chao Zhang, Yunyang Wan, Yuqi Liu, Zeyu Cui, Zhenru Zhang, Zihan Qiu, Shanghaoran Quan, and Zekun Wang.
\newblock Qwen2.5 technical report.
\newblock {\em ArXiv}, abs/2412.15115, 2024.

\bibitem{yang2025qwen3}
An~Yang, Anfeng Li, Baosong Yang, Beichen Zhang, Binyuan Hui, Bo~Zheng, Bowen Yu, Chang Gao, Chengen Huang, Chenxu Lv, et~al.
\newblock Qwen3 technical report.
\newblock {\em arXiv preprint arXiv:2505.09388}, 2025.

\bibitem{touvron2023llama}
Hugo Touvron, Thibaut Lavril, Gautier Izacard, Xavier Martinet, Marie-Anne Lachaux, Timoth{\'e}e Lacroix, Baptiste Rozi{\`e}re, Naman Goyal, Eric Hambro, Faisal Azhar, et~al.
\newblock Llama: Open and efficient foundation language models.
\newblock {\em arXiv preprint arXiv:2302.13971}, 2023.

\bibitem{touvron2023llama2}
Hugo Touvron, Louis Martin, Kevin Stone, Peter Albert, Amjad Almahairi, Yasmine Babaei, Nikolay Bashlykov, Soumya Batra, Prajjwal Bhargava, Shruti Bhosale, et~al.
\newblock Llama 2: Open foundation and fine-tuned chat models.
\newblock {\em arXiv preprint arXiv:2307.09288}, 2023.

\bibitem{grattafiori2024llama}
Aaron Grattafiori, Abhimanyu Dubey, Abhinav Jauhri, Abhinav Pandey, Abhishek Kadian, Ahmad Al-Dahle, Aiesha Letman, Akhil Mathur, Alan Schelten, Alex Vaughan, et~al.
\newblock The llama 3 herd of models.
\newblock {\em arXiv preprint arXiv:2407.21783}, 2024.

\bibitem{meta2025llama}
AI~Meta.
\newblock The llama 4 herd: The beginning of a new era of natively multimodal ai innovation.
\newblock {\em https://ai. meta. com/blog/llama-4-multimodal-intelligence/, checked on}, 4(7):2025, 2025.

\bibitem{glm2024chatglm}
Team GLM, Aohan Zeng, Bin Xu, Bowen Wang, Chenhui Zhang, Da~Yin, Dan Zhang, Diego Rojas, Guanyu Feng, Hanlin Zhao, et~al.
\newblock Chatglm: A family of large language models from glm-130b to glm-4 all tools.
\newblock {\em arXiv preprint arXiv:2406.12793}, 2024.

\bibitem{li2025minimax}
Aonian Li, Bangwei Gong, Bo~Yang, Boji Shan, Chang Liu, Cheng Zhu, Chunhao Zhang, Congchao Guo, Da~Chen, Dong Li, et~al.
\newblock Minimax-01: Scaling foundation models with lightning attention.
\newblock {\em arXiv preprint arXiv:2501.08313}, 2025.

\bibitem{team2023internlm}
InternLM Team.
\newblock Internlm: A multilingual language model with progressively enhanced capabilities, 2023.

\bibitem{cai2024internlm2}
Zheng Cai, Maosong Cao, Haojiong Chen, Kai Chen, Keyu Chen, Xin Chen, Xun Chen, Zehui Chen, Zhi Chen, Pei Chu, et~al.
\newblock Internlm2 technical report.
\newblock {\em arXiv preprint arXiv:2403.17297}, 2024.

\bibitem{sun2024hunyuan}
Xingwu Sun, Yanfeng Chen, Yiqing Huang, Ruobing Xie, Jiaqi Zhu, Kai Zhang, Shuaipeng Li, Zhen Yang, Jonny Han, Xiaobo Shu, et~al.
\newblock Hunyuan-large: An open-source moe model with 52 billion activated parameters by tencent.
\newblock {\em arXiv preprint arXiv:2411.02265}, 2024.

\bibitem{liu2025hunyuan}
Ao~Liu, Botong Zhou, Can Xu, Chayse Zhou, ChenChen Zhang, Chengcheng Xu, Chenhao Wang, Decheng Wu, Dengpeng Wu, Dian Jiao, et~al.
\newblock Hunyuan-turbos: Advancing large language models through mamba-transformer synergy and adaptive chain-of-thought.
\newblock {\em arXiv preprint arXiv:2505.15431}, 2025.

\bibitem{wang2024qwen2}
Peng Wang, Shuai Bai, Sinan Tan, Shijie Wang, Zhihao Fan, Jinze Bai, Keqin Chen, Xuejing Liu, Jialin Wang, Wenbin Ge, et~al.
\newblock Qwen2-vl: Enhancing vision-language model's perception of the world at any resolution.
\newblock {\em arXiv preprint arXiv:2409.12191}, 2024.

\bibitem{bai2025qwen2}
Shuai Bai, Keqin Chen, Xuejing Liu, Jialin Wang, Wenbin Ge, Sibo Song, Kai Dang, Peng Wang, Shijie Wang, Jun Tang, et~al.
\newblock Qwen2.5-vl technical report.
\newblock {\em arXiv preprint arXiv:2502.13923}, 2025.

\bibitem{xu2025qwen2}
Jin Xu, Zhifang Guo, Jinzheng He, Hangrui Hu, Ting He, Shuai Bai, Keqin Chen, Jialin Wang, Yang Fan, Kai Dang, et~al.
\newblock Qwen2. 5-omni technical report.
\newblock {\em arXiv preprint arXiv:2503.20215}, 2025.

\bibitem{chen2024internvl}
Zhe Chen, Jiannan Wu, Wenhai Wang, Weijie Su, Guo Chen, Sen Xing, Muyan Zhong, Qinglong Zhang, Xizhou Zhu, Lewei Lu, et~al.
\newblock Internvl: Scaling up vision foundation models and aligning for generic visual-linguistic tasks.
\newblock In {\em Proceedings of the IEEE/CVF Conference on Computer Vision and Pattern Recognition}, pages 24185--24198, 2024.

\bibitem{chen2024far}
Zhe Chen, Weiyun Wang, Hao Tian, Shenglong Ye, Zhangwei Gao, Erfei Cui, Wenwen Tong, Kongzhi Hu, Jiapeng Luo, Zheng Ma, et~al.
\newblock How far are we to gpt-4v? closing the gap to commercial multimodal models with open-source suites.
\newblock {\em Science China Information Sciences}, 67(12):220101, 2024.

\bibitem{chen2024expanding}
Zhe Chen, Weiyun Wang, Yue Cao, Yangzhou Liu, Zhangwei Gao, Erfei Cui, Jinguo Zhu, Shenglong Ye, Hao Tian, Zhaoyang Liu, et~al.
\newblock Expanding performance boundaries of open-source multimodal models with model, data, and test-time scaling.
\newblock {\em arXiv preprint arXiv:2412.05271}, 2024.

\bibitem{zhu2025internvl3}
Jinguo Zhu, Weiyun Wang, Zhe Chen, Zhaoyang Liu, Shenglong Ye, Lixin Gu, Hao Tian, Yuchen Duan, Weijie Su, Jie Shao, et~al.
\newblock Internvl3: Exploring advanced training and test-time recipes for open-source multimodal models.
\newblock {\em arXiv preprint arXiv:2504.10479}, 2025.

\bibitem{guo2025seed1}
Dong Guo, Faming Wu, Feida Zhu, Fuxing Leng, Guang Shi, Haobin Chen, Haoqi Fan, Jian Wang, Jianyu Jiang, Jiawei Wang, et~al.
\newblock Seed1. 5-vl technical report.
\newblock {\em arXiv preprint arXiv:2505.07062}, 2025.

\bibitem{team2025kimivl}
Kimi Team, Angang Du, Bohong Yin, Bowei Xing, Bowen Qu, Bowen Wang, Cheng Chen, Chenlin Zhang, Chenzhuang Du, Chu Wei, et~al.
\newblock Kimi-vl technical report.
\newblock {\em arXiv preprint arXiv:2504.07491}, 2025.

\bibitem{seed2025seed1}
ByteDance Seed, Jiaze Chen, Tiantian Fan, Xin Liu, Lingjun Liu, Zhiqi Lin, Mingxuan Wang, Chengyi Wang, Xiangpeng Wei, Wenyuan Xu, et~al.
\newblock Seed1. 5-thinking: Advancing superb reasoning models with reinforcement learning.
\newblock {\em arXiv preprint arXiv:2504.13914}, 2025.

\bibitem{chen2025minimax}
Aili Chen, Aonian Li, Bangwei Gong, Binyang Jiang, Bo~Fei, Bo~Yang, Boji Shan, Changqing Yu, Chao Wang, Cheng Zhu, et~al.
\newblock Minimax-m1: Scaling test-time compute efficiently with lightning attention.
\newblock {\em arXiv preprint arXiv:2506.13585}, 2025.

\bibitem{team2025kimi}
Kimi Team, Angang Du, Bofei Gao, Bowei Xing, Changjiu Jiang, Cheng Chen, Cheng Li, Chenjun Xiao, Chenzhuang Du, Chonghua Liao, et~al.
\newblock Kimi k1. 5: Scaling reinforcement learning with llms.
\newblock {\em arXiv preprint arXiv:2501.12599}, 2025.

\bibitem{kimiteam2025kimik2openagentic}
Kimi Team, Yifan Bai, Yiping Bao, Guanduo Chen, Jiahao Chen, Ningxin Chen, Ruijue Chen, Yanru Chen, Yuankun Chen, Yutian Chen, Zhuofu Chen, Jialei Cui, Hao Ding, Mengnan Dong, Angang Du, Chenzhuang Du, Dikang Du, Yulun Du, Yu~Fan, Yichen Feng, Kelin Fu, Bofei Gao, Hongcheng Gao, Peizhong Gao, Tong Gao, Xinran Gu, Longyu Guan, Haiqing Guo, Jianhang Guo, Hao Hu, Xiaoru Hao, Tianhong He, Weiran He, Wenyang He, Chao Hong, Yangyang Hu, Zhenxing Hu, Weixiao Huang, Zhiqi Huang, Zihao Huang, Tao Jiang, Zhejun Jiang, Xinyi Jin, Yongsheng Kang, Guokun Lai, Cheng Li, Fang Li, Haoyang Li, Ming Li, Wentao Li, Yanhao Li, Yiwei Li, Zhaowei Li, Zheming Li, Hongzhan Lin, Xiaohan Lin, Zongyu Lin, Chengyin Liu, Chenyu Liu, Hongzhang Liu, Jingyuan Liu, Junqi Liu, Liang Liu, Shaowei Liu, T.~Y. Liu, Tianwei Liu, Weizhou Liu, Yangyang Liu, Yibo Liu, Yiping Liu, Yue Liu, Zhengying Liu, Enzhe Lu, Lijun Lu, Shengling Ma, Xinyu Ma, Yingwei Ma, Shaoguang Mao, Jie Mei, Xin Men, Yibo Miao, Siyuan Pan, Yebo Peng, Ruoyu Qin, Bowen Qu, Zeyu
  Shang, Lidong Shi, Shengyuan Shi, Feifan Song, Jianlin Su, Zhengyuan Su, Xinjie Sun, Flood Sung, Heyi Tang, Jiawen Tao, Qifeng Teng, Chensi Wang, Dinglu Wang, Feng Wang, Haiming Wang, Jianzhou Wang, Jiaxing Wang, Jinhong Wang, Shengjie Wang, Shuyi Wang, Yao Wang, Yejie Wang, Yiqin Wang, Yuxin Wang, Yuzhi Wang, Zhaoji Wang, Zhengtao Wang, Zhexu Wang, Chu Wei, Qianqian Wei, Wenhao Wu, Xingzhe Wu, Yuxin Wu, Chenjun Xiao, Xiaotong Xie, Weimin Xiong, Boyu Xu, Jing Xu, Jinjing Xu, L.~H. Xu, Lin Xu, Suting Xu, Weixin Xu, Xinran Xu, Yangchuan Xu, Ziyao Xu, Junjie Yan, Yuzi Yan, Xiaofei Yang, Ying Yang, Zhen Yang, Zhilin Yang, Zonghan Yang, Haotian Yao, Xingcheng Yao, Wenjie Ye, Zhuorui Ye, Bohong Yin, Longhui Yu, Enming Yuan, Hongbang Yuan, Mengjie Yuan, Haobing Zhan, Dehao Zhang, Hao Zhang, Wanlu Zhang, Xiaobin Zhang, Yangkun Zhang, Yizhi Zhang, Yongting Zhang, Yu~Zhang, Yutao Zhang, Yutong Zhang, Zheng Zhang, Haotian Zhao, Yikai Zhao, Huabin Zheng, Shaojie Zheng, Jianren Zhou, Xinyu Zhou, Zaida Zhou, Zhen Zhu,
  Weiyu Zhuang, and Xinxing Zu.
\newblock Kimi k2: Open agentic intelligence, 2025.

\bibitem{wei2022chain}
Jason Wei, Xuezhi Wang, Dale Schuurmans, Maarten Bosma, Fei Xia, Ed~Chi, Quoc~V Le, Denny Zhou, et~al.
\newblock Chain-of-thought prompting elicits reasoning in large language models.
\newblock {\em Advances in neural information processing systems}, 35:24824--24837, 2022.

\bibitem{qu2025survey}
Xiaoye Qu, Yafu Li, Zhaochen Su, Weigao Sun, Jianhao Yan, Dongrui Liu, Ganqu Cui, Daizong Liu, Shuxian Liang, Junxian He, et~al.
\newblock A survey of efficient reasoning for large reasoning models: Language, multimodality, and beyond.
\newblock {\em arXiv preprint arXiv:2503.21614}, 2025.

\bibitem{cottier2024rising}
Ben Cottier, Robi Rahman, Loredana Fattorini, Nestor Maslej, Tamay Besiroglu, and David Owen.
\newblock The rising costs of training frontier ai models.
\newblock {\em arXiv preprint arXiv:2405.21015}, 2024.

\bibitem{bogmans2025power}
Christian Bogmans, Patricia Gomez-Gonzalez, Giovanni Melina, Jorge Miranda-Pinto, Andrea Pescatori, and Sneha Thube.
\newblock Power hungry: How ai will drive energy demand.
\newblock Technical report, International Monetary Fund, 2025.

\bibitem{faiz2023llmcarbon}
Ahmad Faiz, Sotaro Kaneda, Ruhan Wang, Rita Osi, Prateek Sharma, Fan Chen, and Lei Jiang.
\newblock Llmcarbon: Modeling the end-to-end carbon footprint of large language models.
\newblock {\em arXiv preprint arXiv:2309.14393}, 2023.

\bibitem{vaswani2017attention}
A~Vaswani.
\newblock Attention is all you need.
\newblock {\em Advances in Neural Information Processing Systems}, 2017.

\bibitem{greff2016lstm}
Klaus Greff, Rupesh~K Srivastava, Jan Koutn{\'\i}k, Bas~R Steunebrink, and J{\"u}rgen Schmidhuber.
\newblock Lstm: A search space odyssey.
\newblock {\em IEEE transactions on neural networks and learning systems}, 28(10):2222--2232, 2016.

\bibitem{katharopoulos2020transformers}
Angelos Katharopoulos, Apoorv Vyas, Nikolaos Pappas, and Fran{\c{c}}ois Fleuret.
\newblock Transformers are rnns: Fast autoregressive transformers with linear attention.
\newblock In {\em International conference on machine learning}, pages 5156--5165. PMLR, 2020.

\bibitem{sun2025linear}
Weigao Sun, Disen Lan, Tong Zhu, Xiaoye Qu, and Yu~Cheng.
\newblock Linear-moe: Linear sequence modeling meets mixture-of-experts.
\newblock {\em arXiv preprint arXiv:2503.05447}, 2025.

\bibitem{qin2025ui}
Yujia Qin, Yining Ye, Junjie Fang, Haoming Wang, Shihao Liang, Shizuo Tian, Junda Zhang, Jiahao Li, Yunxin Li, Shijue Huang, et~al.
\newblock Ui-tars: Pioneering automated gui interaction with native agents.
\newblock {\em arXiv preprint arXiv:2501.12326}, 2025.

\bibitem{yin2024survey}
Shukang Yin, Chaoyou Fu, Sirui Zhao, Ke~Li, Xing Sun, Tong Xu, and Enhong Chen.
\newblock A survey on multimodal large language models.
\newblock {\em National Science Review}, 11(12):nwae403, 2024.

\bibitem{masoudnia2014mixture}
Saeed Masoudnia and Reza Ebrahimpour.
\newblock Mixture of experts: a literature survey.
\newblock {\em Artificial Intelligence Review}, 42(2):275--293, 2014.

\bibitem{peng2021abc}
Hao Peng, Jungo Kasai, Nikolaos Pappas, Dani Yogatama, Zhaofeng Wu, Lingpeng Kong, Roy Schwartz, and Noah~A Smith.
\newblock Abc: Attention with bounded-memory control.
\newblock {\em arXiv preprint arXiv:2110.02488}, 2021.

\bibitem{qin2024various}
Zhen Qin, Weigao Sun, Dong Li, Xuyang Shen, Weixuan Sun, and Yiran Zhong.
\newblock Various lengths, constant speed: Efficient language modeling with lightning attention.
\newblock {\em arXiv preprint arXiv:2405.17381}, 2024.

\bibitem{yang2023gated}
Songlin Yang, Bailin Wang, Yikang Shen, Rameswar Panda, and Yoon Kim.
\newblock Gated linear attention transformers with hardware-efficient training.
\newblock {\em arXiv preprint arXiv:2312.06635}, 2023.

\bibitem{zhang2024gated}
Yu~Zhang, Songlin Yang, Ruijie Zhu, Yue Zhang, Leyang Cui, Yiqiao Wang, Bolun Wang, Freda Shi, Bailin Wang, Wei Bi, et~al.
\newblock Gated slot attention for efficient linear-time sequence modeling.
\newblock {\em arXiv preprint arXiv:2409.07146}, 2024.

\bibitem{qin2024you}
Zhen Qin, Yuxin Mao, Xuyang Shen, Dong Li, Jing Zhang, Yuchao Dai, and Yiran Zhong.
\newblock You only scan once: Efficient multi-dimension sequential modeling with lightnet.
\newblock {\em arXiv preprint arXiv:2405.21022}, 2024.

\bibitem{arora2024simple}
Simran Arora, Sabri Eyuboglu, Michael Zhang, Aman Timalsina, Silas Alberti, Dylan Zinsley, James Zou, Atri Rudra, and Christopher R{\'e}.
\newblock Simple linear attention language models balance the recall-throughput tradeoff.
\newblock {\em arXiv preprint arXiv:2402.18668}, 2024.

\bibitem{aksenov2024rebased}
Yaroslav Aksenov, Nikita Balagansky, Sofia Maria Lo~Cicero Vaina, Boris Shaposhnikov, Alexey Gorbatovski, and Daniil Gavrilov.
\newblock Linear transformers with learnable kernel functions are better in-context models.
\newblock {\em arXiv preprint arXiv:2402.10644}, 2024.

\bibitem{yang2024parallelizing}
Songlin Yang, Bailin Wang, Yu~Zhang, Yikang Shen, and Yoon Kim.
\newblock Parallelizing linear transformers with the delta rule over sequence length.
\newblock {\em arXiv preprint arXiv:2406.06484}, 2024.

\bibitem{yang2024gated}
Songlin Yang, Jan Kautz, and Ali Hatamizadeh.
\newblock Gated delta networks: Improving mamba2 with delta rule.
\newblock {\em arXiv preprint arXiv:2412.06464}, 2024.

\bibitem{du2025mom}
Jusen Du, Weigao Sun, Disen Lan, Jiaxi Hu, and Yu~Cheng.
\newblock Mom: Linear sequence modeling with mixture-of-memories.
\newblock {\em arXiv preprint arXiv:2502.13685}, 2025.

\bibitem{qin2024hierarchically}
Zhen Qin, Songlin Yang, and Yiran Zhong.
\newblock Hierarchically gated recurrent neural network for sequence modeling.
\newblock {\em Advances in Neural Information Processing Systems}, 36, 2024.

\bibitem{qin2024hgrn2}
Zhen Qin, Songlin Yang, Weixuan Sun, Xuyang Shen, Dong Li, Weigao Sun, and Yiran Zhong.
\newblock Hgrn2: Gated linear rnns with state expansion.
\newblock {\em arXiv preprint arXiv:2404.07904}, 2024.

\bibitem{peng2023rwkv}
Bo~Peng, Eric Alcaide, Quentin Anthony, Alon Albalak, Samuel Arcadinho, Stella Biderman, Huanqi Cao, Xin Cheng, Michael Chung, Matteo Grella, et~al.
\newblock Rwkv: Reinventing rnns for the transformer era.
\newblock {\em arXiv preprint arXiv:2305.13048}, 2023.

\bibitem{peng2024eagle}
Bo~Peng, Daniel Goldstein, Quentin Anthony, Alon Albalak, Eric Alcaide, Stella Biderman, Eugene Cheah, Xingjian Du, Teddy Ferdinan, Haowen Hou, et~al.
\newblock Eagle and finch: Rwkv with matrix-valued states and dynamic recurrence.
\newblock {\em arXiv preprint arXiv:2404.05892}, 2024.

\bibitem{peng2025rwkv}
Bo~Peng, Ruichong Zhang, Daniel Goldstein, Eric Alcaide, Haowen Hou, Janna Lu, William Merrill, Guangyu Song, Kaifeng Tan, Saiteja Utpala, et~al.
\newblock Rwkv-7" goose" with expressive dynamic state evolution.
\newblock {\em arXiv preprint arXiv:2503.14456}, 2025.

\bibitem{orvieto2023resurrecting}
Antonio Orvieto, Samuel~L Smith, Albert Gu, Anushan Fernando, Caglar Gulcehre, Razvan Pascanu, and Soham De.
\newblock Resurrecting recurrent neural networks for long sequences.
\newblock In {\em International Conference on Machine Learning}, pages 26670--26698. PMLR, 2023.

\bibitem{beck2024xlstm}
Maximilian Beck, Korbinian P{\"o}ppel, Markus Spanring, Andreas Auer, Oleksandra Prudnikova, Michael Kopp, G{\"u}nter Klambauer, Johannes Brandstetter, and Sepp Hochreiter.
\newblock xlstm: Extended long short-term memory.
\newblock {\em arXiv preprint arXiv:2405.04517}, 2024.

\bibitem{katsch2023gateloop}
Tobias Katsch.
\newblock Gateloop: Fully data-controlled linear recurrence for sequence modeling.
\newblock {\em arXiv preprint arXiv:2311.01927}, 2023.

\bibitem{voelker2019legendre}
Aaron Voelker, Ivana Kaji{\'c}, and Chris Eliasmith.
\newblock Legendre memory units: Continuous-time representation in recurrent neural networks.
\newblock {\em Advances in neural information processing systems}, 32, 2019.

\bibitem{gu2020hippo}
Albert Gu, Tri Dao, Stefano Ermon, Atri Rudra, and Christopher R{\'e}.
\newblock Hippo: Recurrent memory with optimal polynomial projections.
\newblock {\em Advances in neural information processing systems}, 33:1474--1487, 2020.

\bibitem{gu2021combining}
Albert Gu, Isys Johnson, Karan Goel, Khaled Saab, Tri Dao, Atri Rudra, and Christopher R{\'e}.
\newblock Combining recurrent, convolutional, and continuous-time models with linear state space layers.
\newblock {\em Advances in neural information processing systems}, 34:572--585, 2021.

\bibitem{gu2021efficiently}
Albert Gu, Karan Goel, and Christopher R{\'e}.
\newblock Efficiently modeling long sequences with structured state spaces.
\newblock {\em arXiv preprint arXiv:2111.00396}, 2021.

\bibitem{gu2022howtotrain}
Albert Gu, Isys Johnson, Aman Timalsina, Atri Rudra, and Christopher R{\'e}.
\newblock How to train your hippo: State space models with generalized orthogonal basis projections.
\newblock {\em arXiv preprint arXiv:2206.12037}, 2022.

\bibitem{gupta2022diagonal}
Ankit Gupta, Albert Gu, and Jonathan Berant.
\newblock Diagonal state spaces are as effective as structured state spaces.
\newblock {\em Advances in Neural Information Processing Systems}, 35:22982--22994, 2022.

\bibitem{gu2022parameterization}
Albert Gu, Karan Goel, Ankit Gupta, and Christopher R{\'e}.
\newblock On the parameterization and initialization of diagonal state space models.
\newblock {\em Advances in Neural Information Processing Systems}, 35:35971--35983, 2022.

\bibitem{smith2022simplified}
Jimmy~TH Smith, Andrew Warrington, and Scott~W Linderman.
\newblock Simplified state space layers for sequence modeling.
\newblock {\em arXiv preprint arXiv:2208.04933}, 2022.

\bibitem{zhang2023effectively}
Michael Zhang, Khaled~K Saab, Michael Poli, Tri Dao, Karan Goel, and Christopher R{\'e}.
\newblock Effectively modeling time series with simple discrete state spaces.
\newblock {\em arXiv preprint arXiv:2303.09489}, 2023.

\bibitem{hu2024time}
Jiaxi Hu, Disen Lan, Ziyu Zhou, Qingsong Wen, and Yuxuan Liang.
\newblock Time-ssm: Simplifying and unifying state space models for time series forecasting.
\newblock {\em arXiv preprint arXiv:2405.16312}, 2024.

\bibitem{wang2023stablessm}
Shida Wang and Qianxiao Li.
\newblock Stablessm: Alleviating the curse of memory in state-space models through stable reparameterization.
\newblock {\em arXiv preprint arXiv:2311.14495}, 2023.

\bibitem{yu2023robustifying}
Annan Yu, Arnur Nigmetov, Dmitriy Morozov, Michael~W Mahoney, and N~Benjamin Erichson.
\newblock Robustifying state-space models for long sequences via approximate diagonalization.
\newblock {\em arXiv preprint arXiv:2310.01698}, 2023.

\bibitem{hasani2022liquid}
Ramin Hasani, Mathias Lechner, Tsun-Hsuan Wang, Makram Chahine, Alexander Amini, and Daniela Rus.
\newblock Liquid structural state-space models.
\newblock {\em arXiv preprint arXiv:2209.12951}, 2022.

\bibitem{liu2024longhorn}
Bo~Liu, Rui Wang, Lemeng Wu, Yihao Feng, Peter Stone, and Qiang Liu.
\newblock Longhorn: State space models are amortized online learners.
\newblock {\em arXiv preprint arXiv:2407.14207}, 2024.

\bibitem{gu2023mamba}
Albert Gu and Tri Dao.
\newblock Mamba: Linear-time sequence modeling with selective state spaces.
\newblock {\em arXiv preprint arXiv:2312.00752}, 2023.

\bibitem{dao2024transformers}
Tri Dao and Albert Gu.
\newblock Transformers are ssms: Generalized models and efficient algorithms through structured state space duality.
\newblock {\em arXiv preprint arXiv:2405.21060}, 2024.

\bibitem{hu2025comba}
Jiaxi Hu, Yongqi Pan, Jusen Du, Disen Lan, Xiaqiang Tang, Qingsong Wen, Yuxuan Liang, and Weigao Sun.
\newblock Comba: Improving nonlinear rnns with closed-loop control.
\newblock {\em arXiv preprint arXiv:2506.02475}, 2025.

\bibitem{sun2024learning}
Yu~Sun, Xinhao Li, Karan Dalal, Jiarui Xu, Arjun Vikram, Genghan Zhang, Yann Dubois, Xinlei Chen, Xiaolong Wang, Sanmi Koyejo, et~al.
\newblock Learning to (learn at test time): Rnns with expressive hidden states.
\newblock {\em arXiv preprint arXiv:2407.04620}, 2024.

\bibitem{behrouz2024titans}
Ali Behrouz, Peilin Zhong, and Vahab Mirrokni.
\newblock Titans: Learning to memorize at test time.
\newblock {\em arXiv preprint arXiv:2501.00663}, 2024.

\bibitem{karami2025lattice}
Mahdi Karami and Vahab Mirrokni.
\newblock Lattice: Learning to efficiently compress the memory.
\newblock {\em arXiv preprint arXiv:2504.05646}, 2025.

\bibitem{behrouz2025s}
Ali Behrouz, Meisam Razaviyayn, Peilin Zhong, and Vahab Mirrokni.
\newblock It's all connected: A journey through test-time memorization, attentional bias, retention, and online optimization.
\newblock {\em arXiv preprint arXiv:2504.13173}, 2025.

\bibitem{behrouz2025atlas}
Ali Behrouz, Zeman Li, Praneeth Kacham, Majid Daliri, Yuan Deng, Peilin Zhong, Meisam Razaviyayn, and Vahab Mirrokni.
\newblock Atlas: Learning to optimally memorize the context at test time.
\newblock {\em arXiv preprint arXiv:2505.23735}, 2025.

\bibitem{von2025mesanet}
Johannes von Oswald, Nino Scherrer, Seijin Kobayashi, Luca Versari, Songlin Yang, Maximilian Schlegel, Kaitlin Maile, Yanick Schimpf, Oliver Sieberling, Alexander Meulemans, et~al.
\newblock Mesanet: Sequence modeling by locally optimal test-time training.
\newblock {\em arXiv preprint arXiv:2506.05233}, 2025.

\bibitem{qin2024unlocking}
Zhen Qin, Xuyang Shen, Dong Li, Weigao Sun, Stan Birchfield, Richard Hartley, and Yiran Zhong.
\newblock Unlocking the secrets of linear complexity sequence model from a unified perspective.
\newblock {\em arXiv preprint arXiv:2405.17383}, 2024.

\bibitem{kasai2021finetuning}
Jungo Kasai, Hao Peng, Yizhe Zhang, Dani Yogatama, Gabriel Ilharco, Nikolaos Pappas, Yi~Mao, Weizhu Chen, and Noah~A Smith.
\newblock Finetuning pretrained transformers into rnns.
\newblock {\em arXiv preprint arXiv:2103.13076}, 2021.

\bibitem{wang2024mamba}
Junxiong Wang, Daniele Paliotta, Avner May, Alexander Rush, and Tri Dao.
\newblock The mamba in the llama: Distilling and accelerating hybrid models.
\newblock {\em Advances in Neural Information Processing Systems}, 37:62432--62457, 2024.

\bibitem{mercat2024linearizing}
Jean Mercat, Igor Vasiljevic, Sedrick Keh, Kushal Arora, Achal Dave, Adrien Gaidon, and Thomas Kollar.
\newblock Linearizing large language models.
\newblock {\em arXiv preprint arXiv:2405.06640}, 2024.

\bibitem{zhang2024lolcats}
Michael Zhang, Simran Arora, Rahul Chalamala, Alan Wu, Benjamin Spector, Aaryan Singhal, Krithik Ramesh, and Christopher R{\'e}.
\newblock Lolcats: On low-rank linearizing of large language models.
\newblock {\em arXiv preprint arXiv:2410.10254}, 2024.

\bibitem{lan2025liger}
Disen Lan, Weigao Sun, Jiaxi Hu, Jusen Du, and Yu~Cheng.
\newblock Liger: Linearizing large language models to gated recurrent structures.
\newblock {\em arXiv preprint arXiv:2503.01496}, 2025.

\bibitem{child2019generating}
Rewon Child, Scott Gray, Alec Radford, and Ilya Sutskever.
\newblock Generating long sequences with sparse transformers.
\newblock {\em arXiv preprint arXiv:1904.10509}, 2019.

\bibitem{guo2019star}
Qipeng Guo, Xipeng Qiu, Pengfei Liu, Yunfan Shao, Xiangyang Xue, and Zheng Zhang.
\newblock Star-transformer.
\newblock {\em arXiv preprint arXiv:1902.09113}, 2019.

\bibitem{qiu2019blockwise}
Jiezhong Qiu, Hao Ma, Omer Levy, Scott Wen-tau Yih, Sinong Wang, and Jie Tang.
\newblock Blockwise self-attention for long document understanding.
\newblock {\em arXiv preprint arXiv:1911.02972}, 2019.

\bibitem{beltagy2020longformer}
Iz~Beltagy, Matthew~E Peters, and Arman Cohan.
\newblock Longformer: The long-document transformer.
\newblock {\em arXiv preprint arXiv:2004.05150}, 2020.

\bibitem{ainslie2020etc}
Joshua Ainslie, Santiago Ontanon, Chris Alberti, Vaclav Cvicek, Zachary Fisher, Philip Pham, Anirudh Ravula, Sumit Sanghai, Qifan Wang, and Li~Yang.
\newblock Etc: Encoding long and structured inputs in transformers.
\newblock {\em arXiv preprint arXiv:2004.08483}, 2020.

\bibitem{zaheer2020big}
Manzil Zaheer, Guru Guruganesh, Kumar~Avinava Dubey, Joshua Ainslie, Chris Alberti, Santiago Ontanon, Philip Pham, Anirudh Ravula, Qifan Wang, Li~Yang, et~al.
\newblock Big bird: Transformers for longer sequences.
\newblock {\em Advances in neural information processing systems}, 33:17283--17297, 2020.

\bibitem{guo2021longt5}
Mandy Guo, Joshua Ainslie, David Uthus, Santiago Ontanon, Jianmo Ni, Yun-Hsuan Sung, and Yinfei Yang.
\newblock Longt5: Efficient text-to-text transformer for long sequences.
\newblock {\em arXiv preprint arXiv:2112.07916}, 2021.

\bibitem{ding2023longnet}
Jiayu Ding, Shuming Ma, Li~Dong, Xingxing Zhang, Shaohan Huang, Wenhui Wang, Nanning Zheng, and Furu Wei.
\newblock Longnet: Scaling transformers to 1,000,000,000 tokens.
\newblock {\em arXiv preprint arXiv:2307.02486}, 2023.

\bibitem{ho2019axial}
Jonathan Ho, Nal Kalchbrenner, Dirk Weissenborn, and Tim Salimans.
\newblock Axial attention in multidimensional transformers.
\newblock {\em arXiv preprint arXiv:1912.12180}, 2019.

\bibitem{kitaev2020reformer}
Nikita Kitaev, {\L}ukasz Kaiser, and Anselm Levskaya.
\newblock Reformer: The efficient transformer.
\newblock {\em arXiv preprint arXiv:2001.04451}, 2020.

\bibitem{roy2021efficient}
Aurko Roy, Mohammad Saffar, Ashish Vaswani, and David Grangier.
\newblock Efficient content-based sparse attention with routing transformers.
\newblock {\em Transactions of the Association for Computational Linguistics}, 9:53--68, 2021.

\bibitem{tay2020sparse}
Yi~Tay, Dara Bahri, Liu Yang, Donald Metzler, and Da-Cheng Juan.
\newblock Sparse sinkhorn attention.
\newblock In {\em International Conference on Machine Learning}, pages 9438--9447. PMLR, 2020.

\bibitem{wu2022memorizing}
Yuhuai Wu, Markus~N Rabe, DeLesley Hutchins, and Christian Szegedy.
\newblock Memorizing transformers.
\newblock {\em arXiv preprint arXiv:2203.08913}, 2022.

\bibitem{bertsch2024unlimiformer}
Amanda Bertsch, Uri Alon, Graham Neubig, and Matthew Gormley.
\newblock Unlimiformer: Long-range transformers with unlimited length input.
\newblock {\em Advances in Neural Information Processing Systems}, 36, 2024.

\bibitem{yuan2025native}
Jingyang Yuan, Huazuo Gao, Damai Dai, Junyu Luo, Liang Zhao, Zhengyan Zhang, Zhenda Xie, YX~Wei, Lean Wang, Zhiping Xiao, et~al.
\newblock Native sparse attention: Hardware-aligned and natively trainable sparse attention.
\newblock {\em arXiv preprint arXiv:2502.11089}, 2025.

\bibitem{pikekos2025mixture}
Piotr Pi{\k{e}}kos, R{\'o}bert Csord{\'a}s, and J{\"u}rgen Schmidhuber.
\newblock Mixture of sparse attention: Content-based learnable sparse attention via expert-choice routing.
\newblock {\em arXiv preprint arXiv:2505.00315}, 2025.

\bibitem{wang2021spatten}
Hanrui Wang, Zhekai Zhang, and Song Han.
\newblock Spatten: Efficient sparse attention architecture with cascade token and head pruning.
\newblock In {\em 2021 IEEE International Symposium on High-Performance Computer Architecture (HPCA)}, pages 97--110. IEEE, 2021.

\bibitem{jiang2024minference}
Huiqiang Jiang, Yucheng Li, Chengruidong Zhang, Qianhui Wu, Xufang Luo, Surin Ahn, Zhenhua Han, Amir~H Abdi, Dongsheng Li, Chin-Yew Lin, et~al.
\newblock Minference 1.0: Accelerating pre-filling for long-context llms via dynamic sparse attention.
\newblock {\em arXiv preprint arXiv:2407.02490}, 2024.

\bibitem{gao2024seerattention}
Yizhao Gao, Zhichen Zeng, Dayou Du, Shijie Cao, Hayden Kwok-Hay So, Ting Cao, Fan Yang, and Mao Yang.
\newblock Seerattention: Learning intrinsic sparse attention in your llms.
\newblock {\em arXiv preprint arXiv:2410.13276}, 2024.

\bibitem{xiao2023efficient}
Guangxuan Xiao, Yuandong Tian, Beidi Chen, Song Han, and Mike Lewis.
\newblock Efficient streaming language models with attention sinks.
\newblock {\em arXiv preprint arXiv:2309.17453}, 2023.

\bibitem{zhang2024h2o}
Zhenyu Zhang, Ying Sheng, Tianyi Zhou, Tianlong Chen, Lianmin Zheng, Ruisi Cai, Zhao Song, Yuandong Tian, Christopher R{\'e}, Clark Barrett, et~al.
\newblock H2o: Heavy-hitter oracle for efficient generative inference of large language models.
\newblock {\em Advances in Neural Information Processing Systems}, 36, 2024.

\bibitem{ge2023fastgen}
Suyu Ge, Yunan Zhang, Liyuan Liu, Minjia Zhang, Jiawei Han, and Jianfeng Gao.
\newblock Model tells you what to discard: Adaptive kv cache compression for llms.
\newblock {\em arXiv preprint arXiv:2310.01801}, 2023.

\bibitem{tang2024quest}
Jiaming Tang, Yilong Zhao, Kan Zhu, Guangxuan Xiao, Baris Kasikci, and Song Han.
\newblock Quest: Query-aware sparsity for efficient long-context llm inference.
\newblock {\em arXiv preprint arXiv:2406.10774}, 2024.

\bibitem{lu2024longheads}
Yi~Lu, Xin Zhou, Wei He, Jun Zhao, Tao Ji, Tao Gui, Qi~Zhang, and Xuanjing Huang.
\newblock Longheads: Multi-head attention is secretly a long context processor.
\newblock {\em arXiv preprint arXiv:2402.10685}, 2024.

\bibitem{yang2025lserve}
Shang Yang, Junxian Guo, Haotian Tang, Qinghao Hu, Guangxuan Xiao, Jiaming Tang, Yujun Lin, Zhijian Liu, Yao Lu, and Song Han.
\newblock Lserve: Efficient long-sequence llm serving with unified sparse attention.
\newblock {\em arXiv preprint arXiv:2502.14866}, 2025.

\bibitem{xu2025xattention}
Ruyi Xu, Guangxuan Xiao, Haofeng Huang, Junxian Guo, and Song Han.
\newblock Xattention: Block sparse attention with antidiagonal scoring.
\newblock {\em arXiv preprint arXiv:2503.16428}, 2025.

\bibitem{dao2022flashattention}
Tri Dao, Dan Fu, Stefano Ermon, Atri Rudra, and Christopher R{\'e}.
\newblock Flashattention: Fast and memory-efficient exact attention with io-awareness.
\newblock {\em Advances in Neural Information Processing Systems}, 35:16344--16359, 2022.

\bibitem{dao2023flashattention}
Tri Dao.
\newblock Flashattention-2: Faster attention with better parallelism and work partitioning.
\newblock {\em arXiv preprint arXiv:2307.08691}, 2023.

\bibitem{lu2025moba}
Enzhe Lu, Zhejun Jiang, Jingyuan Liu, Yulun Du, Tao Jiang, Chao Hong, Shaowei Liu, Weiran He, Enming Yuan, Yuzhi Wang, et~al.
\newblock Moba: Mixture of block attention for long-context llms.
\newblock {\em arXiv preprint arXiv:2502.13189}, 2025.

\bibitem{shah2024flashattention}
Jay Shah, Ganesh Bikshandi, Ying Zhang, Vijay Thakkar, Pradeep Ramani, and Tri Dao.
\newblock Flashattention-3: Fast and accurate attention with asynchrony and low-precision.
\newblock {\em Advances in Neural Information Processing Systems}, 37:68658--68685, 2024.

\bibitem{shazeer2019fast}
Noam Shazeer.
\newblock Fast transformer decoding: One write-head is all you need.
\newblock {\em arXiv preprint arXiv:1911.02150}, 2019.

\bibitem{ainslie2023gqa}
Joshua Ainslie, James Lee-Thorp, Michiel De~Jong, Yury Zemlyanskiy, Federico Lebr{\'o}n, and Sumit Sanghai.
\newblock Gqa: Training generalized multi-query transformer models from multi-head checkpoints.
\newblock {\em arXiv preprint arXiv:2305.13245}, 2023.

\bibitem{zadouri2025hardware}
Ted Zadouri, Hubert Strauss, and Tri Dao.
\newblock Hardware-efficient attention for fast decoding.
\newblock {\em arXiv preprint arXiv:2505.21487}, 2025.

\bibitem{fu2024moa}
Tianyu Fu, Haofeng Huang, Xuefei Ning, Genghan Zhang, Boju Chen, Tianqi Wu, Hongyi Wang, Zixiao Huang, Shiyao Li, Shengen Yan, et~al.
\newblock Moa: Mixture of sparse attention for automatic large language model compression.
\newblock {\em arXiv preprint arXiv:2406.14909}, 2024.

\bibitem{Qu2024LLaMAMoEVE}
Xiaoye Qu, Daize Dong, Xuyang Hu, Tong Zhu, Weigao Sun, and Yu~Cheng.
\newblock Llama-moe v2: Exploring sparsity of llama from perspective of mixture-of-experts with post-training.
\newblock {\em ArXiv}, abs/2411.15708, 2024.

\bibitem{jin2024moh}
Peng Jin, Bo~Zhu, Li~Yuan, and Shuicheng Yan.
\newblock Moh: Multi-head attention as mixture-of-head attention.
\newblock {\em arXiv preprint arXiv:2410.11842}, 2024.

\bibitem{zhang2024sageattention}
Jintao Zhang, Jia Wei, Haofeng Huang, Pengle Zhang, Jun Zhu, and Jianfei Chen.
\newblock Sageattention: Accurate 8-bit attention for plug-and-play inference acceleration.
\newblock {\em arXiv preprint arXiv:2410.02367}, 2024.

\bibitem{zhang2024sageattention2}
Jintao Zhang, Haofeng Huang, Pengle Zhang, Jia Wei, Jun Zhu, and Jianfei Chen.
\newblock Sageattention2 technical report: Accurate 4 bit attention for plug-and-play inference acceleration.
\newblock {\em arXiv preprint arXiv:2411.10958}, 2024.

\bibitem{zhang2025sageattention3}
Jintao Zhang, Jia Wei, Pengle Zhang, Xiaoming Xu, Haofeng Huang, Haoxu Wang, Kai Jiang, Jun Zhu, and Jianfei Chen.
\newblock Sageattention3: Microscaling fp4 attention for inference and an exploration of 8-bit training.
\newblock {\em arXiv preprint arXiv:2505.11594}, 2025.

\bibitem{shen2020q}
Sheng Shen, Zhen Dong, Jiayu Ye, Linjian Ma, Zhewei Yao, Amir Gholami, Michael~W Mahoney, and Kurt Keutzer.
\newblock Q-bert: Hessian based ultra low precision quantization of bert.
\newblock In {\em Proceedings of the AAAI Conference on Artificial Intelligence}, volume~34, pages 8815--8821, 2020.

\bibitem{kim2021bert}
Sehoon Kim, Amir Gholami, Zhewei Yao, Michael~W Mahoney, and Kurt Keutzer.
\newblock I-bert: Integer-only bert quantization.
\newblock In {\em International conference on machine learning}, pages 5506--5518. PMLR, 2021.

\bibitem{chen2024int}
Shimao Chen, Zirui Liu, Zhiying Wu, Ce~Zheng, Peizhuang Cong, Zihan Jiang, Yuhan Wu, Lei Su, and Tong Yang.
\newblock Int-flashattention: Enabling flash attention for int8 quantization.
\newblock {\em arXiv preprint arXiv:2409.16997}, 2024.

\bibitem{zafrir2019q8bert}
Ofir Zafrir, Guy Boudoukh, Peter Izsak, and Moshe Wasserblat.
\newblock Q8bert: Quantized 8bit bert.
\newblock In {\em 2019 Fifth Workshop on Energy Efficient Machine Learning and Cognitive Computing-NeurIPS Edition (EMC2-NIPS)}, pages 36--39. IEEE, 2019.

\bibitem{prato2019fully}
Gabriele Prato, Ella Charlaix, and Mehdi Rezagholizadeh.
\newblock Fully quantized transformer for machine translation.
\newblock {\em arXiv preprint arXiv:1910.10485}, 2019.

\bibitem{kang2024turboattention}
Hao Kang, Srikant Bharadwaj, James Hensman, Tushar Krishna, Victor Ruhle, and Saravan Rajmohan.
\newblock Turboattention: Efficient attention approximation for high throughputs llms.
\newblock {\em arXiv preprint arXiv:2412.08585}, 2024.

\bibitem{zhang2025hack}
Zeyu Zhang, Haiying Shen, Shay Vargaftik, Ran~Ben Basat, Michael Mitzenmacher, and Minlan Yu.
\newblock Hack: Homomorphic acceleration via compression of the key-value cache for disaggregated llm inference.
\newblock {\em arXiv preprint arXiv:2502.03589}, 2025.

\bibitem{du2024bitdistiller}
Dayou Du, Yijia Zhang, Shijie Cao, Jiaqi Guo, Ting Cao, Xiaowen Chu, and Ningyi Xu.
\newblock Bitdistiller: Unleashing the potential of sub-4-bit llms via self-distillation.
\newblock {\em arXiv preprint arXiv:2402.10631}, 2024.

\bibitem{zhou2022mixture}
Yanqi Zhou, Tao Lei, Hanxiao Liu, Nan Du, Yanping Huang, Vincent Zhao, Andrew~M Dai, Quoc~V Le, James Laudon, et~al.
\newblock Mixture-of-experts with expert choice routing.
\newblock {\em Advances in Neural Information Processing Systems}, 35:7103--7114, 2022.

\bibitem{Lewis2021BASELS}
Mike Lewis, Shruti Bhosale, Tim Dettmers, Naman Goyal, and Luke Zettlemoyer.
\newblock Base layers: Simplifying training of large, sparse models.
\newblock {\em ArXiv}, abs/2103.16716, 2021.

\bibitem{Roller2021HashLF}
Stephen Roller, Sainbayar Sukhbaatar, Arthur Szlam, and Jason Weston.
\newblock Hash layers for large sparse models.
\newblock In {\em Neural Information Processing Systems}, 2021.

\bibitem{Huang2024HarderTN}
Quzhe Huang, Zhenwei An, Zhuang Nan, Mingxu Tao, Chen Zhang, Yang Jin, Kun Xu, Liwei Chen, Songfang Huang, and Yansong Feng.
\newblock Harder tasks need more experts: Dynamic routing in moe models.
\newblock {\em ArXiv}, abs/2403.07652, 2024.

\bibitem{Guo2024DynamicMO}
Yongxin Guo, Zhenglin Cheng, Xiaoying Tang, and Tao Lin.
\newblock Dynamic mixture of experts: An auto-tuning approach for efficient transformer models.
\newblock {\em ArXiv}, abs/2405.14297, 2024.

\bibitem{Zeng2024AdaMoETR}
Zihao Zeng, Yibo Miao, Hongcheng Gao, Hao Zhang, and Zhijie Deng.
\newblock Adamoe: Token-adaptive routing with null experts for mixture-of-experts language models.
\newblock {\em ArXiv}, abs/2406.13233, 2024.

\bibitem{Yue2024AdaKRB}
Tongtian Yue, Longteng Guo, Jie Cheng, Xuange Gao, and Jing Liu.
\newblock Ada-k routing: Boosting the efficiency of moe-based llms.
\newblock {\em ArXiv}, abs/2410.10456, 2024.

\bibitem{Wang2024AuxiliaryLossFreeLB}
Lean Wang, Huazuo Gao, Chenggang Zhao, Xu~Sun, and Damai Dai.
\newblock Auxiliary-loss-free load balancing strategy for mixture-of-experts.
\newblock {\em ArXiv}, abs/2408.15664, 2024.

\bibitem{Qiu2025DemonsIT}
Zihan Qiu, Zeyu Huang, Bo~Zheng, Kaiyue Wen, Zekun Wang, Rui Men, Ivan Titov, Dayiheng Liu, Jingren Zhou, and Junyang Lin.
\newblock Demons in the detail: On implementing load balancing loss for training specialized mixture-of-expert models.
\newblock {\em ArXiv}, abs/2501.11873, 2025.

\bibitem{dai2024deepseekmoe}
Damai Dai, Chengqi Deng, Chenggang Zhao, RX~Xu, Huazuo Gao, Deli Chen, Jiashi Li, Wangding Zeng, Xingkai Yu, Yu~Wu, et~al.
\newblock Deepseekmoe: Towards ultimate expert specialization in mixture-of-experts language models.
\newblock {\em arXiv preprint arXiv:2401.06066}, 2024.

\bibitem{Rajbhandari2022DeepSpeedMoEAM}
Samyam Rajbhandari, Conglong Li, Zhewei Yao, Minjia Zhang, Reza~Yazdani Aminabadi, Ammar~Ahmad Awan, Jeff Rasley, and Yuxiong He.
\newblock Deepspeed-moe: Advancing mixture-of-experts inference and training to power next-generation ai scale.
\newblock {\em ArXiv}, abs/2201.05596, 2022.

\bibitem{muennighoff2024olmoe}
Niklas Muennighoff, Luca Soldaini, Dirk Groeneveld, Kyle Lo, Jacob Morrison, Sewon Min, Weijia Shi, Pete Walsh, Oyvind Tafjord, Nathan Lambert, et~al.
\newblock Olmoe: Open mixture-of-experts language models.
\newblock {\em arXiv preprint arXiv:2409.02060}, 2024.

\bibitem{Raposo2024MixtureofDepthsDA}
David Raposo, Sam Ritter, Blake Richards, Timothy~P. Lillicrap, Peter Humphreys, and Adam Santoro.
\newblock Mixture-of-depths: Dynamically allocating compute in transformer-based language models.
\newblock {\em ArXiv}, abs/2404.02258, 2024.

\bibitem{Zuo2022MoEBERTFB}
Simiao Zuo, Qingru Zhang, Chen Liang, Pengcheng He, Tuo Zhao, and Weizhu Chen.
\newblock Moebert: from bert to mixture-of-experts via importance-guided adaptation.
\newblock In {\em North American Chapter of the Association for Computational Linguistics}, 2022.

\bibitem{Zhang2021MoEficationTF}
Zhengyan Zhang, Yankai Lin, Zhiyuan Liu, Peng Li, Maosong Sun, and Jie Zhou.
\newblock Moefication: Transformer feed-forward layers are mixtures of experts.
\newblock In {\em Findings}, 2021.

\bibitem{zhu-etal-2024-llama}
Tong Zhu, Xiaoye Qu, Daize Dong, Jiacheng Ruan, Jingqi Tong, Conghui He, and Yu~Cheng.
\newblock {LL}a{MA}-{M}o{E}: Building mixture-of-experts from {LL}a{MA} with continual pre-training.
\newblock In Yaser Al-Onaizan, Mohit Bansal, and Yun-Nung Chen, editors, {\em Proceedings of the 2024 Conference on Empirical Methods in Natural Language Processing}, pages 15913--15923, Miami, Florida, USA, November 2024. Association for Computational Linguistics.

\bibitem{Komatsuzaki2022SparseUT}
Aran Komatsuzaki, Joan Puigcerver, James Lee-Thorp, Carlos~Riquelme Ruiz, Basil Mustafa, Joshua Ainslie, Yi~Tay, Mostafa Dehghani, and Neil Houlsby.
\newblock Sparse upcycling: Training mixture-of-experts from dense checkpoints.
\newblock {\em ArXiv}, abs/2212.05055, 2022.

\bibitem{Li2022BranchTrainMergeEP}
Margaret Li, Suchin Gururangan, Tim Dettmers, Mike Lewis, Tim Althoff, Noah~A. Smith, and Luke Zettlemoyer.
\newblock Branch-train-merge: Embarrassingly parallel training of expert language models.
\newblock {\em ArXiv}, abs/2208.03306, 2022.

\bibitem{Sukhbaatar2024BranchTrainMiXME}
Sainbayar Sukhbaatar, Olga Golovneva, Vasu Sharma, Hu~Xu, Xi~Victoria Lin, Baptiste Rozi{\`e}re, Jacob Kahn, Shang-Wen Li, Wen tau Yih, Jason~E Weston, and Xian Li.
\newblock Branch-train-mix: Mixing expert llms into a mixture-of-experts llm.
\newblock {\em ArXiv}, abs/2403.07816, 2024.

\bibitem{zamba}
Paolo Glorioso, Quentin Anthony, Yury Tokpanov, James Whittington, Jonathan Pilault, Adam Ibrahim, and Beren Millidge.
\newblock Zamba: A compact 7b ssm hybrid model.
\newblock {\em arXiv preprint arXiv:2405.16712}, 2024.

\bibitem{glorioso2024zamba2}
Paolo Glorioso, Quentin Anthony, Yury Tokpanov, Anna Golubeva, Vasudev Shyam, James Whittington, Jonathan Pilault, and Beren Millidge.
\newblock The zamba2 suite: Technical report.
\newblock {\em arXiv preprint arXiv:2411.15242}, 2024.

\bibitem{samba}
Liliang Ren, Yang Liu, Yadong Lu, Yelong Shen, Chen Liang, and Weizhu Chen.
\newblock Samba: Simple hybrid state space models for efficient unlimited context language modeling.
\newblock {\em arXiv preprint arXiv:2406.07522}, 2024.

\bibitem{jamba}
Opher Lieber, Barak Lenz, Hofit Bata, Gal Cohen, Jhonathan Osin, Itay Dalmedigos, Erez Safahi, Shaked Meirom, Yonatan Belinkov, Shai Shalev-Shwartz, et~al.
\newblock Jamba: A hybrid transformer-mamba language model.
\newblock {\em arXiv preprint arXiv:2403.19887}, 2024.

\bibitem{rwkv-x}
Haowen Hou, Zhiyi Huang, Kaifeng Tan, Rongchang Lu, and Fei~Richard Yu.
\newblock {RWKV}-{X}: {A} {Linear} {Complexity} {Hybrid} {Language} {Model}, May 2025.

\bibitem{yang2025zebra}
Mingyu Yang, Mehdi Rezagholizadeh, Guihong Li, Vikram Appia, and Emad Barsoum.
\newblock Zebra-llama: Towards extremely efficient hybrid models.
\newblock {\em arXiv preprint arXiv:2505.17272}, 2025.

\bibitem{sun2024yoco}
Yutao Sun, Li~Dong, Yi~Zhu, Shaohan Huang, Wenhui Wang, Shuming Ma, Quanlu Zhang, Jianyong Wang, and Furu Wei.
\newblock You only cache once: Decoder-decoder architectures for language models.
\newblock {\em Advances in Neural Information Processing Systems}, 37:7339--7361, 2024.

\bibitem{botev2024recurrentgemma}
Aleksandar Botev, Soham De, Samuel~L Smith, Anushan Fernando, George-Cristian Muraru, Ruba Haroun, Leonard Berrada, Razvan Pascanu, Pier~Giuseppe Sessa, Robert Dadashi, et~al.
\newblock Recurrentgemma: Moving past transformers for efficient open language models.
\newblock {\em arXiv preprint arXiv:2404.07839}, 2024.

\bibitem{zhang2025test}
Tianyuan Zhang, Sai Bi, Yicong Hong, Kai Zhang, Fujun Luan, Songlin Yang, Kalyan Sunkavalli, William~T Freeman, and Hao Tan.
\newblock Test-time training done right.
\newblock {\em arXiv preprint arXiv:2505.23884}, 2025.

\bibitem{hymba}
Xin Dong, Yonggan Fu, Shizhe Diao, Wonmin Byeon, Zijia Chen, Ameya~Sunil Mahabaleshwarkar, Shih-Yang Liu, Matthijs Van~Keirsbilck, Min-Hung Chen, Yoshi Suhara, et~al.
\newblock Hymba: A hybrid-head architecture for small language models.
\newblock {\em arXiv preprint arXiv:2411.13676}, 2024.

\bibitem{transmamba}
Yixing Li, Ruobing Xie, Zhen Yang, Xingwu Sun, Shuaipeng Li, Weidong Han, Zhanhui Kang, Yu~Cheng, Chengzhong Xu, Di~Wang, et~al.
\newblock Transmamba: Flexibly switching between transformer and mamba.
\newblock {\em arXiv preprint arXiv:2503.24067}, 2025.

\bibitem{mcdermott2025lola}
Luke McDermott, Robert~W Heath~Jr, and Rahul Parhi.
\newblock Lola: Low-rank linear attention with sparse caching.
\newblock {\em arXiv preprint arXiv:2505.23666}, 2025.

\bibitem{nie2025largelanguagediffusionmodels}
Shen Nie, Fengqi Zhu, Zebin You, Xiaolu Zhang, Jingyang Ou, Jun Hu, Jun Zhou, Yankai Lin, Ji-Rong Wen, and Chongxuan Li.
\newblock Large language diffusion models, 2025.

\bibitem{li2022diffusionlmimprovescontrollabletext}
Xiang~Lisa Li, John Thickstun, Ishaan Gulrajani, Percy Liang, and Tatsunori~B. Hashimoto.
\newblock Diffusion-lm improves controllable text generation, 2022.

\bibitem{gong2023diffuseqsequencesequencetext}
Shansan Gong, Mukai Li, Jiangtao Feng, Zhiyong Wu, and Lingpeng Kong.
\newblock Diffuseq: Sequence to sequence text generation with diffusion models, 2023.

\bibitem{lou2024discretediffusionmodelingestimating}
Aaron Lou, Chenlin Meng, and Stefano Ermon.
\newblock Discrete diffusion modeling by estimating the ratios of the data distribution, 2024.

\bibitem{gulrajani2023likelihoodbaseddiffusionlanguagemodels}
Ishaan Gulrajani and Tatsunori~B. Hashimoto.
\newblock Likelihood-based diffusion language models, 2023.

\bibitem{arriola2025blockdiffusioninterpolatingautoregressive}
Marianne Arriola, Aaron Gokaslan, Justin~T Chiu, Zhihan Yang, Zhixuan Qi, Jiaqi Han, Subham~Sekhar Sahoo, and Volodymyr Kuleshov.
\newblock Block diffusion: Interpolating between autoregressive and diffusion language models, 2025.

\bibitem{gong2025scalingdiffusionlanguagemodels}
Shansan Gong, Shivam Agarwal, Yizhe Zhang, Jiacheng Ye, Lin Zheng, Mukai Li, Chenxin An, Peilin Zhao, Wei Bi, Jiawei Han, Hao Peng, and Lingpeng Kong.
\newblock Scaling diffusion language models via adaptation from autoregressive models, 2025.

\bibitem{you2025lladavlargelanguagediffusion}
Zebin You, Shen Nie, Xiaolu Zhang, Jun Hu, Jun Zhou, Zhiwu Lu, Ji-Rong Wen, and Chongxuan Li.
\newblock Llada-v: Large language diffusion models with visual instruction tuning, 2025.

\bibitem{swerdlow2025unified}
Alexander Swerdlow, Mihir Prabhudesai, Siddharth Gandhi, Deepak Pathak, and Katerina Fragkiadaki.
\newblock Unified multimodal discrete diffusion.
\newblock {\em arXiv preprint arXiv:2503.20853}, 2025.

\bibitem{li2025lavida}
Shufan Li, Konstantinos Kallidromitis, Hritik Bansal, Akash Gokul, Yusuke Kato, Kazuki Kozuka, Jason Kuen, Zhe Lin, Kai-Wei Chang, and Aditya Grover.
\newblock Lavida: A large diffusion language model for multimodal understanding.
\newblock {\em arXiv preprint arXiv:2505.16839}, 2025.

\bibitem{yang2025mmadamultimodallargediffusion}
Ling Yang, Ye~Tian, Bowen Li, Xinchen Zhang, Ke~Shen, Yunhai Tong, and Mengdi Wang.
\newblock Mmada: Multimodal large diffusion language models, 2025.

\bibitem{liao2025vig}
Bencheng Liao, Xinggang Wang, Lianghui Zhu, Qian Zhang, and Chang Huang.
\newblock Vig: Linear-complexity visual sequence learning with gated linear attention.
\newblock In {\em Proceedings of the AAAI Conference on Artificial Intelligence}, volume~39, pages 5182--5190, 2025.

\bibitem{duan2024vision}
Yuchen Duan, Weiyun Wang, Zhe Chen, Xizhou Zhu, Lewei Lu, Tong Lu, Yu~Qiao, Hongsheng Li, Jifeng Dai, and Wenhai Wang.
\newblock Vision-rwkv: Efficient and scalable visual perception with rwkv-like architectures.
\newblock {\em arXiv preprint arXiv:2403.02308}, 2024.

\bibitem{hwang2023tutel}
Changho Hwang, Wei Cui, Yifan Xiong, Ziyue Yang, Ze~Liu, Han Hu, Zilong Wang, Rafael Salas, Jithin Jose, Prabhat Ram, et~al.
\newblock Tutel: Adaptive mixture-of-experts at scale.
\newblock {\em Proceedings of Machine Learning and Systems}, 5:269--287, 2023.

\bibitem{wang2025insectmamba}
Qianning Wang, Chenglin Wang, Zhixin Lai, and Yucheng Zhou.
\newblock Insectmamba: State space model with adaptive composite features for insect recognition.
\newblock In {\em ICASSP 2025-2025 IEEE International Conference on Acoustics, Speech and Signal Processing (ICASSP)}, pages 1--5. IEEE, 2025.

\bibitem{zhang2024voxel}
Guowen Zhang, Lue Fan, Chenhang He, Zhen Lei, ZHAO-XIANG ZHANG, and Lei Zhang.
\newblock Voxel mamba: Group-free state space models for point cloud based 3d object detection.
\newblock {\em Advances in Neural Information Processing Systems}, 37:81489--81509, 2024.

\bibitem{mo2024scaling}
Shentong Mo and Yapeng Tian.
\newblock Scaling diffusion mamba with bidirectional ssms for efficient image and video generation.
\newblock {\em arXiv preprint arXiv:2405.15881}, 2024.

\bibitem{ma2024u}
Jun Ma, Feifei Li, and Bo~Wang.
\newblock U-mamba: Enhancing long-range dependency for biomedical image segmentation.
\newblock {\em arXiv preprint arXiv:2401.04722}, 2024.

\bibitem{ruan2024vm}
Jiacheng Ruan, Jincheng Li, and Suncheng Xiang.
\newblock Vm-unet: Vision mamba unet for medical image segmentation.
\newblock {\em arXiv preprint arXiv:2402.02491}, 2024.

\bibitem{jiang2025rwkv}
Juntao Jiang, Jiangning Zhang, Weixuan Liu, Muxuan Gao, Xiaobin Hu, Xiaoxiao Yan, Feiyue Huang, and Yong Liu.
\newblock Rwkv-unet: Improving unet with long-range cooperation for effective medical image segmentation.
\newblock {\em arXiv preprint arXiv:2501.08458}, 2025.

\bibitem{you2024mambabev}
Zihan You, Ni~Wang, Hao Wang, Qichao Zhao, and Jinxiang Wang.
\newblock Mambabev: An efficient 3d detection model with mamba2.
\newblock {\em arXiv preprint arXiv:2410.12673}, 2024.

\bibitem{lin2024audio}
Jiaju Lin and Haoxuan Hu.
\newblock Audio mamba: Pretrained audio state space model for audio tagging.
\newblock {\em arXiv preprint arXiv:2405.13636}, 2024.

\bibitem{zhang2024MAMC}
Yezhuo Zhang, Zinan Zhou, Yichao Cao, Guangyu Li, and Xuanpeng Li.
\newblock Mamca -- optimal on accuracy and efficiency for automatic modulation classification with extended signal length.
\newblock {\em arXiv preprint arXiv:2405.11263}, 2024.
\newblock Also accepted in IEEE Communications Letters (Early Access).

\bibitem{chen2024rawbmamba}
Yujie Chen, Jiangyan Yi, Jun Xue, Chenglong Wang, Xiaohui Zhang, Shunbo Dong, Siding Zeng, Jianhua Tao, Lv~Zhao, and Cunhang Fan.
\newblock Rawbmamba: End-to-end bidirectional state space model for audio deepfake detection.
\newblock {\em arXiv preprint arXiv:2406.06086}, 2024.

\bibitem{goel2022its_raw}
Karan Goel, Albert Gu, Chris Donahue, and Christopher Re.
\newblock It’s raw! audio generation with state‑space models.
\newblock In {\em Proceedings of the 39th International Conference on Machine Learning (ICML 2022)}, volume 162 of {\em Proceedings of Machine Learning Research}, pages 7616--7633, 2022.

\bibitem{liu2024perturbing}
Shipei Liu, Xiaoya Fan, and Guowei Wu.
\newblock Why perturbing symbolic music is necessary: Fitting the distribution of never-used notes through a joint probabilistic diffusion model.
\newblock {\em arXiv preprint arXiv:2408.01950}, 2024.

\bibitem{zhang2024mamba}
Xiangyu Zhang, Qiquan Zhang, Hexin Liu, Tianyi Xiao, Xinyuan Qian, Beena Ahmed, Eliathamby Ambikairajah, Haizhou Li, and Julien Epps.
\newblock Mamba in speech: Towards an alternative to self-attention.
\newblock {\em arXiv preprint arXiv:2405.12609}, 2024.

\bibitem{jiang2024dual}
Xilin Jiang, Cong Han, and Nima Mesgarani.
\newblock Dual-path mamba: Short and long-term bidirectional selective structured state space models for speech separation.
\newblock arXiv preprint arXiv:2403.18257, 2024.

\bibitem{li2024spmamba}
Kai Li, Chen Guo, and Hu~Xiaolin.
\newblock Spmamba: State-space model is all you need in speech separation.
\newblock arXiv preprint arXiv:2404.02063, 2024.

\bibitem{zuo2023advancing}
Lingyun Zuo, Keyu An, Shiliang Zhang, and Zhijie Yan.
\newblock Advancing vad systems based on multi-task learning with improved model structures.
\newblock {\em arXiv preprint arXiv:2312.14860}, 2023.

\bibitem{li2024mamba}
Xinran Li, Xiaomao Fan, Qingyang Wu, Xiaojiang Peng, and Ye~Li.
\newblock Mamba-enhanced text-audio-video alignment network for emotion recognition in conversations.
\newblock {\em arXiv preprint arXiv:2409.05243}, 2024.

\bibitem{gong2025avs}
Sitong Gong, Yunzhi Zhuge, Lu~Zhang, Yifan Wang, Pingping Zhang, Lijun Wang, and Huchuan Lu.
\newblock Avs-mamba: Exploring temporal and multi-modal mamba for audio-visual segmentation.
\newblock {\em arXiv preprint arXiv:2501.07810}, 2025.

\bibitem{li2024visualrwkv}
Zihang Li and Haowen Hou.
\newblock Visualrwkv-hd and uhd: Advancing high-resolution processing for visual language models.
\newblock {\em arXiv preprint arXiv:2410.11665}, 2024.

\bibitem{shen2023scaling}
Sheng Shen, Zhewei Yao, Chunyuan Li, Trevor Darrell, Kurt Keutzer, and Yuxiong He.
\newblock Scaling vision-language models with sparse mixture of experts.
\newblock {\em arXiv preprint arXiv:2303.07226}, 2023.

\bibitem{lin2024moe}
Bin Lin, Zhenyu Tang, Yang Ye, Jiaxi Cui, Bin Zhu, Peng Jin, Jinfa Huang, Junwu Zhang, Yatian Pang, Munan Ning, et~al.
\newblock Moe-llava: Mixture of experts for large vision-language models.
\newblock {\em arXiv preprint arXiv:2401.15947}, 2024.

\bibitem{gou2023mixture}
Yunhao Gou, Zhili Liu, Kai Chen, Lanqing Hong, Hang Xu, Aoxue Li, Dit-Yan Yeung, James~T Kwok, and Yu~Zhang.
\newblock Mixture of cluster-conditional lora experts for vision-language instruction tuning.
\newblock {\em arXiv preprint arXiv:2312.12379}, 2023.

\bibitem{chen2024llava}
Shaoxiang Chen, Zequn Jie, and Lin Ma.
\newblock Llava-mole: Sparse mixture of lora experts for mitigating data conflicts in instruction finetuning mllms.
\newblock {\em arXiv preprint arXiv:2401.16160}, 2024.

\bibitem{tay2022efficient}
Yi~Tay, Mostafa Dehghani, Dara Bahri, and Donald Metzler.
\newblock Efficient transformers: A survey.
\newblock {\em ACM Computing Surveys}, 55(6):1--28, 2022.

\bibitem{patro2024mamba}
Badri~Narayana Patro and Vijay~Srinivas Agneeswaran.
\newblock Mamba-360: Survey of state space models as transformer alternative for long sequence modelling: Methods, applications, and challenges.
\newblock {\em arXiv preprint arXiv:2404.16112}, 2024.

\bibitem{tiezzi2025back}
Matteo Tiezzi, Michele Casoni, Alessandro Betti, Tommaso Guidi, Marco Gori, and Stefano Melacci.
\newblock Back to recurrent processing at the crossroad of transformers and state-space models.
\newblock {\em Nature Machine Intelligence}, pages 1--11, 2025.

\bibitem{sun2025efficient}
Yutao Sun, Zhenyu Li, Yike Zhang, Tengyu Pan, Bowen Dong, Yuyi Guo, and Jianyong Wang.
\newblock Efficient attention mechanisms for large language models: A survey.
\newblock {\em arXiv preprint arXiv:2507.19595}, 2025.

\bibitem{peng2021random}
Hao Peng, Nikolaos Pappas, Dani Yogatama, Roy Schwartz, Noah~A Smith, and Lingpeng Kong.
\newblock Random feature attention.
\newblock {\em arXiv preprint arXiv:2103.02143}, 2021.

\bibitem{rahimi2007random}
Ali Rahimi and Benjamin Recht.
\newblock Random features for large-scale kernel machines.
\newblock {\em Advances in neural information processing systems}, 20, 2007.

\bibitem{xiong2021nystromformer}
Yunyang Xiong, Zhanpeng Zeng, Rudrasis Chakraborty, Mingxing Tan, Glenn Fung, Yin Li, and Vikas Singh.
\newblock Nystr{\"o}mformer: A nystr{\"o}m-based algorithm for approximating self-attention.
\newblock In {\em Proceedings of the AAAI conference on artificial intelligence}, volume~35, pages 14138--14148, 2021.

\bibitem{chen2021skyformer}
Yifan Chen, Qi~Zeng, Heng Ji, and Yun Yang.
\newblock Skyformer: Remodel self-attention with gaussian kernel and nystr$\backslash$" om method.
\newblock {\em Advances in Neural Information Processing Systems}, 34:2122--2135, 2021.

\bibitem{han2023flatten}
Dongchen Han, Xuran Pan, Yizeng Han, Shiji Song, and Gao Huang.
\newblock Flatten transformer: Vision transformer using focused linear attention.
\newblock In {\em Proceedings of the IEEE/CVF international conference on computer vision}, pages 5961--5971, 2023.

\bibitem{zhang2024hedgehog}
Michael Zhang, Kush Bhatia, Hermann Kumbong, and Christopher R{\'e}.
\newblock The hedgehog \& the porcupine: Expressive linear attentions with softmax mimicry.
\newblock {\em arXiv preprint arXiv:2402.04347}, 2024.

\bibitem{qin2022devil}
Zhen Qin, Xiaodong Han, Weixuan Sun, Dongxu Li, Lingpeng Kong, Nick Barnes, and Yiran Zhong.
\newblock The devil in linear transformer.
\newblock {\em arXiv preprint arXiv:2210.10340}, 2022.

\bibitem{lu2025regla}
Peng Lu, Ivan Kobyzev, Mehdi Rezagholizadeh, Boxing Chen, and Philippe Langlais.
\newblock Regla: Refining gated linear attention.
\newblock {\em arXiv preprint arXiv:2502.01578}, 2025.

\bibitem{qin2023transnormerllm}
Zhen Qin, Dong Li, Weigao Sun, Weixuan Sun, Xuyang Shen, Xiaodong Han, Yunshen Wei, Baohong Lv, Xiao Luo, Yu~Qiao, et~al.
\newblock Transnormerllm: A faster and better large language model with improved transnormer.
\newblock {\em arXiv preprint arXiv:2307.14995}, 2023.

\bibitem{qin2024lightning}
Zhen Qin, Weigao Sun, Dong Li, Xuyang Shen, Weixuan Sun, and Yiran Zhong.
\newblock Lightning attention-2: A free lunch for handling unlimited sequence lengths in large language models.
\newblock {\em arXiv preprint arXiv:2401.04658}, 2024.

\bibitem{sun2023retentive}
Yutao Sun, Li~Dong, Shaohan Huang, Shuming Ma, Yuqing Xia, Jilong Xue, Jianyong Wang, and Furu Wei.
\newblock Retentive network: A successor to transformer for large language models.
\newblock {\em arXiv preprint arXiv:2307.08621}, 2023.

\bibitem{chou2024metala}
Yuhong Chou, Man Yao, Kexin Wang, Yuqi Pan, Rui-Jie Zhu, Jibin Wu, Yiran Zhong, Yu~Qiao, Bo~Xu, and Guoqi Li.
\newblock Metala: Unified optimal linear approximation to softmax attention map.
\newblock {\em Advances in Neural Information Processing Systems}, 37:71034--71067, 2024.

\bibitem{chakraverty2019hebbian}
Snehashish Chakraverty, Deepti~Moyi Sahoo, Nisha~Rani Mahato, Snehashish Chakraverty, Deepti~Moyi Sahoo, and Nisha~Rani Mahato.
\newblock Hebbian learning rule.
\newblock {\em Concepts of Soft Computing: Fuzzy and ANN with Programming}, pages 175--182, 2019.

\bibitem{prados1989neural}
DL~Prados and SC~Kak.
\newblock Neural network capacity using delta rule.
\newblock {\em Electronics Letters}, 25(3):197--199, 1989.

\bibitem{widrow1988adaptive}
Bernard Widrow and Marcian~E Hoff.
\newblock Adaptive switching circuits, 1988.

\bibitem{von2023uncovering}
Johannes Von~Oswald, Maximilian Schlegel, Alexander Meulemans, Seijin Kobayashi, Eyvind Niklasson, Nicolas Zucchet, Nino Scherrer, Nolan Miller, Mark Sandler, Max Vladymyrov, et~al.
\newblock Uncovering mesa-optimization algorithms in transformers.
\newblock {\em arXiv preprint arXiv:2309.05858}, 2023.

\bibitem{guo2025log}
Han Guo, Songlin Yang, Tarushii Goel, Eric~P Xing, Tri Dao, and Yoon Kim.
\newblock Log-linear attention.
\newblock {\em arXiv preprint arXiv:2506.04761}, 2025.

\bibitem{yau2025sequential}
Morris Yau, Sharut Gupta, Valerie Engelmayer, Kazuki Irie, Stefanie Jegelka, and Jacob Andreas.
\newblock Sequential-parallel duality in prefix scannable models.
\newblock {\em arXiv preprint arXiv:2506.10918}, 2025.

\bibitem{hu2024attractor}
Jiaxi Hu, Yuehong Hu, Wei Chen, Ming Jin, Shirui Pan, Qingsong Wen, and Yuxuan Liang.
\newblock Attractor memory for long-term time series forecasting: A chaos perspective.
\newblock {\em arXiv preprint arXiv:2402.11463}, 2024.

\bibitem{blelloch1990prefix}
Guy~E Blelloch.
\newblock Prefix sums and their applications, 1990.

\bibitem{chung2014empirical}
Junyoung Chung, Caglar Gulcehre, KyungHyun Cho, and Yoshua Bengio.
\newblock Empirical evaluation of gated recurrent neural networks on sequence modeling.
\newblock {\em arXiv preprint arXiv:1412.3555}, 2014.

\bibitem{zhai2021attention}
Shuangfei Zhai, Walter Talbott, Nitish Srivastava, Chen Huang, Hanlin Goh, Ruixiang Zhang, and Josh Susskind.
\newblock An attention free transformer.
\newblock {\em arXiv preprint arXiv:2105.14103}, 2021.

\bibitem{kalman1960new}
Rudolph~Emil Kalman.
\newblock A new approach to linear filtering and prediction problems, 1960.

\bibitem{glasser1985control}
William Glasser.
\newblock {\em Control theory}.
\newblock Harper and Row New York, 1985.

\bibitem{he2016deep}
Kaiming He, Xiangyu Zhang, Shaoqing Ren, and Jian Sun.
\newblock Deep residual learning for image recognition.
\newblock In {\em Proceedings of the IEEE conference on computer vision and pattern recognition}, pages 770--778, 2016.

\bibitem{fu2022hungry}
Daniel~Y Fu, Tri Dao, Khaled~K Saab, Armin~W Thomas, Atri Rudra, and Christopher R{\'e}.
\newblock Hungry hungry hippos: Towards language modeling with state space models.
\newblock {\em arXiv preprint arXiv:2212.14052}, 2022.

\bibitem{joffrain2006accumulating}
Thierry Joffrain, Tze~Meng Low, Enrique~S Quintana-Ort{\'\i}, Robert van~de Geijn, and Field G~Van Zee.
\newblock Accumulating householder transformations, revisited.
\newblock {\em ACM Transactions on Mathematical Software (TOMS)}, 32(2):169--179, 2006.

\bibitem{bischof1987wy}
Christian Bischof and Charles Van~Loan.
\newblock The wy representation for products of householder matrices.
\newblock {\em SIAM Journal on Scientific and Statistical Computing}, 8(1):s2--s13, 1987.

\bibitem{munkhdalai2017neural}
Tsendsuren Munkhdalai and Hong Yu.
\newblock Neural semantic encoders.
\newblock In {\em Proceedings of the conference. Association for Computational Linguistics. Meeting}, volume~1, page 397, 2017.

\bibitem{munkhdalai2019metalearned}
Tsendsuren Munkhdalai, Alessandro Sordoni, Tong Wang, and Adam Trischler.
\newblock Metalearned neural memory.
\newblock {\em Advances in Neural Information Processing Systems}, 32, 2019.

\bibitem{irie2021going}
Kazuki Irie, Imanol Schlag, R{\'o}bert Csord{\'a}s, and J{\"u}rgen Schmidhuber.
\newblock Going beyond linear transformers with recurrent fast weight programmers.
\newblock {\em Advances in neural information processing systems}, 34:7703--7717, 2021.

\bibitem{schlag2021linear}
Imanol Schlag, Kazuki Irie, and J{\"u}rgen Schmidhuber.
\newblock Linear transformers are secretly fast weight programmers.
\newblock In {\em International Conference on Machine Learning}, pages 9355--9366. PMLR, 2021.

\bibitem{jordan2024muon}
Keller Jordan, Yuchen Jin, Vlado Boza, Jiacheng You, Franz Cesista, Laker Newhouse, and Jeremy Bernstein.
\newblock Muon: An optimizer for hidden layers in neural networks.
\newblock {\em Cited on}, page~10, 2024.

\bibitem{wang2025test}
Ke~Alexander Wang, Jiaxin Shi, and Emily~B Fox.
\newblock Test-time regression: a unifying framework for designing sequence models with associative memory.
\newblock {\em arXiv preprint arXiv:2501.12352}, 2025.

\bibitem{siems2025deltaproduct}
Julien Siems, Timur Carstensen, Arber Zela, Frank Hutter, Massimiliano Pontil, and Riccardo Grazzi.
\newblock Deltaproduct: Improving state-tracking in linear rnns via householder products.
\newblock {\em arXiv preprint arXiv:2502.10297}, 2025.

\bibitem{zancato2024b}
Luca Zancato, Arjun Seshadri, Yonatan Dukler, Aditya~Sharad Golatkar, Yantao Shen, Benjamin Bowman, Matthew Trager, Alessandro Achille, and Stefano Soatto.
\newblock B'mojo: Hybrid state space realizations of foundation models with eidetic and fading memory.
\newblock {\em Advances in Neural Information Processing Systems}, 37:130433--130462, 2024.

\bibitem{huang2024compression}
Yuzhen Huang, Jinghan Zhang, Zifei Shan, and Junxian He.
\newblock Compression represents intelligence linearly.
\newblock {\em arXiv preprint arXiv:2404.09937}, 2024.

\bibitem{lecun2006tutorial}
Yann LeCun, Sumit Chopra, Raia Hadsell, M~Ranzato, Fujie Huang, et~al.
\newblock A tutorial on energy-based learning.
\newblock {\em Predicting structured data}, 1(0), 2006.

\bibitem{mceliece1987capacity}
ROBERTJ McEliece, Edwardc Posner, EUGENER Rodemich, and SANTOSHS Venkatesh.
\newblock The capacity of the hopfield associative memory.
\newblock {\em IEEE transactions on Information Theory}, 33(4):461--482, 1987.

\bibitem{farhat1985optical}
Nabil~H Farhat, Demetri Psaltis, Aluizio Prata, and Eung Paek.
\newblock Optical implementation of the hopfield model.
\newblock {\em Applied optics}, 24(10):1469--1475, 1985.

\bibitem{wang2020linformer}
Sinong Wang, Belinda~Z Li, Madian Khabsa, Han Fang, and Hao Ma.
\newblock Linformer: Self-attention with linear complexity.
\newblock {\em arXiv preprint arXiv:2006.04768}, 2020.

\bibitem{gershman2025key}
Samuel~J Gershman, Ila Fiete, and Kazuki Irie.
\newblock Key-value memory in the brain.
\newblock {\em arXiv preprint arXiv:2501.02950}, 2025.

\bibitem{bruni1974bilinear}
Carlo Bruni, Gianni DiPillo, and Giorgio Koch.
\newblock Bilinear systems: An appealing class of" nearly linear" systems in theory and applications.
\newblock {\em IEEE Transactions on automatic control}, 19(4):334--348, 1974.

\bibitem{zhao2016gramian}
Yingbo Zhao and Jorge Cort{\'e}s.
\newblock Gramian-based reachability metrics for bilinear networks.
\newblock {\em IEEE Transactions on Control of Network Systems}, 4(3):620--631, 2016.

\bibitem{wang2023expectation}
Xinyue Wang, Junxia Ma, and Weili Xiong.
\newblock Expectation-maximization algorithm for bilinear state-space models with time-varying delays under non-gaussian noise.
\newblock {\em International Journal of Adaptive Control and Signal Processing}, 37(10):2706--2724, 2023.

\bibitem{pardalos2010optimization}
Panos~M Pardalos and Vitaliy~A Yatsenko.
\newblock {\em Optimization and control of bilinear systems: theory, algorithms, and applications}, volume~11.
\newblock Springer Science \& Business Media, 2010.

\bibitem{hochreiter1997long}
Sepp Hochreiter and J{\"u}rgen Schmidhuber.
\newblock Long short-term memory.
\newblock {\em Neural computation}, 9(8):1735--1780, 1997.

\bibitem{cho2014learning}
Kyunghyun Cho, Bart Van~Merri{\"e}nboer, Caglar Gulcehre, Dzmitry Bahdanau, Fethi Bougares, Holger Schwenk, and Yoshua Bengio.
\newblock Learning phrase representations using rnn encoder-decoder for statistical machine translation.
\newblock {\em arXiv preprint arXiv:1406.1078}, 2014.

\bibitem{merrill2024illusion}
William Merrill, Jackson Petty, and Ashish Sabharwal.
\newblock The illusion of state in state-space models.
\newblock {\em arXiv preprint arXiv:2404.08819}, 2024.

\bibitem{mao2022fine}
Huanru~Henry Mao.
\newblock Fine-tuning pre-trained transformers into decaying fast weights.
\newblock {\em arXiv preprint arXiv:2210.04243}, 2022.

\bibitem{chen2024dijiang}
Hanting Chen, Zhicheng Liu, Xutao Wang, Yuchuan Tian, and Yunhe Wang.
\newblock Dijiang: Efficient large language models through compact kernelization.
\newblock {\em arXiv preprint arXiv:2403.19928}, 2024.

\bibitem{hinton2015distilling}
Geoffrey Hinton, Oriol Vinyals, and Jeff Dean.
\newblock Distilling the knowledge in a neural network.
\newblock {\em arXiv preprint arXiv:1503.02531}, 2015.

\bibitem{bick2025llamba}
Aviv Bick, Tobias Katsch, Nimit Sohoni, Arjun Desai, and Albert Gu.
\newblock Llamba: Scaling distilled recurrent models for efficient language processing.
\newblock {\em arXiv preprint arXiv:2502.14458}, 2025.

\bibitem{bick2024transformers}
Aviv Bick, Kevin Li, Eric Xing, J~Zico Kolter, and Albert Gu.
\newblock Transformers to ssms: Distilling quadratic knowledge to subquadratic models.
\newblock {\em Advances in Neural Information Processing Systems}, 37:31788--31812, 2024.

\bibitem{zhang2025lighttransfer}
Xuan Zhang, Fengzhuo Zhang, Cunxiao Du, Chao Du, Tianyu Pang, Wei Gao, and Min Lin.
\newblock Lighttransfer: Your long-context llm is secretly a hybrid model with effortless adaptation.
\newblock In {\em Workshop on Reasoning and Planning for Large Language Models}, 2025.

\bibitem{van2025lizard}
Chien Van~Nguyen, Ruiyi Zhang, Hanieh Deilamsalehy, Puneet Mathur, Viet~Dac Lai, Haoliang Wang, Jayakumar Subramanian, Ryan~A Rossi, Trung Bui, Nikos Vlassis, et~al.
\newblock Lizard: An efficient linearization framework for large language models.
\newblock {\em arXiv preprint arXiv:2507.09025}, 2025.

\bibitem{ro2025onthefly}
Yeonju Ro, Zhenyu Zhang, Souvik Kundu, Zhangyang Wang, and Aditya Akella.
\newblock On-the-fly adaptive distillation of transformer to dual-state linear attention, 2025.

\bibitem{li2025system}
Zhong-Zhi Li, Duzhen Zhang, Ming-Liang Zhang, Jiaxin Zhang, Zengyan Liu, Yuxuan Yao, Haotian Xu, Junhao Zheng, Pei-Jie Wang, Xiuyi Chen, et~al.
\newblock From system 1 to system 2: A survey of reasoning large language models.
\newblock {\em arXiv preprint arXiv:2502.17419}, 2025.

\bibitem{chen2025towards}
Qiguang Chen, Libo Qin, Jinhao Liu, Dengyun Peng, Jiannan Guan, Peng Wang, Mengkang Hu, Yuhang Zhou, Te~Gao, and Wanxiang Che.
\newblock Towards reasoning era: A survey of long chain-of-thought for reasoning large language models.
\newblock {\em arXiv preprint arXiv:2503.09567}, 2025.

\bibitem{paliotta2025thinking}
Daniele Paliotta, Junxiong Wang, Matteo Pagliardini, Kevin~Y Li, Aviv Bick, J~Zico Kolter, Albert Gu, Fran{\c{c}}ois Fleuret, and Tri Dao.
\newblock Thinking slow, fast: Scaling inference compute with distilled reasoners.
\newblock {\em arXiv preprint arXiv:2502.20339}, 2025.

\bibitem{wang2025m1}
Junxiong Wang, Wen-Ding Li, Daniele Paliotta, Daniel Ritter, Alexander~M Rush, and Tri Dao.
\newblock M1: Towards scalable test-time compute with mamba reasoning models.
\newblock {\em arXiv preprint arXiv:2504.10449}, 2025.

\bibitem{yang2024fla}
Songlin Yang and Yu~Zhang.
\newblock Fla: A triton-based library for hardware-efficient implementations of linear attention mechanism, 2024.

\bibitem{nawrot2025sparse}
Piotr Nawrot, Robert Li, Renjie Huang, Sebastian Ruder, Kelly Marchisio, and Edoardo~M Ponti.
\newblock The sparse frontier: Sparse attention trade-offs in transformer llms.
\newblock {\em arXiv preprint arXiv:2504.17768}, 2025.

\bibitem{gupta2020gmat}
Ankit Gupta and Jonathan Berant.
\newblock Gmat: Global memory augmentation for transformers.
\newblock {\em arXiv preprint arXiv:2006.03274}, 2020.

\bibitem{opensora}
Zangwei Zheng, Xiangyu Peng, Tianji Yang, Chenhui Shen, Shenggui Li, Hongxin Liu, Yukun Zhou, Tianyi Li, and Yang You.
\newblock Open-sora: Democratizing efficient video production for all, March 2024.

\bibitem{dai2019transformer}
Zihang Dai, Zhilin Yang, Yiming Yang, Jaime Carbonell, Quoc~V Le, and Ruslan Salakhutdinov.
\newblock Transformer-xl: Attentive language models beyond a fixed-length context.
\newblock {\em arXiv preprint arXiv:1901.02860}, 2019.

\bibitem{rae2019compressive}
Jack~W Rae, Anna Potapenko, Siddhant~M Jayakumar, and Timothy~P Lillicrap.
\newblock Compressive transformers for long-range sequence modelling.
\newblock {\em arXiv preprint arXiv:1911.05507}, 2019.

\bibitem{sukhbaatar2019adaptive}
Sainbayar Sukhbaatar, Edouard Grave, Piotr Bojanowski, and Armand Joulin.
\newblock Adaptive attention span in transformers.
\newblock {\em arXiv preprint arXiv:1905.07799}, 2019.

\bibitem{ainslie2023colt5}
Joshua Ainslie, Tao Lei, Michiel de~Jong, Santiago Onta{\~n}{\'o}n, Siddhartha Brahma, Yury Zemlyanskiy, David Uthus, Mandy Guo, James Lee-Thorp, Yi~Tay, et~al.
\newblock Colt5: Faster long-range transformers with conditional computation.
\newblock {\em arXiv preprint arXiv:2303.09752}, 2023.

\bibitem{chen2023longlora}
Yukang Chen, Shengju Qian, Haotian Tang, Xin Lai, Zhijian Liu, Song Han, and Jiaya Jia.
\newblock Longlora: Efficient fine-tuning of long-context large language models.
\newblock {\em arXiv preprint arXiv:2309.12307}, 2023.

\bibitem{gao2025seerattention}
Yizhao Gao, Shuming Guo, Shijie Cao, Yuqing Xia, Yu~Cheng, Lei Wang, Lingxiao Ma, Yutao Sun, Tianzhu Ye, Li~Dong, et~al.
\newblock Seerattention-r: Sparse attention adaptation for long reasoning.
\newblock {\em arXiv preprint arXiv:2506.08889}, 2025.

\bibitem{oren2024tova}
Matanel Oren, Michael Hassid, Yossi Adi, and Roy Schwartz.
\newblock Transformers are multi-state rnns.
\newblock {\em arXiv preprint arXiv:2401.06104}, 2024.

\bibitem{liu2024retrievalattention}
Di~Liu, Meng Chen, Baotong Lu, Huiqiang Jiang, Zhenhua Han, Qianxi Zhang, Qi~Chen, Chengruidong Zhang, Bailu Ding, Kai Zhang, et~al.
\newblock Retrievalattention: Accelerating long-context llm inference via vector retrieval.
\newblock {\em arXiv preprint arXiv:2409.10516}, 2024.

\bibitem{xiao2024duoattention}
Guangxuan Xiao, Jiaming Tang, Jingwei Zuo, Junxian Guo, Shang Yang, Haotian Tang, Yao Fu, and Song Han.
\newblock Duoattention: Efficient long-context llm inference with retrieval and streaming heads.
\newblock {\em arXiv preprint arXiv:2410.10819}, 2024.

\bibitem{sun2024shadowkv}
Hanshi Sun, Li-Wen Chang, Wenlei Bao, Size Zheng, Ningxin Zheng, Xin Liu, Harry Dong, Yuejie Chi, and Beidi Chen.
\newblock Shadowkv: Kv cache in shadows for high-throughput long-context llm inference.
\newblock {\em arXiv preprint arXiv:2410.21465}, 2024.

\bibitem{zhang2025pqcache}
Hailin Zhang, Xiaodong Ji, Yilin Chen, Fangcheng Fu, Xupeng Miao, Xiaonan Nie, Weipeng Chen, and Bin Cui.
\newblock Pqcache: Product quantization-based kvcache for long context llm inference.
\newblock {\em Proceedings of the ACM on Management of Data}, 3(3):1--30, 2025.

\bibitem{csordas2024switchhead}
R{\'o}bert Csord{\'a}s, Piotr Pi{\k{e}}kos, Kazuki Irie, and J{\"u}rgen Schmidhuber.
\newblock Switchhead: Accelerating transformers with mixture-of-experts attention.
\newblock {\em Advances in Neural Information Processing Systems}, 37:74411--74438, 2024.

\bibitem{su2024roformer}
Jianlin Su, Murtadha Ahmed, Yu~Lu, Shengfeng Pan, Wen Bo, and Yunfeng Liu.
\newblock Roformer: Enhanced transformer with rotary position embedding.
\newblock {\em Neurocomputing}, 568:127063, 2024.

\bibitem{shazeer2017outrageously}
Noam Shazeer, Azalia Mirhoseini, Krzysztof Maziarz, Andy Davis, Quoc Le, Geoffrey Hinton, and Jeff Dean.
\newblock Outrageously large neural networks: The sparsely-gated mixture-of-experts layer.
\newblock {\em arXiv preprint arXiv:1701.06538}, 2017.

\bibitem{lepikhin2020gshard}
Dmitry Lepikhin, HyoukJoong Lee, Yuanzhong Xu, Dehao Chen, Orhan Firat, Yanping Huang, Maxim Krikun, Noam Shazeer, and Zhifeng Chen.
\newblock Gshard: Scaling giant models with conditional computation and automatic sharding.
\newblock {\em arXiv preprint arXiv:2006.16668}, 2020.

\bibitem{fedus2022switch}
William Fedus, Barret Zoph, and Noam Shazeer.
\newblock Switch transformers: Scaling to trillion parameter models with simple and efficient sparsity.
\newblock {\em Journal of Machine Learning Research}, 23(120):1--39, 2022.

\bibitem{Chen2024MoERBenchTB}
Guanjie Chen, Xinyu Zhao, Tianlong Chen, and Yu~Cheng.
\newblock Moe-rbench: Towards building reliable language models with sparse mixture-of-experts.
\newblock {\em ArXiv}, abs/2406.11353, 2024.

\bibitem{li2024your}
Ziyue Li and Tianyi Zhou.
\newblock Your mixture-of-experts llm is secretly an embedding model for free.
\newblock {\em arXiv preprint arXiv:2410.10814}, 2024.

\bibitem{zhu2024dynamic}
Tong Zhu, Daize Dong, Xiaoye Qu, Jiacheng Ruan, Wenliang Chen, and Yu~Cheng.
\newblock Dynamic data mixing maximizes instruction tuning for mixture-of-experts.
\newblock {\em arXiv preprint arXiv:2406.11256}, 2024.

\bibitem{fedus2022review}
William Fedus, Jeff Dean, and Barret Zoph.
\newblock A review of sparse expert models in deep learning.
\newblock {\em arXiv preprint arXiv:2209.01667}, 2022.

\bibitem{pmlr-v162-du22c}
Nan Du, Yanping Huang, Andrew~M Dai, Simon Tong, Dmitry Lepikhin, Yuanzhong Xu, Maxim Krikun, Yanqi Zhou, Adams~Wei Yu, Orhan Firat, Barret Zoph, Liam Fedus, Maarten~P Bosma, Zongwei Zhou, Tao Wang, Emma Wang, Kellie Webster, Marie Pellat, Kevin Robinson, Kathleen Meier-Hellstern, Toju Duke, Lucas Dixon, Kun Zhang, Quoc Le, Yonghui Wu, Zhifeng Chen, and Claire Cui.
\newblock {GL}a{M}: Efficient scaling of language models with mixture-of-experts.
\newblock In Kamalika Chaudhuri, Stefanie Jegelka, Le~Song, Csaba Szepesvari, Gang Niu, and Sivan Sabato, editors, {\em Proceedings of the 39th International Conference on Machine Learning}, volume 162 of {\em Proceedings of Machine Learning Research}, pages 5547--5569. PMLR, 17--23 Jul 2022.

\bibitem{pmlr-v162-clark22a}
Aidan Clark, Diego De~Las~Casas, Aurelia Guy, Arthur Mensch, Michela Paganini, Jordan Hoffmann, Bogdan Damoc, Blake Hechtman, Trevor Cai, Sebastian Borgeaud, George~Bm Van Den~Driessche, Eliza Rutherford, Tom Hennigan, Matthew~J Johnson, Albin Cassirer, Chris Jones, Elena Buchatskaya, David Budden, Laurent Sifre, Simon Osindero, Oriol Vinyals, Marc'Aurelio Ranzato, Jack Rae, Erich Elsen, Koray Kavukcuoglu, and Karen Simonyan.
\newblock Unified scaling laws for routed language models.
\newblock In Kamalika Chaudhuri, Stefanie Jegelka, Le~Song, Csaba Szepesvari, Gang Niu, and Sivan Sabato, editors, {\em Proceedings of the 39th International Conference on Machine Learning}, volume 162 of {\em Proceedings of Machine Learning Research}, pages 4057--4086. PMLR, 17--23 Jul 2022.

\bibitem{cai2025survey}
Weilin Cai, Juyong Jiang, Fan Wang, Jing Tang, Sunghun Kim, and Jiayi Huang.
\newblock A survey on mixture of experts in large language models.
\newblock {\em IEEE Transactions on Knowledge and Data Engineering}, 2025.

\bibitem{mu2025comprehensive}
Siyuan Mu and Sen Lin.
\newblock A comprehensive survey of mixture-of-experts: Algorithms, theory, and applications.
\newblock {\em arXiv preprint arXiv:2503.07137}, 2025.

\bibitem{gupta2024dbrx}
Nikhil Gupta and Jason Yip.
\newblock Dbrx: Creating an llm from scratch using databricks.
\newblock In {\em Databricks Data Intelligence Platform: Unlocking the GenAI Revolution}, pages 311--330. Springer, 2024.

\bibitem{xue2024openmoe}
Fuzhao Xue, Zian Zheng, Yao Fu, Jinjie Ni, Zangwei Zheng, Wangchunshu Zhou, and Yang You.
\newblock Openmoe: An early effort on open mixture-of-experts language models.
\newblock {\em arXiv preprint arXiv:2402.01739}, 2024.

\bibitem{wei2024skywork}
Tianwen Wei, Bo~Zhu, Liang Zhao, Cheng Cheng, Biye Li, Weiwei L{\"u}, Peng Cheng, Jianhao Zhang, Xiaoyu Zhang, Liang Zeng, et~al.
\newblock Skywork-moe: A deep dive into training techniques for mixture-of-experts language models.
\newblock {\em arXiv preprint arXiv:2406.06563}, 2024.

\bibitem{jiang2024mixtral}
Albert~Q Jiang, Alexandre Sablayrolles, Antoine Roux, Arthur Mensch, Blanche Savary, Chris Bamford, Devendra~Singh Chaplot, Diego de~las Casas, Emma~Bou Hanna, Florian Bressand, et~al.
\newblock Mixtral of experts.
\newblock {\em arXiv preprint arXiv:2401.04088}, 2024.

\bibitem{lu2024twin}
Zhenyi Lu, Chenghao Fan, Wei Wei, Xiaoye Qu, Dangyang Chen, and Yu~Cheng.
\newblock Twin-merging: Dynamic integration of modular expertise in model merging.
\newblock {\em Advances in Neural Information Processing Systems}, 37:78905--78935, 2024.

\bibitem{liu2025muon}
Jingyuan Liu, Jianlin Su, Xingcheng Yao, Zhejun Jiang, Guokun Lai, Yulun Du, Yidao Qin, Weixin Xu, Enzhe Lu, Junjie Yan, et~al.
\newblock Muon is scalable for llm training.
\newblock {\em arXiv preprint arXiv:2502.16982}, 2025.

\bibitem{NEURIPS2022_df4f371f}
Zewen Chi, Li~Dong, Shaohan Huang, Damai Dai, Shuming Ma, Barun Patra, Saksham Singhal, Payal Bajaj, XIA SONG, Xian-Ling Mao, Heyan Huang, and Furu Wei.
\newblock On the representation collapse of sparse mixture of experts.
\newblock In S.~Koyejo, S.~Mohamed, A.~Agarwal, D.~Belgrave, K.~Cho, and A.~Oh, editors, {\em Advances in Neural Information Processing Systems}, volume~35, pages 34600--34613. Curran Associates, Inc., 2022.

\bibitem{nguyen2024sigmoid}
Huy Nguyen, Nhat Ho, and Alessandro Rinaldo.
\newblock Sigmoid gating is more sample efficient than softmax gating in mixture of experts.
\newblock {\em arXiv preprint arXiv:2405.13997}, 2024.

\bibitem{NEURIPS2021_48237d9f}
Carlos Riquelme, Joan Puigcerver, Basil Mustafa, Maxim Neumann, Rodolphe Jenatton, Andr\'{e} Susano~Pinto, Daniel Keysers, and Neil Houlsby.
\newblock Scaling vision with sparse mixture of experts.
\newblock In M.~Ranzato, A.~Beygelzimer, Y.~Dauphin, P.S. Liang, and J.~Wortman Vaughan, editors, {\em Advances in Neural Information Processing Systems}, volume~34, pages 8583--8595. Curran Associates, Inc., 2021.

\bibitem{Dosovitskiy2020AnII}
Alexey Dosovitskiy, Lucas Beyer, Alexander Kolesnikov, Dirk Weissenborn, Xiaohua Zhai, Thomas Unterthiner, Mostafa Dehghani, Matthias Minderer, Georg Heigold, Sylvain Gelly, Jakob Uszkoreit, and Neil Houlsby.
\newblock An image is worth 16x16 words: Transformers for image recognition at scale.
\newblock {\em ArXiv}, abs/2010.11929, 2020.

\bibitem{wang2024remoe}
Ziteng Wang, Jun Zhu, and Jianfei Chen.
\newblock Remoe: Fully differentiable mixture-of-experts with relu routing.
\newblock {\em arXiv preprint arXiv:2412.14711}, 2024.

\bibitem{song2025blockffn}
Chenyang Song, Weilin Zhao, Xu~Han, Chaojun Xiao, Yingfa Chen, Yuxuan Li, Zhiyuan Liu, and Maosong Sun.
\newblock Blockffn: Towards end-side acceleration-friendly mixture-of-experts with chunk-level activation sparsity.
\newblock {\em arXiv preprint arXiv:2507.08771}, 2025.

\bibitem{Jin2024MoEAM}
Peng Jin, Bo~Zhu, Li~Yuan, and Shuicheng Yan.
\newblock Moe++: Accelerating mixture-of-experts methods with zero-computation experts.
\newblock {\em ArXiv}, abs/2410.07348, 2024.

\bibitem{Zoph2022STMoEDS}
Barret Zoph, Irwan Bello, Sameer Kumar, Nan Du, Yanping Huang, Jeff Dean, Noam~M. Shazeer, and William Fedus.
\newblock St-moe: Designing stable and transferable sparse expert models, 2022.

\bibitem{wang2024let}
Zihan Wang, Deli Chen, Damai Dai, Runxin Xu, Zhuoshu Li, and Yu~Wu.
\newblock Let the expert stick to his last: Expert-specialized fine-tuning for sparse architectural large language models.
\newblock {\em arXiv preprint arXiv:2407.01906}, 2024.

\bibitem{Krajewski2024ScalingLF}
Jakub Krajewski, Jan Ludziejewski, Kamil Adamczewski, Maciej Pi'oro, Michal Krutul, Szymon Antoniak, Kamil Ciebiera, Krystian Kr'ol, Tomasz Odrzyg'o'zd'z, Piotr Sankowski, Marek Cygan, and Sebastian Jaszczur.
\newblock Scaling laws for fine-grained mixture of experts.
\newblock {\em ArXiv}, abs/2402.07871, 2024.

\bibitem{Yang2024Qwen2TR}
An~Yang, Baosong Yang, Binyuan Hui, Bo~Zheng, Bowen Yu, Chang Zhou, Chengpeng Li, Chengyuan Li, Dayiheng Liu, Fei Huang, Guanting Dong, Haoran Wei, Huan Lin, Jialong Tang, Jialin Wang, Jian Yang, Jianhong Tu, Jianwei Zhang, Jianxin Ma, Jin Xu, Jingren Zhou, Jinze Bai, Jinzheng He, Junyang Lin, Kai Dang, Keming Lu, Ke-Yang Chen, Kexin Yang, Mei Li, Min Xue, Na~Ni, Pei Zhang, Peng Wang, Ru~Peng, Rui Men, Ruize Gao, Runji Lin, Shijie Wang, Shuai Bai, Sinan Tan, Tianhang Zhu, Tianhao Li, Tianyu Liu, Wenbin Ge, Xiaodong Deng, Xiaohuan Zhou, Xingzhang Ren, Xinyu Zhang, Xipin Wei, Xuancheng Ren, Yang Fan, Yang Yao, Yichang Zhang, Yunyang Wan, Yunfei Chu, Zeyu Cui, Zhenru Zhang, and Zhi-Wei Fan.
\newblock Qwen2 technical report.
\newblock {\em ArXiv}, abs/2407.10671, 2024.

\bibitem{Elhoushi2024LayerSkipEE}
Mostafa Elhoushi, Akshat Shrivastava, Diana Liskovich, Basil Hosmer, Bram Wasti, Liangzhen Lai, Anas Mahmoud, Bilge Acun, Saurabh Agarwal, Ahmed Roman, Ahmed Aly, Beidi Chen, and Carole-Jean Wu.
\newblock Layerskip: Enabling early exit inference and self-speculative decoding.
\newblock {\em ArXiv}, abs/2404.16710, 2024.

\bibitem{Puigcerver2023FromST}
Joan Puigcerver, Carlos Riquelme, Basil Mustafa, and Neil Houlsby.
\newblock From sparse to soft mixtures of experts.
\newblock {\em ArXiv}, abs/2308.00951, 2023.

\bibitem{Shen2023ModuleFormerLM}
Yikang Shen, Zheyu Zhang, Tianyou Cao, Shawn Tan, Zhenfang Chen, and Chuang Gan.
\newblock Moduleformer: Learning modular large language models from uncurated data.
\newblock {\em ArXiv}, abs/2306.04640, 2023.

\bibitem{Dou2024LoRAMoEAW}
Shihan Dou, Enyu Zhou, Yan Liu, Songyang Gao, Wei Shen, Limao Xiong, Yuhao Zhou, Xiao Wang, Zhiheng Xi, Xiaoran Fan, Shiliang Pu, Jiang Zhu, Rui Zheng, Tao Gui, Qi~Zhang, and Xuanjing Huang.
\newblock Loramoe: Alleviating world knowledge forgetting in large language models via moe-style plugin.
\newblock In {\em Annual Meeting of the Association for Computational Linguistics}, 2024.

\bibitem{Luo2024MoELoRACL}
Tongxu Luo, Jiahe Lei, Fangyu Lei, Weihao Liu, Shizhu He, Jun Zhao, and Kang Liu.
\newblock Moelora: Contrastive learning guided mixture of experts on parameter-efficient fine-tuning for large language models.
\newblock {\em ArXiv}, abs/2402.12851, 2024.

\bibitem{Wu2024MixtureOL}
Xun Wu, Shaohan Huang, and Furu Wei.
\newblock Mixture of lora experts.
\newblock {\em ArXiv}, abs/2404.13628, 2024.

\bibitem{fan2025make}
Chenghao Fan, Zhenyi Lu, Sichen Liu, Chengfeng Gu, Xiaoye Qu, Wei Wei, and Yu~Cheng.
\newblock Make lora great again: Boosting lora with adaptive singular values and mixture-of-experts optimization alignment.
\newblock {\em arXiv preprint arXiv:2502.16894}, 2025.

\bibitem{He2024MixtureOA}
Xu~Owen He.
\newblock Mixture of a million experts.
\newblock {\em ArXiv}, abs/2407.04153, 2024.

\bibitem{zhang2024clip}
Jihai Zhang, Xiaoye Qu, Tong Zhu, and Yu~Cheng.
\newblock Clip-moe: Towards building mixture of experts for clip with diversified multiplet upcycling.
\newblock {\em arXiv preprint arXiv:2409.19291}, 2024.

\bibitem{Molchanov2019ImportanceEF}
Pavlo Molchanov, Arun Mallya, Stephen Tyree, Iuri Frosio, and Jan Kautz.
\newblock Importance estimation for neural network pruning.
\newblock {\em 2019 IEEE/CVF Conference on Computer Vision and Pattern Recognition (CVPR)}, pages 11256--11264, 2019.

\bibitem{Touvron2023Llama2O}
Hugo Touvron, Louis Martin, Kevin~R. Stone, Peter Albert, Amjad Almahairi, Yasmine Babaei, Niko lay Bashlykov, Soumya Batra, Prajjwal Bhargava, Shruti Bhosale, Daniel~M. Bikel, Lukas Blecher, Cris tian Cant{\'o}n~Ferrer, Moya Chen, Guillem Cucurull, David Esiobu, Jude Fernandes, Jeremy Fu, Wenyin Fu, Brian Fuller, Cynthia Gao, Vedanuj Goswami, Naman Goyal, Anthony~S. Hartshorn, Saghar Hosseini, Rui Hou, Hakan Inan, Marcin Kardas, Viktor Kerkez, Madian Khabsa, Isabel~M. Kloumann, Artem Korenev, Punit~Singh Koura, Marie-Anne Lachaux, Thibaut Lavril, Jenya Lee, Diana Liskovich, Yinghai Lu, Yuning Mao, Xavier Martinet, Todor Mihaylov, Pushkar Mishra, Igor Molybog, Yixin Nie, Andrew Poulton, Jeremy Reizenstein, Rashi Rungta, Kalyan Saladi, Alan Schelten, Ruan Silva, Eric~Michael Smith, R.~Subramanian, Xia Tan, Binh Tang, Ross Taylor, Adina Williams, Jian~Xiang Kuan, Puxin Xu, Zhengxu Yan, Iliyan Zarov, Yuchen Zhang, Angela Fan, Melissa Hall~Melanie Kambadur, Sharan Narang, Aur'elien Rodriguez, Robert Stojnic,
  Sergey Edunov, and Thomas Scialom.
\newblock Llama 2: Open foundation and fine-tuned chat models.
\newblock {\em ArXiv}, abs/2307.09288, 2023.

\bibitem{Dubey2024TheL3}
Abhimanyu Dubey, Abhinav Jauhri, Abhinav Pandey, Abhishek Kadian, Ahmad Al-Dahle, Aiesha Letman, Akhil Mathur, Alan Schelten, Amy Yang, Angela Fan, Anirudh Goyal, Anthony~S. Hartshorn, Aobo Yang, Archi Mitra, Archie Sravankumar, Artem Korenev, Arthur Hinsvark, Arun Rao, Aston Zhang, Aur'elien Rodriguez, Austen Gregerson, Ava Spataru, Baptiste Rozi{\`e}re, Bethany Biron, Binh Tang, Bobbie Chern, Charlotte Caucheteux, Chaya Nayak, Chloe Bi, Chris Marra, Chris McConnell, Christian Keller, Christophe Touret, Chunyang Wu, Corinne Wong, Cris tian Cant{\'o}n~Ferrer, Cyrus Nikolaidis, Damien Allonsius, Daniel Song, Danielle Pintz, Danny Livshits, David Esiobu, Dhruv Choudhary, Dhruv Mahajan, Diego Garcia-Olano, Diego Perino, Dieuwke Hupkes, Egor Lakomkin, Ehab~A. AlBadawy, Elina Lobanova, Emily Dinan, Eric~Michael Smith, Filip Radenovic, Frank Zhang, Gabriele Synnaeve, Gabrielle Lee, Georgia~Lewis Anderson, Graeme Nail, Gr{\'e}goire Mialon, Guanglong Pang, Guillem Cucurell, Hailey Nguyen, Hannah Korevaar, Hu~Xu, Hugo
  Touvron, Iliyan Zarov, Imanol~Arrieta Ibarra, Isabel~M. Kloumann, Ishan Misra, Ivan Evtimov, Jade Copet, Jaewon Lee, Jan Geffert, Jana Vranes, Jason Park, Jay Mahadeokar, Jeet Shah, Jelmer van~der Linde, Jennifer Billock, Jenny Hong, Jenya Lee, Jeremy Fu, Jianfeng Chi, Jianyu Huang, Jiawen Liu, Jie Wang, Jiecao Yu, Joanna Bitton, Joe Spisak, Jongsoo Park, Joseph Rocca, Joshua Johnstun, Joshua Saxe, Ju-Qing Jia, Kalyan~Vasuden Alwala, K.~Upasani, Kate Plawiak, Keqian Li, Ken-591 neth Heafield, Kevin~R. Stone, Khalid El-Arini, Krithika Iyer, Kshitiz Malik, Kuen ley Chiu, Kunal Bhalla, Lauren Rantala-Yeary, Laurens van~der Maaten, Lawrence Chen, Liang Tan, Liz Jenkins, Louis Martin, Lovish Madaan, Lubo Malo, Lukas Blecher, Lukas Landzaat, Luke de~Oliveira, Madeline Muzzi, Mahesh Pasupuleti, Mannat Singh, Manohar Paluri, Marcin Kardas, Mathew Oldham, Mathieu Rita, Maya Pavlova, Melissa Hall~Melanie Kambadur, Mike Lewis, Min Si, Mitesh~Kumar Singh, Mona Hassan, Naman Goyal, Narjes Torabi, Niko lay Bashlykov,
  Nikolay Bogoychev, Niladri~S. Chatterji, Olivier Duchenne, Onur cCelebi, Patrick Alrassy, Pengchuan Zhang, Pengwei Li, Petar Vasi{\'c}, Peter Weng, Prajjwal Bhargava, Pratik Dubal, Praveen Krishnan, Punit~Singh Koura, Puxin Xu, Qing He, Qingxiao Dong, Ragavan Srinivasan, Raj Ganapathy, Ramon Calderer, Ricardo~Silveira Cabral, Robert Stojnic, Roberta Raileanu, Rohit Girdhar, Rohit Patel, Ro~main Sauvestre, Ron nie Polidoro, Roshan Sumbaly, Ross Taylor, Ruan Silva, Rui Hou, Rui Wang, Saghar Hosseini, Sa~hana Chennabasappa, Sanjay Singh, Sean Bell, Seohyun~Sonia Kim, Sergey Edunov, Shaoliang Nie, Sharan Narang, Sharath~Chandra Raparthy, Sheng Shen, Shengye Wan, Shruti Bhosale, Shun Zhang, Simon Vandenhende, Soumya Batra, Spencer Whitman, Sten Sootla, Stephane Collot, Suchin Gururangan, Sydney Borodinsky, Tamar Herman, Tara Fowler, Tarek Sheasha, Thomas Georgiou, Thomas Scialom, Tobias Speckbacher, Todor Mihaylov, Tong Xiao, Ujjwal Karn, Vedanuj Goswami, Vibhor Gupta, Vignesh Ramanathan, Viktor Kerkez, Vincent
  Gonguet, Vir ginie Do, Vish Vogeti, Vladan Petrovic, Weiwei Chu, Wenhan Xiong, Wenyin Fu, Whit ney Meers, Xavier Martinet, Xiaodong Wang, Xiaoqing~Ellen Tan, Xinfeng Xie, Xuchao Jia, Xuewei Wang, Yaelle Goldschlag, Yashesh Gaur, Yasmine Babaei, Yiqian Wen, Yiwen Song, Yuchen Zhang, Yue Li, Yuning Mao, Zacharie~Delpierre Coudert, Zhengxu Yan, Zhengxing Chen, Zoe Papakipos, Aaditya~K. Singh, Aaron Grattafiori, Abha Jain, Adam Kelsey, Adam Shajnfeld, Adi Gangidi, Adolfo Victoria, Ahuva Goldstand, Ajay Menon, Ajay Sharma, Alex Boesenberg, Alex Vaughan, Alexei Baevski, Allie Feinstein, Amanda Kallet, Amit Sangani, Anam Yunus, Andrei Lupu, Andres Alvarado, Andrew Caples, Andrew Gu, Andrew Ho, Andrew Poulton, Andrew Ryan, Ankit Ramchandani, Annie Franco, Aparajita Saraf, Arkabandhu Chowdhury, Ashley Gabriel, Ashwin Bharambe, Assaf Eisenman, Azadeh Yazdan, Beau James, Ben Maurer, Benjamin Leonhardi, Po-Yao~(Bernie) Huang, Beth Loyd, Beto de~Paola, Bhargavi Paranjape, Bing Liu, Bo~Wu, Boyu Ni, Braden Hancock, Bram
  Wasti, Brandon Spence, Brani Stojkovic, Brian Gamido, Britt Montalvo, Carl Parker, Carly Burton, Catalina Mejia, Changhan Wang, Changkyu Kim, Chao Zhou, Chester Hu, Ching-Hsiang Chu, Chris Cai, Chris Tindal, Christoph Feichtenhofer, Damon Civin, Dana Beaty, Daniel Kreymer, Shang-Wen Li, Danny Wyatt, David Adkins, David Xu, Davide Testuggine, Delia David, Devi Parikh, Diana Liskovich, Didem Foss, Dingkang Wang, Duc Le, Dustin Holland, Edward Dowling, Eissa Jamil, Elaine Montgomery, Eleonora Presani, Emily Hahn, Emily Wood, Erik Brinkman, Esteban Arcaute, Evan Dunbar, Evan Smothers, Fei Sun, Felix Kreuk, Feng Tian, Firat Ozgenel, Francesco Caggioni, Francisco Guzm’an, Frank~J. Kanayet, Frank Seide, Gabriela~Medina Florez, Gabriella Schwarz, Gada Badeer, Georgia Swee, Gil Halpern, Govind Thattai, Grant Herman, Grigory~G. Sizov, Guangyi Zhang, Guna Lakshminarayanan, Hamid Shojanazeri, Han Zou, Hannah Wang, Han Zha, Haroun Habeeb, Harrison Rudolph, Helen Suk, Henry Aspegren, Hunter Goldman, Igor Molybog, Igor
  Tufanov, Irina-Elena Veliche, Itai Gat, Jake Weissman, James Geboski, James Kohli, Japhet Asher, Jean-Baptiste Gaya, Jeff Marcus, Jeff Tang, Jennifer Chan, Jenny Zhen, Jeremy Reizenstein, Jeremy Teboul, Jessica Zhong, Jian Jin, Jingyi Yang, Joe Cummings, Jon Carvill, Jon Shepard, Jonathan McPhie, Jonathan Torres, Josh Ginsburg, Junjie Wang, Kaixing(Kai) Wu, U~KamHou, Karan Saxena, Karthik Prasad, Kartikay Khandelwal, Katayoun Zand, Kathy Matosich, Kaushik Veeraraghavan, Kelly Michelena, Keqian Li, Kun Huang, Kunal Chawla, Kushal Lakhotia, Kyle Huang, Lailin Chen, Lakshya Garg, A~Lavender, Leandro Silva, Lee Bell, Lei Zhang, Liangpeng Guo, Licheng Yu, Liron Moshkovich, Luca Wehrstedt, Madian Khabsa, Manav Avalani, Manish Bhatt, Maria Tsimpoukelli, Martynas Mankus, Matan Hasson, Matthew Lennie, Matthias Reso, Maxim Groshev, Maxim Naumov, Maya Lathi, Meghan Keneally, Michael~L. Seltzer, Michal Valko, Michelle Restrepo, Mihir Patel, Mik Vyatskov, Mikayel Samvelyan, Mike Clark, Mike Macey, Mike Wang,
  Miquel~Jubert Hermoso, Mo~Metanat, Mohammad Rastegari, Munish Bansal, Nandhini Santhanam, Natascha Parks, Natasha White, Navy ata Bawa, Nayan Singhal, Nick Egebo, Nicolas Usunier, Nikolay~Pavlovich Laptev, Ning Dong, Ning Zhang, Norman Cheng, Oleg Chernoguz, Olivia Hart, Omkar Salpekar, Ozlem Kalinli, Parkin Kent, Parth Parekh, Paul Saab, Pavan Balaji, Pe~dro Rittner, Philip Bontrager, Pierre Roux, Piotr Doll{\'a}r, Polina Zvyagina, Prashant Ratanchandani, Pritish Yuvraj, Qian Liang, Rachad Alao, Rachel Rodriguez, Rafi Ayub, Raghotham Murthy, Raghu Nayani, Rahul Mitra, Raymond Li, Rebekkah Hogan, Robin Battey, Rocky Wang, Rohan Maheswari, Russ Howes, Ruty Rinott, Sai~Jayesh Bondu, Samyak Datta, Sara Chugh, Sara Hunt, Sargun Dhillon, Sasha Sidorov, Satadru Pan, Saurabh Verma, Seiji Yamamoto, Sharadh Ramaswamy, Shaun Lindsay, Sheng Feng, Shenghao Lin, Shengxin~Cindy Zha, Shiva Shankar, Shuqiang Zhang, Sinong Wang, Sneha Agarwal, Soji Sajuyigbe, Soumith Chintala, Stephanie Max, Stephen Chen, Steve Kehoe, Steve
  Satterfield, Sudarshan Govindaprasad, Sumit Gupta, Sung-Bae Cho, Sunny Virk, Suraj Subramanian, Sy~Choudhury, Sydney Goldman, Tal Remez, Tamar Glaser, Tamara Best, Thilo Kohler, Thomas Robinson, Tianhe Li, Tianjun Zhang, Tim Matthews, Timothy Chou, Tzook Shaked, Varun Vontimitta, Victoria Ajayi, Victoria Montanez, Vijai Mohan, Vinay~Satish Kumar, Vishal Mangla, Vlad Ionescu, Vlad~Andrei Poenaru, Vlad~T. Mihailescu, Vladimir Ivanov, Wei Li, Wenchen Wang, Wenwen Jiang, Wes Bouaziz, Will Constable, Xia Tang, Xiaofang Wang, Xiaojian Wu, Xiaolan Wang, Xide Xia, Xilun Wu, Xinbo Gao, Yanjun Chen, Ye~Hu, Ye~Jia, Ye~Qi, Yenda Li, Yilin Zhang, Ying Zhang, Yossi Adi, Youngjin Nam, Yu~Wang, Yuchen Hao, Yundi Qian, Yuzi He, Zach Rait, Zachary DeVito, Zef Rosnbrick, Zhaoduo Wen, Zhenyu Yang, and Zhiwei Zhao.
\newblock The llama 3 herd of models.
\newblock {\em ArXiv}, abs/2407.21783, 2024.

\bibitem{Tan2024DLODL}
Zhen Tan, Daize Dong, Xinyu Zhao, Jie Peng, Yu~Cheng, and Tianlong Chen.
\newblock Dlo: Dynamic layer operation for efficient vertical scaling of llms.
\newblock {\em ArXiv}, abs/2407.11030, 2024.

\bibitem{Zhang2024MoDificationMO}
Chen Zhang, Meizhi Zhong, Qimeng Wang, Xuantao Lu, Zheyu Ye, Chengqiang Lu, Yan Gao, Yao Hu, Kehai Chen, Min Zhang, and Dawei Song.
\newblock Modification: Mixture of depths made easy.
\newblock {\em ArXiv}, abs/2410.14268, 2024.

\bibitem{jiang2023mistral}
Albert~Q Jiang, Alexandre Sablayrolles, Arthur Mensch, Chris Bamford, Devendra~Singh Chaplot, Diego de~las Casas, Florian Bressand, Gianna Lengyel, Guillaume Lample, Lucile Saulnier, et~al.
\newblock Mistral 7b.
\newblock {\em arXiv preprint arXiv:2310.06825}, 2023.

\bibitem{de2024griffin}
Soham De, Samuel~L Smith, Anushan Fernando, Aleksandar Botev, George Cristian-Muraru, Albert Gu, Ruba Haroun, Leonard Berrada, Yutian Chen, Srivatsan Srinivasan, et~al.
\newblock Griffin: Mixing gated linear recurrences with local attention for efficient language models.
\newblock {\em arXiv preprint arXiv:2402.19427}, 2024.

\bibitem{wuneng}
Liu Xiao, Li~Zhiyuan, and Lin Yueyu.
\newblock {WuNeng}: {Hybrid} {State} with {Attention}, April 2025.

\bibitem{yu2025discrete}
Runpeng Yu, Qi~Li, and Xinchao Wang.
\newblock Discrete diffusion in large language and multimodal models: A survey.
\newblock {\em arXiv preprint arXiv:2506.13759}, 2025.

\bibitem{sahoo2024simpleeffectivemaskeddiffusion}
Subham~Sekhar Sahoo, Marianne Arriola, Yair Schiff, Aaron Gokaslan, Edgar Marroquin, Justin~T Chiu, Alexander Rush, and Volodymyr Kuleshov.
\newblock Simple and effective masked diffusion language models, 2024.

\bibitem{devlin2019bert}
Jacob Devlin, Ming-Wei Chang, Kenton Lee, and Kristina Toutanova.
\newblock Bert: Pre-training of deep bidirectional transformers for language understanding.
\newblock In {\em Proceedings of the 2019 conference of the North American chapter of the association for computational linguistics: human language technologies, volume 1 (long and short papers)}, pages 4171--4186, 2019.

\bibitem{zhao2025d1scalingreasoningdiffusion}
Siyan Zhao, Devaansh Gupta, Qinqing Zheng, and Aditya Grover.
\newblock d1: Scaling reasoning in diffusion large language models via reinforcement learning, 2025.

\bibitem{liu2023visual}
Haotian Liu, Chunyuan Li, Qingyang Wu, and Yong~Jae Lee.
\newblock Visual instruction tuning.
\newblock {\em Advances in neural information processing systems}, 36:34892--34916, 2023.

\bibitem{su2025openthinkimg}
Zhaochen Su, Linjie Li, Mingyang Song, Yunzhuo Hao, Zhengyuan Yang, Jun Zhang, Guanjie Chen, Jiawei Gu, Juntao Li, Xiaoye Qu, et~al.
\newblock Openthinkimg: Learning to think with images via visual tool reinforcement learning.
\newblock {\em arXiv preprint arXiv:2505.08617}, 2025.

\bibitem{su2025thinking}
Zhaochen Su, Peng Xia, Hangyu Guo, Zhenhua Liu, Yan Ma, Xiaoye Qu, Jiaqi Liu, Yanshu Li, Kaide Zeng, Zhengyuan Yang, et~al.
\newblock Thinking with images for multimodal reasoning: Foundations, methods, and future frontiers.
\newblock {\em arXiv preprint arXiv:2506.23918}, 2025.

\bibitem{chen2025advancing}
Shuang Chen, Yue Guo, Zhaochen Su, Yafu Li, Yulun Wu, Jiacheng Chen, Jiayu Chen, Weijie Wang, Xiaoye Qu, and Yu~Cheng.
\newblock Advancing multimodal reasoning: From optimized cold start to staged reinforcement learning.
\newblock {\em arXiv preprint arXiv:2506.04207}, 2025.

\bibitem{yu2025dimplediscretediffusionmultimodal}
Runpeng Yu, Xinyin Ma, and Xinchao Wang.
\newblock Dimple: Discrete diffusion multimodal large language model with parallel decoding, 2025.

\bibitem{li2024llavanext-ablations}
Bo~Li, Hao Zhang, Kaichen Zhang, Dong Guo, Yuanhan Zhang, Renrui Zhang, Feng Li, Ziwei Liu, and Chunyuan Li.
\newblock Llava-next: What else influences visual instruction tuning beyond data?, May 2024.

\bibitem{zhang2023robust}
Yihua Zhang, Ruisi Cai, Tianlong Chen, Guanhua Zhang, Huan Zhang, Pin-Yu Chen, Shiyu Chang, Zhangyang Wang, and Sijia Liu.
\newblock Robust mixture-of-expert training for convolutional neural networks.
\newblock In {\em Proceedings of the IEEE/CVF International Conference on Computer Vision}, pages 90--101, 2023.

\bibitem{chowdhury2023patch}
Mohammed Nowaz~Rabbani Chowdhury, Shuai Zhang, Meng Wang, Sijia Liu, and Pin-Yu Chen.
\newblock Patch-level routing in mixture-of-experts is provably sample-efficient for convolutional neural networks.
\newblock In {\em International Conference on Machine Learning}, pages 6074--6114. PMLR, 2023.

\bibitem{chen2024res}
Chi-Sheng Chen, Guan-Ying Chen, Dong Zhou, Di~Jiang, and Dai-Shi Chen.
\newblock Res-vmamba: Fine-grained food category visual classification using selective state space models with deep residual learning.
\newblock {\em arXiv preprint arXiv:2402.15761}, 2024.

\bibitem{fang2024mammil}
Zijie Fang, Yifeng Wang, Ye~Zhang, Zhi Wang, Jian Zhang, Xiangyang Ji, and Yongbing Zhang.
\newblock Mammil: Multiple instance learning for whole slide images with state space models.
\newblock In {\em 2024 IEEE International Conference on Bioinformatics and Biomedicine (BIBM)}, pages 3200--3205. IEEE, 2024.

\bibitem{yao2024spectralmamba}
Jing Yao, Danfeng Hong, Chenyu Li, and Jocelyn Chanussot.
\newblock Spectralmamba: Efficient mamba for hyperspectral image classification.
\newblock {\em arXiv preprint arXiv:2404.08489}, 2024.

\bibitem{wang2024memorymamba}
Qianning Wang, He~Hu, and Yucheng Zhou.
\newblock Memorymamba: Memory-augmented state space model for defect recognition.
\newblock {\em arXiv preprint arXiv:2405.03673}, 2024.

\bibitem{wang2024mambayolo}
Zeyu Wang, Chen Li, Huiying Xu, and Xinzhong Zhu.
\newblock Mamba yolo: Ssms-based yolo for object detection.
\newblock {\em arXiv preprint arXiv:2406.05835}, 2024.

\bibitem{chen2024mim}
Tianxiang Chen, Zi~Ye, Zhentao Tan, Tao Gong, Yue Wu, Qi~Chu, Bin Liu, Nenghai Yu, and Jieping Ye.
\newblock Mim-istd: Mamba-in-mamba for efficient infrared small target detection.
\newblock {\em IEEE Transactions on Geoscience and Remote Sensing}, 2024.

\bibitem{shen2025htd}
Dunbin Shen, Xuanbing Zhu, Jiacheng Tian, Jianjun Liu, Zhenrong Du, Hongyu Wang, and Xiaorui Ma.
\newblock Htd-mamba: Efficient hyperspectral target detection with pyramid state space model.
\newblock {\em IEEE Transactions on Geoscience and Remote Sensing}, 2025.

\bibitem{verma2024soar}
Tushar Verma, Jyotsna Singh, Yash Bhartari, Rishi Jarwal, Suraj Singh, and Shubhkarman Singh.
\newblock Soar: Advancements in small body object detection for aerial imagery using state space models and programmable gradients.
\newblock {\em arXiv preprint arXiv:2405.01699}, 2024.

\bibitem{yuan2024mamba}
Haobo Yuan, Xiangtai Li, Lu~Qi, Tao Zhang, Ming-Hsuan Yang, Shuicheng Yan, and Chen~Change Loy.
\newblock Mamba or rwkv: Exploring high-quality and high-efficiency segment anything model.
\newblock {\em arXiv preprint arXiv:2406.19369}, 2024.

\bibitem{fu2024segman}
Yunxiang Fu, Meng Lou, and Yizhou Yu.
\newblock Segman: Omni-scale context modeling with state space models and local attention for semantic segmentation.
\newblock {\em arXiv preprint arXiv:2412.11890}, 2024.

\bibitem{chen2024vision}
Zhaohui Chen, Elyas~Asadi Shamsabadi, Sheng Jiang, Luming Shen, and Daniel Dias-da Costa.
\newblock Vision mamba-based autonomous crack segmentation on concrete, asphalt, and masonry surfaces.
\newblock {\em arXiv preprint arXiv:2406.16518}, 2024.

\bibitem{wang2024pyramidmamba}
Libo Wang, Dongxu Li, Sijun Dong, Xiaoliang Meng, Xiaokang Zhang, and Danfeng Hong.
\newblock Pyramidmamba: rethinking pyramid feature fusion with selective space state model for semantic segmentation of remote sensing imagery.
\newblock {\em arXiv preprint arXiv:2406.10828}, 2024.

\bibitem{yang2024restore}
Zhiwen Yang, Jiayin Li, Hui Zhang, Dan Zhao, Bingzheng Wei, and Yan Xu.
\newblock Restore-rwkv: Efficient and effective medical image restoration with rwkv.
\newblock {\em arXiv preprint arXiv:2407.11087}, 2024.

\bibitem{lei2024dvmsr}
Xiaoyan Lei, Wenlong Zhang, and Weifeng Cao.
\newblock Dvmsr: Distillated vision mamba for efficient super-resolution.
\newblock In {\em Proceedings of the IEEE/CVF Conference on Computer Vision and Pattern Recognition}, pages 6536--6546, 2024.

\bibitem{chen2025q}
Yujie Chen, Haotong Qin, Zhang Zhang, Michelo Magno, Luca Benini, and Yawei Li.
\newblock Q-mambair: Accurate quantized mamba for efficient image restoration.
\newblock {\em arXiv preprint arXiv:2503.21970}, 2025.

\bibitem{du2024exploring}
Yuzhen Du, Teng Hu, Jiangning Zhang, Ran Yi~Chengming Xu, Xiaobin Hu, Kai Wu, Donghao Luo, Yabiao Wang, and Lizhuang Ma.
\newblock Exploring real\&synthetic dataset and linear attention in image restoration.
\newblock {\em arXiv preprint arXiv:2412.03814}, 2024.

\bibitem{lin2024pixmamba}
Wei-Tung Lin, Yong-Xiang Lin, Jyun-Wei Chen, and Kai-Lung Hua.
\newblock Pixmamba: Leveraging state space models in a dual-level architecture for underwater image enhancement.
\newblock In {\em Proceedings of the Asian Conference on Computer Vision}, pages 3622--3637, 2024.

\bibitem{guan2024watermamba}
Meisheng Guan, Haiyong Xu, Gangyi Jiang, Mei Yu, Yeyao Chen, Ting Luo, and Yang Song.
\newblock Watermamba: Visual state space model for underwater image enhancement.
\newblock {\em arXiv preprint arXiv:2405.08419}, 2024.

\bibitem{bai2024retinexmamba}
Jiesong Bai, Yuhao Yin, Qiyuan He, Yuanxian Li, and Xiaofeng Zhang.
\newblock Retinexmamba: Retinex-based mamba for low-light image enhancement.
\newblock {\em arXiv preprint arXiv:2405.03349}, 2024.

\bibitem{zhang2024llemamba}
Xuanqi Zhang, Haijin Zeng, Jinwang Pan, Qiangqiang Shen, and Yongyong Chen.
\newblock Llemamba: Low-light enhancement via relighting-guided mamba with deep unfolding network.
\newblock {\em arXiv preprint arXiv:2406.01028}, 2024.

\bibitem{li2024fouriermamba}
Dong Li, Yidi Liu, Xueyang Fu, Senyan Xu, and Zheng-Jun Zha.
\newblock Fouriermamba: Fourier learning integration with state space models for image deraining.
\newblock {\em arXiv preprint arXiv:2405.19450}, 2024.

\bibitem{shi2025vmambair}
Yuan Shi, Bin Xia, Xiaoyu Jin, Xing Wang, Tianyu Zhao, Xin Xia, Xuefeng Xiao, and Wenming Yang.
\newblock Vmambair: Visual state space model for image restoration.
\newblock {\em IEEE Transactions on Circuits and Systems for Video Technology}, 2025.

\bibitem{sepehri2024serpent}
Mohammad~Shahab Sepehri, Zalan Fabian, and Mahdi Soltanolkotabi.
\newblock Serpent: Scalable and efficient image restoration via multi-scale structured state space models.
\newblock {\em arXiv preprint arXiv:2403.17902}, 2024.

\bibitem{wen2025matir}
Juan Wen, Weiyan Hou, Luc Van~Gool, and Radu Timofte.
\newblock Matir: A hybrid mamba-transformer image restoration model.
\newblock {\em arXiv preprint arXiv:2501.18401}, 2025.

\bibitem{deng2024cu}
Rui Deng and Tianpei Gu.
\newblock Cu-mamba: Selective state space models with channel learning for image restoration.
\newblock In {\em 2024 IEEE 7th International Conference on Multimedia Information Processing and Retrieval (MIPR)}, pages 328--334. IEEE, 2024.

\bibitem{lu2024lfmamba}
Yao Lu, Shunzhou Wang, Ziqi Wang, Peiqi Xia, Tianfei Zhou, et~al.
\newblock Lfmamba: light field image super-resolution with state space model.
\newblock {\em arXiv preprint arXiv:2406.12463}, 2024.

\bibitem{fei2024scalable}
Zhengcong Fei, Mingyuan Fan, Changqian Yu, and Junshi Huang.
\newblock Scalable diffusion models with state space backbone.
\newblock {\em arXiv preprint arXiv:2402.05608}, 2024.

\bibitem{teng2024dim}
Yao Teng, Yue Wu, Han Shi, Xuefei Ning, Guohao Dai, Yu~Wang, Zhenguo Li, and Xihui Liu.
\newblock Dim: Diffusion mamba for efficient high-resolution image synthesis.
\newblock {\em arXiv preprint arXiv:2405.14224}, 2024.

\bibitem{hu2024zigma}
Vincent~Tao Hu, Stefan~Andreas Baumann, Ming Gui, Olga Grebenkova, Pingchuan Ma, Johannes Fischer, and Bj{\"o}rn Ommer.
\newblock Zigma: A dit-style zigzag mamba diffusion model.
\newblock In {\em European Conference on Computer Vision}, pages 148--166. Springer, 2024.

\bibitem{li2024scalable}
Haopeng Li, Jinyue Yang, Kexin Wang, Xuerui Qiu, Yuhong Chou, Xin Li, and Guoqi Li.
\newblock Scalable autoregressive image generation with mamba.
\newblock {\em arXiv preprint arXiv:2408.12245}, 2024.

\bibitem{chen2024maskmamba}
Wenchao Chen, Liqiang Niu, Ziyao Lu, Fandong Meng, and Jie Zhou.
\newblock Maskmamba: A hybrid mamba-transformer model for masked image generation.
\newblock {\em arXiv preprint arXiv:2409.19937}, 2024.

\bibitem{fei2024dimba}
Zhengcong Fei, Mingyuan Fan, Changqian Yu, Debang Li, Youqiang Zhang, and Junshi Huang.
\newblock Dimba: Transformer-mamba diffusion models.
\newblock {\em arXiv preprint arXiv:2406.01159}, 2024.

\bibitem{mo2024efficient}
Shentong Mo.
\newblock Efficient 3d shape generation via diffusion mamba with bidirectional ssms.
\newblock {\em arXiv preprint arXiv:2406.05038}, 2024.

\bibitem{shen2025gamba}
Qiuhong Shen, Zike Wu, Xuanyu Yi, Pan Zhou, Hanwang Zhang, Shuicheng Yan, and Xinchao Wang.
\newblock Gamba: Marry gaussian splatting with mamba for single-view 3d reconstruction.
\newblock {\em IEEE Transactions on Pattern Analysis and Machine Intelligence}, 2025.

\bibitem{fei2024diffusion}
Zhengcong Fei, Mingyuan Fan, Changqian Yu, Debang Li, and Junshi Huang.
\newblock Diffusion-rwkv: Scaling rwkv-like architectures for diffusion models.
\newblock {\em arXiv preprint arXiv:2404.04478}, 2024.

\bibitem{yang2024sdit}
Shu Yang, Hanzhi Ma, Chengting Yu, Aili Wang, and Er-Ping Li.
\newblock Sdit: Spiking diffusion model with transformer.
\newblock {\em arXiv preprint arXiv:2402.11588}, 2024.

\bibitem{xing2024segmamba}
Zhaohu Xing, Tian Ye, Yijun Yang, Guang Liu, and Lei Zhu.
\newblock Segmamba: Long-range sequential modeling mamba for 3d medical image segmentation.
\newblock In {\em International Conference on Medical Image Computing and Computer-Assisted Intervention}, pages 578--588. Springer, 2024.

\bibitem{chen2025zig}
Tianxiang Chen, Xudong Zhou, Zhentao Tan, Yue Wu, Ziyang Wang, Zi~Ye, Tao Gong, Qi~Chu, Nenghai Yu, and Le~Lu.
\newblock Zig-rir: Zigzag rwkv-in-rwkv for efficient medical image segmentation.
\newblock {\em IEEE Transactions on Medical Imaging}, 2025.

\bibitem{zhou2024bsbp}
Xudong Zhou and Tianxiang Chen.
\newblock Bsbp-rwkv: Background suppression with boundary preservation for efficient medical image segmentation.
\newblock In {\em Proceedings of the 32nd ACM International Conference on Multimedia}, pages 4938--4946, 2024.

\bibitem{yang2024mambamil}
Shu Yang, Yihui Wang, and Hao Chen.
\newblock Mambamil: Enhancing long sequence modeling with sequence reordering in computational pathology.
\newblock In {\em International Conference on Medical Image Computing and Computer-Assisted Intervention}, pages 296--306. Springer, 2024.

\bibitem{zou2024mmr}
Jing Zou, Lanqing Liu, Qi~Chen, Shujun Wang, Xiaohan Xing, and Jing Qin.
\newblock Mmr-mamba: Multi-contrast mri reconstruction with mamba and spatial-frequency information fusion.
\newblock {\em arXiv e-prints}, pages arXiv--2406, 2024.

\bibitem{huang2024mambamir}
Jiahao Huang, Liutao Yang, Fanwen Wang, Yang Nan, Angelica~I Aviles-Rivero, Carola-Bibiane Sch{\"o}nlieb, Daoqiang Zhang, and Guang Yang.
\newblock Mambamir: An arbitrary-masked mamba for joint medical image reconstruction and uncertainty estimation.
\newblock {\em arXiv preprint arXiv:2402.18451}, 2024.

\bibitem{wang2024vmambamorph}
Ziyang Wang, Jian-Qing Zheng, Chao Ma, and Tao Guo.
\newblock Vmambamorph: a visual mamba-based framework with cross-scan module for deformable 3d image registration.
\newblock {\em arXiv e-prints}, pages arXiv--2404, 2024.

\bibitem{atli2024i2i}
Omer~F Atli, Bilal Kabas, Fuat Arslan, Arda~C Demirtas, Mahmut Yurt, Onat Dalmaz, and Tolga Cukur.
\newblock I2i-mamba: Multi-modal medical image synthesis via selective state space modeling.
\newblock {\em arXiv preprint arXiv:2405.14022}, 2024.

\bibitem{lu2025delta}
Rongchang Lu, Bingcheng Liao, Haowen Hou, Jiahang Lv, and Xin Hai.
\newblock Delta-wkv: A novel meta-in-context learner for mri super-resolution.
\newblock {\em arXiv preprint arXiv:2502.20852}, 2025.

\bibitem{wang2024occrwkv}
Junming Wang, Wei Yin, Xiaoxiao Long, Xingyu Zhang, Zebin Xing, Xiaoyang Guo, and Qian Zhang.
\newblock Occrwkv: Rethinking efficient 3d semantic occupancy prediction with linear complexity.
\newblock {\em arXiv preprint arXiv:2409.19987}, 2024.

\bibitem{chen2025h}
Siran Chen, Yuxiao Luo, Yue Ma, Yu~Qiao, and Yali Wang.
\newblock H-mba: Hierarchical mamba adaptation for multi-modal video understanding in autonomous driving.
\newblock {\em arXiv preprint arXiv:2501.04302}, 2025.

\bibitem{huang2025trajectory}
Yizhou Huang, Yihua Cheng, and Kezhi Wang.
\newblock Trajectory mamba: Efficient attention-mamba forecasting model based on selective ssm.
\newblock In {\em Proceedings of the Computer Vision and Pattern Recognition Conference}, pages 12058--12067, 2025.

\bibitem{yuan2024drama}
Chengran Yuan, Zhanqi Zhang, Jiawei Sun, Shuo Sun, Zefan Huang, Christina Dao~Wen Lee, Dongen Li, Yuhang Han, Anthony Wong, Keng~Peng Tee, et~al.
\newblock Drama: An efficient end-to-end motion planner for autonomous driving with mamba.
\newblock {\em arXiv preprint arXiv:2408.03601}, 2024.

\bibitem{zhao2025salm2}
Chunyu Zhao, Wentao Mu, Xian Zhou, Wenbo Liu, Fei Yan, and Tao Deng.
\newblock Salm$^2$: An extremely lightweight saliency mamba model for real-time cognitive awareness of driver attention.
\newblock In {\em Proceedings of the AAAI Conference on Artificial Intelligence}, volume~39, pages 1647--1655, 2025.

\bibitem{zhao2024rs}
Sijie Zhao, Hao Chen, Xueliang Zhang, Pengfeng Xiao, Lei Bai, and Wanli Ouyang.
\newblock Rs-mamba for large remote sensing image dense prediction.
\newblock {\em IEEE Transactions on Geoscience and Remote Sensing}, 2024.

\bibitem{zhu2024samba}
Qinfeng Zhu, Yuanzhi Cai, Yuan Fang, Yihan Yang, Cheng Chen, Lei Fan, and Anh Nguyen.
\newblock Samba: Semantic segmentation of remotely sensed images with state space model.
\newblock {\em Heliyon}, 10(19), 2024.

\bibitem{chen2024changemamba}
Hongruixuan Chen, Jian Song, Chengxi Han, Junshi Xia, and Naoto Yokoya.
\newblock Changemamba: Remote sensing change detection with spatio-temporal state space model.
\newblock {\em IEEE Transactions on Geoscience and Remote Sensing}, 2024.

\bibitem{ma2024rs3mamba}
Xianping Ma, Xiaokang Zhang, and Man-On Pun.
\newblock Rs 3 mamba: Visual state space model for remote sensing image semantic segmentation.
\newblock {\em IEEE Geoscience and Remote Sensing Letters}, 2024.

\bibitem{yang2024hsimamba}
Judy~X Yang, Jun Zhou, Jing Wang, Hui Tian, and Alan Wee~Chung Liew.
\newblock Hsimamba: Hyperpsectral imaging efficient feature learning with bidirectional state space for classification.
\newblock {\em arXiv preprint arXiv:2404.00272}, 2024.

\bibitem{he2024pan}
Xuanhua He, Ke~Cao, Jie Zhang, Keyu Yan, Yingying Wang, Rui Li, Chengjun Xie, Danfeng Hong, and Man Zhou.
\newblock Pan-mamba: Effective pan-sharpening with state space model.
\newblock {\em Information Fusion}, 115:102779, 2025.

\bibitem{cao2024novel}
Zihan Cao, Xiao Wu, Liang-Jian Deng, and Yu~Zhong.
\newblock A novel state space model with local enhancement and state sharing for image fusion.
\newblock In {\em Proceedings of the 32nd ACM International Conference on Multimedia}, pages 1235--1244, 2024.

\bibitem{zhou2024rsdehamba}
Huiling Zhou, Xianhao Wu, Hongming Chen, Xiang Chen, and Xin He.
\newblock Rsdehamba: lightweight vision mamba for remote sensing satellite image dehazing.
\newblock {\em arXiv preprint arXiv:2405.10030}, 2024.

\bibitem{xiao2024frequency}
Yi~Xiao, Qiangqiang Yuan, Kui Jiang, Yuzeng Chen, Qiang Zhang, and Chia-Wen Lin.
\newblock Frequency-assisted mamba for remote sensing image super-resolution.
\newblock {\em IEEE Transactions on Multimedia}, 2024.

\bibitem{liu2024rscama}
Chenyang Liu, Keyan Chen, Bowen Chen, Haotian Zhang, Zhengxia Zou, and Zhenwei Shi.
\newblock Rscama: Remote sensing image change captioning with state space model.
\newblock {\em IEEE Geoscience and Remote Sensing Letters}, 2024.

\bibitem{erol2024audio_mamba}
Mehmet~Hamza Erol, Arda Senocak, Jiu Feng, and Joon~Son Chung.
\newblock Audio mamba: Bidirectional state space model for audio representation learning.
\newblock {\em IEEE Signal Processing Letters}, 31:2975--2979, 2024.

\bibitem{yadav2024audio}
Sarthak Yadav and Zheng-Hua Tan.
\newblock Audio mamba: Selective state spaces for self-supervised audio representations.
\newblock {\em arXiv preprint arXiv:2406.02178}, 2024.

\bibitem{shams2024ssamba}
Siavash Shams, Sukru~Samet Dindar, Xilin Jiang, and Nima Mesgarani.
\newblock Ssamba: Self‑supervised audio representation learning with mamba state space model.
\newblock In {\em arXiv preprint arXiv:2405.11831}, 2024.

\bibitem{chao2024investigation}
Rong Chao, Wen‐Huang Cheng, Moreno La~Quatra, Sabato~Marco Siniscalchi, Chao‐Han~Huck Yang, Szu‐Wei Fu, and Yu~Tsao.
\newblock An investigation of incorporating mamba for speech enhancement.
\newblock arXiv preprint arXiv:2405.06573, 2024.

\bibitem{sui2024tramba}
Yueyuan Sui, Minghui Zhao, Junxi Xia, Xiaofan Jiang, and Stephen Xia.
\newblock Tramba: A hybrid transformer and mamba architecture for practical audio and bone conduction speech super resolution and enhancement on mobile and wearable platforms.
\newblock {\em Proceedings of the ACM on Interactive Mobile, Wearable and Ubiquitous Technologies}, 8(4):1--29, 2024.

\bibitem{quan2024multichannel_speech_enhancement}
Changsheng Quan and Xiaofei Li.
\newblock Multichannel long‑term streaming neural speech enhancement for static and moving speakers.
\newblock {\em arXiv preprint arXiv:2403.07675}, 2024.

\bibitem{huang2024av}
Ziru Huang, Jia Li, Wenjie Zhao, Yunhui Guo, and Yapeng Tian.
\newblock Av-mamba: Cross-modality selective state space models for audio-visual question answering.
\newblock In {\em Proceedings of the IEEE/CVF Conference on Computer Vision and Pattern Recognition Workshop (CVPRW)}, pages 1--4, 2024.

\bibitem{gu2024rwkv}
Tiancheng Gu, Kaicheng Yang, Xiang An, Ziyong Feng, Dongnan Liu, Weidong Cai, and Jiankang Deng.
\newblock Rwkv-clip: a robust vision-language representation learner.
\newblock {\em arXiv preprint arXiv:2406.06973}, 2024.

\bibitem{mustafa2022multimodal}
Basil Mustafa, Carlos Riquelme, Joan Puigcerver, Rodolphe Jenatton, and Neil Houlsby.
\newblock Multimodal contrastive learning with limoe: the language-image mixture of experts.
\newblock {\em Advances in Neural Information Processing Systems}, 35:9564--9576, 2022.

\bibitem{li2025uni}
Yunxin Li, Shenyuan Jiang, Baotian Hu, Longyue Wang, Wanqi Zhong, Wenhan Luo, Lin Ma, and Min Zhang.
\newblock Uni-moe: Scaling unified multimodal llms with mixture of experts.
\newblock {\em IEEE Transactions on Pattern Analysis and Machine Intelligence}, 2025.

\bibitem{li2023pace}
Yunshui Li, Binyuan Hui, ZhiChao Yin, Min Yang, Fei Huang, and Yongbin Li.
\newblock Pace: Unified multi-modal dialogue pre-training with progressive and compositional experts.
\newblock {\em arXiv preprint arXiv:2305.14839}, 2023.

\bibitem{yue2024medmamba}
Yubiao Yue and Zhenzhang Li.
\newblock Medmamba: Vision mamba for medical image classification.
\newblock {\em arXiv preprint arXiv:2403.03849}, 2024.

\bibitem{chen2024rsmamba}
Keyan Chen, Bowen Chen, Chenyang Liu, Wenyuan Li, Zhengxia Zou, and Zhenwei Shi.
\newblock Rsmamba: Remote sensing image classification with state space model.
\newblock {\em IEEE Geoscience and Remote Sensing Letters}, 2024.

\bibitem{dong2024fusion}
Wenhao Dong, Haodong Zhu, Shaohui Lin, Xiaoyan Luo, Yunhang Shen, Xuhui Liu, Juan Zhang, Guodong Guo, and Baochang Zhang.
\newblock Fusion-mamba for cross-modality object detection.
\newblock {\em arXiv preprint arXiv:2404.09146}, 2024.

\bibitem{gao2024mamba}
Ruisheng Gao, Zeyu Xiao, and Zhiwei Xiong.
\newblock Mamba-based light field super-resolution with efficient subspace scanning.
\newblock In {\em Proceedings of the Asian Conference on Computer Vision}, pages 531--547, 2024.

\bibitem{nasiri2024vision}
Ali Nasiri-Sarvi, Mahdi~S Hosseini, and Hassan Rivaz.
\newblock Vision mamba for classification of breast ultrasound images.
\newblock In {\em Deep Breast Workshop on AI and Imaging for Diagnostic and Treatment Challenges in Breast Care}, pages 148--158. Springer, 2024.

\bibitem{ji2023rnn}
Gaoyuan Ji and Pei Liu.
\newblock Rnn-based multiple instance learning for the classification of histopathology whole slide images.
\newblock In {\em International Conference on Medical Imaging and Computer-Aided Diagnosis}, pages 329--339. Springer, 2023.

\bibitem{zhang2024motion2}
Yuelin Zhang, Kim Yan, Chun~Ping Lam, Chengyu Fang, Wenxuan Xie, Yufu Qiu, Raymond Shing-Yan Tang, and Shing~Shin Cheng.
\newblock Motion-guided dual-camera tracker for endoscope tracking and motion analysis in a mechanical gastric simulator.
\newblock {\em arXiv preprint arXiv:2403.05146}, 2024.

\bibitem{meng2025enhancing}
Zelin Meng and Takanori Fukao.
\newblock Enhancing autonomous driving perception with mamba-based dual-branch depth estimation.
\newblock {\em IEEE Transactions on Intelligent Transportation Systems}, 2025.

\bibitem{an2023exploring}
Keyu An and Shiliang Zhang.
\newblock Exploring rwkv for memory efficient and low latency streaming asr.
\newblock {\em arXiv preprint arXiv:2309.14758}, 2023.

\end{thebibliography}





\end{document}